\documentclass{article}

\usepackage{arxiv}

\usepackage[utf8]{inputenc} 
\usepackage[T1]{fontenc}    
\usepackage{hyperref}       
\usepackage{url}            
\usepackage{multirow}
\usepackage{bm}
\usepackage[square,numbers]{natbib}
\usepackage{amsmath}
\usepackage{booktabs}       
\usepackage{amsfonts}       
\usepackage{nicefrac}       
\usepackage{microtype}      
\usepackage{lipsum}		
\usepackage{graphicx}
\usepackage{natbib}
\usepackage{doi}

\usepackage{float}

\title{Deep Learning Approaches on Image Captioning: A Review}

\author{ {Taraneh Ghandi} \\
	Faculty of Engineering\\
    McMaster University\\
    Hamilton, Ontario, Canada\\
	\texttt{ghandit@mcmaster.ca} \\
	\And
	{Hamidreza Pourreza} \\
	Computer Engineering Department\\
	Ferdowsi University of Mashhad\\
	Mashhad, Iran \\
	\texttt{hpourreza@um.ac.ir} \\
    \And
    {Hamidreza Mahyar}\thanks{Corresponding author}\\
    Faculty of Engineering\\
    McMaster University\\
    Hamilton, Ontario, Canada\\
    \texttt{mahyarh@mcmaster.ca} \\
}

\renewcommand{\headeright}{}
\renewcommand{\undertitle}{}
\renewcommand{\shorttitle}{Deep Learning Approaches on Image Captioning: A Review}

\hypersetup{
pdftitle={DeepLearningApproachesOnImageCaptioning:AReview},
pdfsubject={q-bio.NC, q-bio.QM},
pdfauthor={Taraneh Ghandi, Hamidreza Pourreza, Hamidreza Mahyar},
pdfkeywords={Image Captioning, Deep Learning, Text Generation, Neural Networks, Machine translation},
}

\begin{document}
\maketitle

\begin{abstract}
	 Image captioning is a research area of immense importance, aiming to generate natural language descriptions for visual content in the form of still images. The advent of deep learning and more recently vision-language pre-training techniques has revolutionized the field, leading to more sophisticated methods and improved performance. In this survey paper, we provide a structured review of deep learning methods in image captioning by presenting a comprehensive taxonomy and discussing each method category in detail.
Additionally, we examine the datasets commonly employed in image captioning research, as well as the evaluation metrics used to assess the performance of different captioning models.
We address the challenges faced in this field by emphasizing issues such as object hallucination, missing context, illumination conditions, contextual understanding, and referring expressions. We rank different deep learning methods' performance according to widely used evaluation metrics, giving insight into the current state of the art.
Furthermore, we identify several potential future directions for research in this area, which include tackling the information misalignment problem between image and text modalities, mitigating dataset bias, incorporating vision-language pre-training methods to enhance caption generation, and developing improved evaluation tools to accurately measure the quality of image captions. 
\end{abstract}


\section{Introduction}
Automatic image captioning is a critical research problem with numerous complexities, attracting a significant amount of work with extensive applications across various domains such as human-computer interaction \cite{li2020oscar, fukui2016multimodal, zhang2021vinvl}, medical image captioning and prescription \cite{pavlopoulos2019survey, huang2021contextualized, ayesha2021automatic}, traffic data analysis \cite{li2020traffic}, quality control in industry \cite{luo2019visual}, and especially assistive technologies for visually impaired individuals \cite{gurari2020captioning, sidorov2020textcaps, dognin2020image, ahsan2021multi, makav2019new}. The field has undergone a revolutionary transformation with the development and growth of deep learning techniques \cite{vaswani2017attention,radford2021learning}, resulting in the emergence of advanced methods and enhanced performance.
Automatic image captioning lies at the intersection of natural language processing and computer vision. This field of research deals with the creation of textual descriptions for images without human intervention. Given an input image $I$, the goal is to generate a caption $C$ describing the visual contents present inside the given image, with $C$ being a set of sentences $C=\{c_1, c_2, ..., c_n\}$ where each $c_i$ is a sentence of the generated caption $C$.

Given the recent advancements in the domain of image captioning, an updated review of the more recent research works can assist researchers in keeping up with the latest progress in this field. There exist numerous literature reviews and surveys on image captioning, providing an extensive collection of research conducted in previous years. Notably, Hossain et al. \cite{hossain2019comprehensive} authored a comprehensive survey paper that served as an inspiration for this work's structure. However, instead of a pairwise comparison like in \cite{hossain2019comprehensive}, we have organized our paper to feature a separate section for each method category and follow the same order of category in our discussion (section \ref{discussion_section}). Furthermore, most surveys typically cover works dating from 2018 and earlier, while more recent research is yet to be addressed. Some surveys \cite{elhagry2021a} are limited in the number of research works covered, while others \cite{chohan2020image} do not delve into the methodologies' specific details. Additionally, considering the recent advancements of vision language pre-training methods, image captioning methods that fall under this category must be addressed and discussed. This category has seldom been explored in previous survey works.
The field of image captioning can be classified into multiple categories which differ in the captioning settings, such as dense captioning methods which provide captions for each entity presented in the image or whole image captioning methods that provide captions for the entirety of the input image. Here, we focus on reviewing "whole image" captioning methods.  

In this paper, we discuss various methods of image captioning introduced in papers published from 2018 to 2022, followed by the most common problems and challenges of image captioning. We provide a comprehensive analysis of each method, covering widely used datasets and evaluation metrics. We also compare the performance of the different covered methods before exploring future directions in the field. The section on problems and challenges provides a detailed overview of the inherent difficulties in image captioning and provides insight into potential solutions to address them. We hope to provide a thorough understanding of image captioning through this review and encourage continued progress in the field.


\section{Common Solutions and Techniques}
Automatic image captioning is usually computationally intensive and structurally complicated. Therefore, it is necessary to study and observe the different methods of solving this problem in order to propose a practical and efficient solution. Despite the recent remarkable advances in hardware design and optimization techniques, utilizing sensible methods and tools is still of vital importance. This section discusses some of the standard solutions and techniques used in image captioning. 

\subsection{Most Common Solutions}
Image captioning is presented chiefly as a sequence-to-sequence problem in machine vision. In sequence-to-sequence problems, the goal is to convert a specific sequence to the appropriate corresponding sequence. One of the essential sequence-to-sequence problems is machine translation. In machine translation, a sequence of words (e.g., a sentence) in a language is translated to its alternative in another language. In order to learn the correspondence between the sequences, the sequences are mapped into a common space in which the distance between two sequences with close meaning is small. 
One of the most common solutions to the image captioning problem is inspired by machine translation and has given promising results according to the performance metrics. In this class of methods, known as the "encoder-decoder" methods, the input image is mapped to an intermediate representation of the image contents. It is then converted to a sequence of words that make up the caption of the image. In the encoding stage, Convolutional Neural Networks (CNNs) are frequently used to detect objects in the image since the last convolutional layer of these networks provides a rich representation of an image.
This layer is used as a feature vector (or multiple feature vectors obtained from different regions in the image). In the decoding stage, recurrent neural networks (RNNs) are commonly used due to their ability to give a proper representation of the human language and texts. After the image and its corresponding caption in the dataset are mapped into a common space, the correspondence between the two representations is learned and new captions are generated for new images.
Despite significant results, these methods usually give general and vague captions for images and do not describe image contents appropriately since all information is compressed into a single vector. This causes problems with learning the information at the beginning of the sequence and the deeper relations between image contents. Many new methods have been proposed to solve these problems, most of them having the encoder-decoder structure as their core component. These methods and their features and possible flaws are discussed later in this survey. 
In addition to these methods, other methods, such as dense captioning \cite{johnson2016densecap}, have been proposed to solve the image captioning problem. However, in recent works on image captioning dating from 2018 to 2022, these methods are seldom used, and methods based on the attention mechanism and graphs have been used more frequently.

\subsection{Some of the Widely Used Techniques in Image Captioning}
Before introducing various methods of image captioning in detail, we discuss some of the frequently used techniques in image captioning methods.

\subsubsection{R-CNNs} When detecting objects, an especially trained Convolutional Neural Network also detects the object’s bounding box inside the image. If a simple CNN is used for object detection, using a grid above the image and processing the individual cells of the grid is one way to detect object bounding boxes. However, objects that appear in images are of various shapes and sizes and can be located anywhere inside the image; therefore, one type of grid with fixed cell sizes will not give desirable results. In order to resolve this issue, grids with different cell sizes must be used to detect objects with different settings, which will be computationally intensive.

To solve this issue, Girshick et al. introduced Region-based CNNs (R-CNN) \cite{girshick2014rich}. These networks use selective search to extract only 2000 regions from the image. The regions are given the term: \textit{region proposals}. The selective search algorithm generates many candidate regions to segment the input image. These regions are merged recursively and form larger regions which are then selected as the final region proposals.
Since no form of actual learning is used inside the selective search algorithm, the algorithm may produce incorrect region proposals.

\textbf{Fast R-CNN: }The same authors that introduced R-CNN introduced another network under the name of Fast R-CNN \cite{girshick2015fast}. The working of Fast R-CNN is very similar to that of R-CNN. However, instead of feeding the region proposals to a CNN, the input image is fed to the CNN to produce a convolutional feature map. Region proposals are then generated using this feature map and the selective search algorithm. 
Fast R-CNN is faster than R-CNN in that the image is only convolved once, and a feature map is extracted, as opposed to R-CNN that fed 2000 region proposals to a CNN.

\textbf{Faster R-CNN: }Despite Fast R-CNN being faster than R-CNN, both use the selective search algorithm, which is time-consuming and affects the network's performance. In \cite{ren2015faster}, a new method was presented in which the network learns the region proposals and does not use the selective search algorithm. Similar to Fast R-CNN, an image is given to the CNN as input, and a convolutional feature map is extracted. Instead of the selective search algorithm, a separate network is used to predict region proposals. 
Faster R-CNN is significantly faster than R-CNN and Fast R-CNN and can be used in real-time object detection. Faster R-CNN is used in many papers covered in this survey to generate a presentation for the input image.

\subsubsection{RNNs}
Recurrent neural networks (RNNs) \cite{rumelhart1985learning, osti_6910294} are a type of artificial neural network that also have internal memory. These networks are recurrent in the sense that they perform the same operation for each input data, and the output from the current input data is dependent on the computations from previous steps. After the output is computed, it is fed back into the network. RNN uses current input and the learned output from the previous step during inference time. In contrast to the feed-forward neural networks, RNNs use their internal memory (also known as the internal state) to process a sequence of input. This feature enables RNNs to perform tasks such as human handwriting or speech detection. In other words, RNNs can be used in applications in which the inputs are related to each other in some way and are not independent.
On the other hand, these networks are prone to problems such as "Vanishing Gradient," difficulty in training, and the inability to process long sequences. If a sequence is lengthy, RNN networks might lose parts of the information at the beginning of the sequence.
\subsubsection{LSTMs and GRUs}
An improved version of RNN networks is the "Long Short-Term Memory Network (LSTM)"\cite{hochreiter1997long}. LSTMs have been explicitly designed to resolve long-term dependency problems \cite{Gers2000learning}. Since LSTMs are effective at capturing long-term temporal dependencies without such optimization issues, they have been used to solve many challenging problems, including handwriting recognition/generation, language modeling/translation, speech synthesis, acoustic modeling of speech, protein secondary structure prediction, and audio and video analysis.\cite{Greff_2017} These networks have an internal "Gate" mechanism that can control and modify the flow of information. These gates can learn which data is essential in a sequence or must be ignored. Therefore, important information is stored all through the sequence. Despite the advantages of LSTMs over RNN, LSTMs ignore the hierarchical structure in a sentence. Also, LSTMs need plenty of storage space due to their memory cells.
LSTM networks are widely used in "Encoder-Decoder"-based methods in image captioning to generate representations of the captions, which are textual data. Another network similar to LSTMs is the "Gated Recurrent Unit (GRU)"\cite{cho2014learning}. These networks are similar to LSTMs in structure but use fewer gates to control the flow of information. Fewer parameters allow GRUs to train easier and faster than their LSTM counterparts. GRUs have been shown to perform better on specific smaller and less frequent datasets \cite{gruber2020gru}.

\subsubsection{ResNet}
Residual neural networks or ResNets are used in object detection \cite{he2016deep}. The structure of these networks is inspired by the pyramidal cells in the cerebral cortex and uses slip connections to connect multiple layers. Usually, ResNets are implemented with two or three skip connections. ResNets are built with "Residual Blocks" placed on top of each other; for example, ResNet-50 consists of 50 layers containing residual blocks. Optimization of this structure is shown to be easier and faster compared to the simple network structure (without skip connections and residual blocks). Some of the research works presented in this survey have used ResNet to detect objects and generate image representations.

\section{Deep Learning-Based Image Captioning}\label{deep_learning_based_image_captioning}
\begin{figure}[h]
  \centering
  \includegraphics[width=0.8\textwidth]{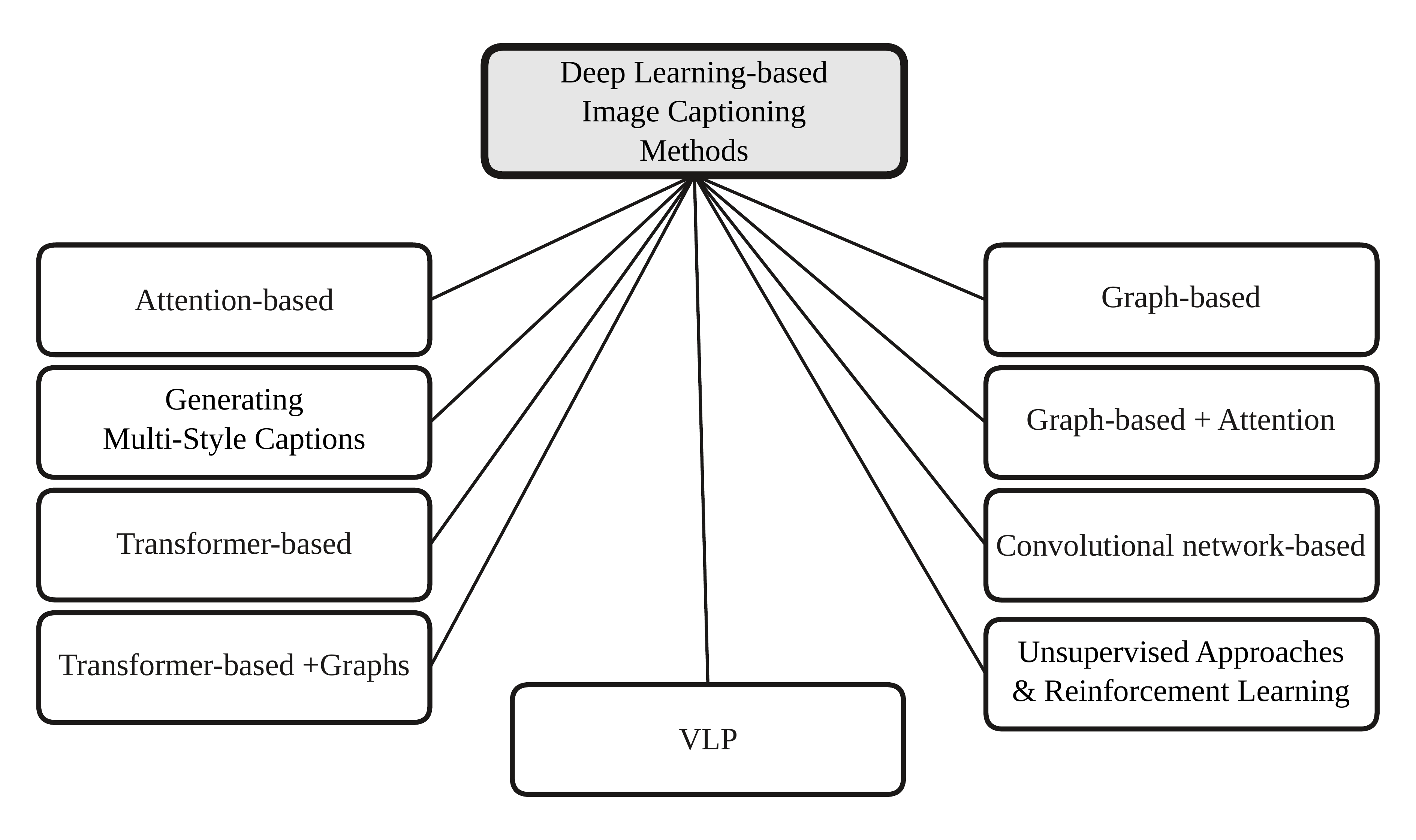}
  \caption{The taxonomy of the image captioning methods covered in this survey paper.}
  \label{fig_general_framework}
\end{figure}

In this section, we have organized and classified the different frameworks, methods, and approaches which were extensively used in recent research works based on their core structure. Some terms and notations in the covered papers have been altered to maintain consistency throughout this review. A figure demonstrating the taxonomy provided in this paper is shown in \ref{fig_general_framework}.
\subsection{Attention-Based Methods}\label{section_attention}
The methods that fall under the attention-based category utilize attention mechanisms to emphasize the most relevant parts of the input image when generating captions.

Attention-based methods \cite{bahdanau2014neural} are inspired by the human attention pattern and the way the human eye focuses on images. When inspecting images, humans focus more on the image's salient features. The same mechanism is implemented in attention-based mechanisms. During the training process, the model is shown "where to look at." To understand the mechanism of attention-based methods, one can imagine a sequential decoder in which, in addition to the previous cell's output and internal state, there is also a context vector under the term 'c.'

\begin{figure}[H]
  \centering
  \includegraphics[width=0.7\textwidth]{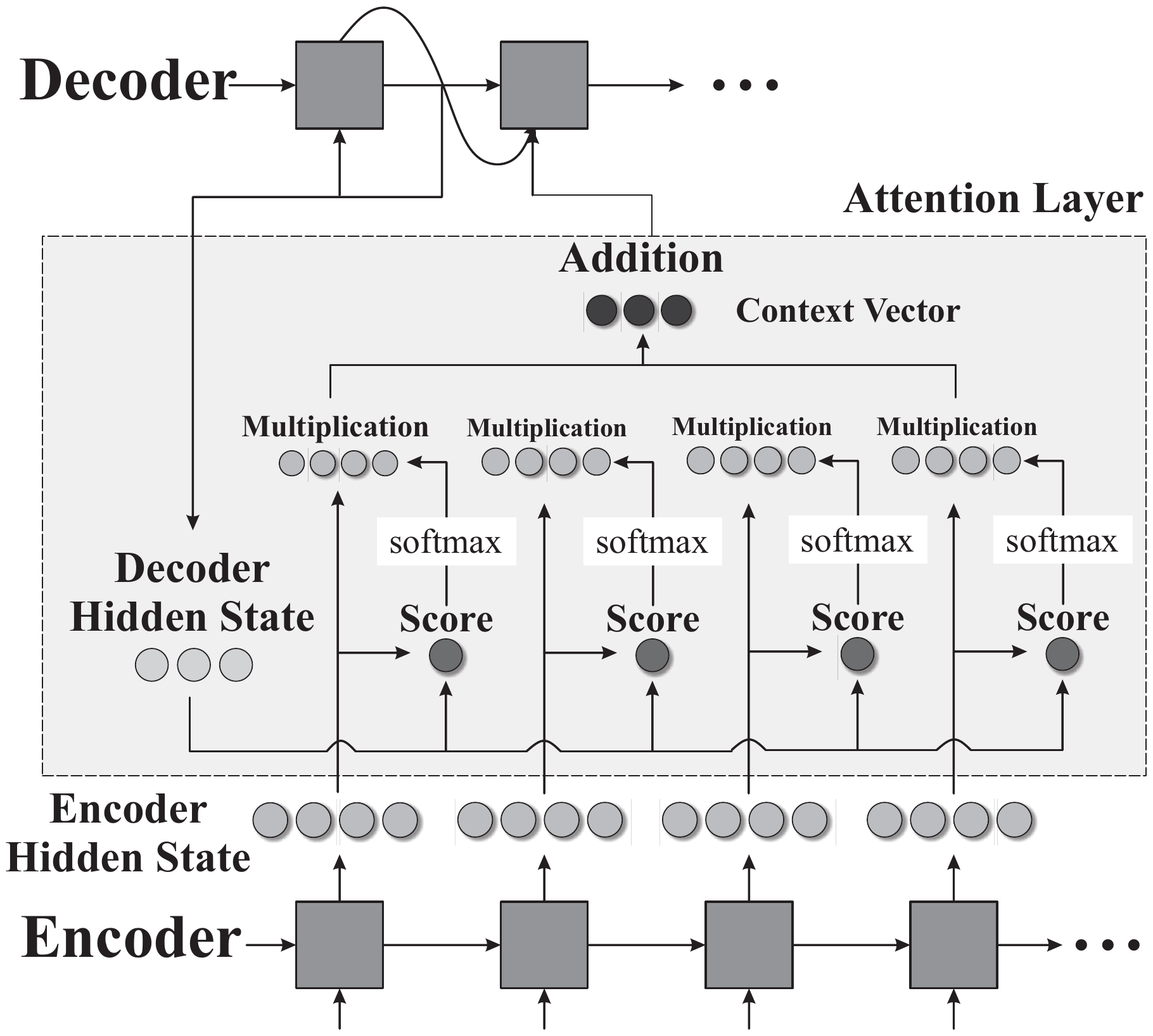}
  \caption{The attention mechanism in an encoder-decoder framework is typically used in machine translation.\cite{attention_mechanism_figure}}
  
  \label{fig_attention}
\end{figure}

\begin{figure}[H]
  \centering
  \includegraphics[width=0.78\textwidth]{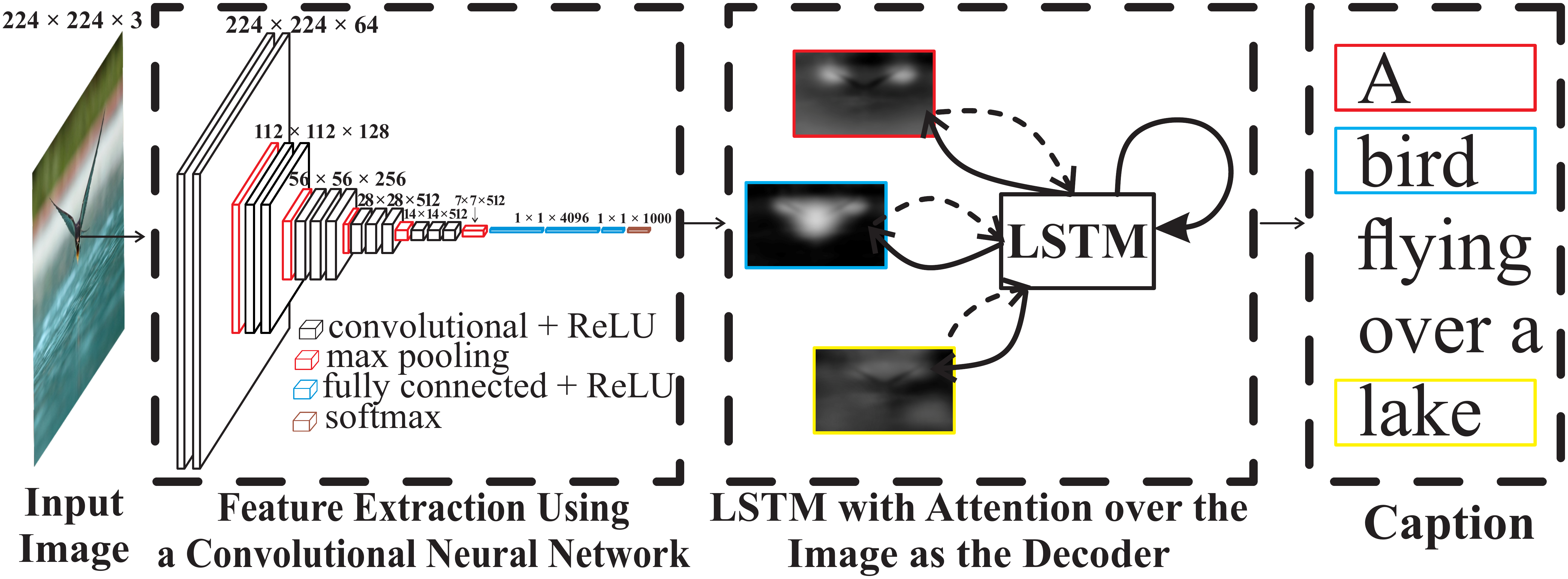}
  \caption{The basis of attention-based methods (best viewed in color). }
 
  \label{fig_attention_workflow}
\end{figure}

Vector c is the weighted sum of hidden states in the encoder.
\begin{equation}
  c_{i}=\sum_{j=1}^{T_{x}} a_{i j} h_{j}
\end{equation}
In the statement above, $a_{ij}$ is the "amount of attention" that output $i$ must pay to input $j$, and $h_j$ is the encoder state in input $j$. $a_{ij}$ is obtained by calculating softmax over attention amounts that are shown with $e$ on inputs and for output $i$:
\begin{equation}
  a_{i j}=\operatorname{softmax}\left(e_{i j}\right)=\frac{\exp \left(e_{i j}\right)}{\sum_{k=1}^{T_{x}} \exp \left(e_{i k}\right)}
\end{equation}

\begin{equation}
  e_{i j}=f\left(S_{i-1}, h_{j}\right)
\end{equation}
where, $f$ is the model that determines how much input at $j$ and output $i$ are correlated, and $S_{i-1}$ is the hidden state from the previous time step. The model $f$ can be estimated with a small neural network and can be optimized with any gradient-based optimization techniques, such as gradient descent. A presentation of the attention mechanism used in an encoder-decoder framework typically used in machine translation is shown in Figure \ref{fig_attention}.
In short, attention-based image captioning methods generate a weighted sum of extracted feature vectors at each time step in their decoder that guides the decoder module. 
Similar to the encoder-decoder framework, attention-based methods were first introduced for the machine translation problem in \cite{bahdanau2014neural}. In most of the attention-based methods, a CNN or a region-based CNN is used in the encoding stage to provide a representation of the image, and an RNN is usually used in the decoding stage. A block diagram of the basis of attention-based methods (which was first proposed by Xu et al. \cite{xu2015show}) is shown in Figure \ref{fig_attention_workflow}. The last layer of a Convolutional Neural Network (Here, VGGnet by Simonyan et al. \cite{simonyan2014very})- just before max pooling- has been used to extract features from the image. The LSTM network \cite{hochreiter1997long} with attention has been used as the decoder. The multiple images surrounding the LSTM shown in this figure demonstrate the attention values over different regions of the image. The lighter areas mean a higher attention value. The colored outline of the generated words in the caption corresponds to the regions outlined by the same colors.

\textbf{Multi-Head Attention} Multi-head attention \cite{vaswani2017attention} is a module for the attention mechanisms, which runs through the attention mechanism several times in parallel. The attention outputs achieved by this method are then concatenated and linearly transformed into the expected dimension. Multiple attention heads help attend to the parts of the sequence which are different in nature, e.g., longer-term dependencies versus shorter-term dependencies. Multi-head attention can be defined as:
\begin{equation}
\begin{aligned}
  \text{MultiHead}(Q,K,V)=[\text{head}_1,...,\text{head}_n]W_0 \\
   \text{where}\ \text{head}_i=\text{Attention}(QW_i^Q,KW_i^K,VW_i^V)
   \end{aligned}
\end{equation}
Where the set of queries is packed together into the matrix $Q$, the keys and values are packed into matrices $K$ and $V$, and $W_i^Q$, $W_i^K$, $W_i^V$, and $W^Q$ are the parameter matrices: $W_i^Q \in \mathbb{R}^{d_{model}\times d_k}$, $W_i^K \in \mathbb{R}^{d_{model}\times d_k }$, $W_i^V \in \mathbb{R}^{d_{model}\times d_v }$ and $W_i^O \in \mathbb{R}^{hd_{v}\times d_{model} }$, $h$=the number of parallel attention layers.

\textbf{Soft Attention and Hard Attention}
Attention is usually implemented in two forms: soft or hard attention \cite{xu2015show}. In soft attention, weighted image features accounted for attention are used as input instead of using an image as an input to the LSTM \cite{hochreiter1997long}. Soft attention disregards irrelevant areas by multiplying the corresponding features map with a low weight. High attention areas keep the original value while low attention areas get closer to 0 (become dark in the visualization) \cite{xu2015show}.

Hard attention uses a stochastic sampling model. Sampling is performed to accurately calculate the gradient descent in the backpropagation, and the results are averaged using the Monte Carlo method. Monte Carlo performs end-to-end episodes to compute an average for all sampling results. The accuracy depends on the number of samples and sampling quality in hard attention. However, soft attention applies the regular backpropagation method to compute the gradient, which is easier to calculate. The accuracy is also subject to the assumption that the weighted average is a good representation of the area of attention.
Attention-based methods are widely used in the encoder-decoder framework. Most of the research works discussed in this survey have used it as their primary framework or have combined it with other methods to improve its performance.

Vinyals et al. \cite{vinyals2015show} have been the first to incorporate deep learning-based encoder-decoder framework for image captioning. The presented model in their work is inspired by machine translation, based on the findings that indicate that given a powerful sequence model, it is possible to achieve remarkable results by directly maximizing the probability of the correct translation. CNNs can produce a rich presentation of an input image by embedding it into a fixed-length vector. Vinyals et al. \cite{vinyals2015show} have presented a model that uses CNN as an image "encoder" by pre-training it for an image classification task first and using the last hidden layer as an input to an RNN "decoder" that generates sentences. The model is trained to maximize the likelihood of the target description sentence given the training image. This work has been used by many other researchers as a basis to expand upon and refine using other modules and techniques \cite{xu2015show}.


Anderson et al. proposed the "bottom-up and top-down" method in \cite{anderson2018bottom}. The bottom-up module proposes the salient regions in the image, and each of the proposed regions is represented as a convolutional feature vector. This module is implemented using Faster R-CNN \cite{ren2015faster}, which was discussed previously. Faster R-CNN works well as a "hard" attention mechanism since a small number of bounding box features are selected from a large number of configurations. Faster R-CNN network is initialized with ResNet-101 \cite{he2016deep} pre-trained for image classification on the ImageNet dataset. Faster R-CNN is then trained using the Visual Genome \cite{krishna2017visual} dataset. The top-down module, designed to caption images, contains two LSTM networks \cite{hochreiter1997long} with the standard implementation. The first LSTM network operates as a top-down visual attention model, and the second LSTM network operates as a language model. The top-down visual attention module estimates a distribution of attention over regions and calculates the extracted feature vector as a weighted sum over total region proposals. The captioning model takes a variably-sized set of $k$ image features: $V=\{v_1,...,v_k\}, v_i \in \mathbb{R}^D$ as input. Each image feature encodes a salient region of the image. These image features can be defined as the output of the bottom-up attention model or as the spatial output layer of a CNN. The input vector to the attention LSTM at each time step consists of the previous output of the language LSTM, the mean-pooled image features $\bar{v}=\frac{1}{k}\sum_{i} v_i$, and an encoding of the word generated previously all concatenated together. The input to the language model LSTM is composed of the attended image feature concatenated with the output of the attention LSTM. 

The two-layer LSTM \cite{hochreiter1997long} structure has also been used by Yao et al. in \cite{yao2018exploring} as the attention mechanism in the final stage (More detail on the workings of this paper is discussed in "Combining Attention-Based and Graph-Based Methods" (section \ref{section_combining_attention_graph})).

Gu et al. \cite{gu2018stack} have presented a multi-stage coarse-to-fine structure for image captioning. This structure contains multiple decoders that each work on the output of the decoder in the previous step, making the captions richer in every step. This paper has used the LSTM network \cite{hochreiter1997long} as the decoder. The structure comprises three LSTM networks, with the first LSTM presenting the coarse details at the first stage and reducing computations in the later stages. The other LSTMs operate as fine-level decoders. 
At each stage, attention weights and hidden vectors generated by the previous stage decoder are used as input to the next stage decoder.
The operation of the coarse decoder is based on the general and global features of the image. However, in many cases, each word belongs to a small region of the image only. Using the general features of the image might yield improper results due to the possible noise from unrelated regions. Therefore, a "Stacked Attention Model \cite{yang2016stacked}" is used to improve the performance of this coarse-to-fine structure. This model enables the structure to extract visual information from finer details for future word predictions. The stacked model generates a spatial map that determines the region of each predicted word. Using this stacked attention model, finer and more precise details are extracted, and noise is gradually reduced. Also, regions that are highly relevant to the words are determined.


Huang et al. \cite{huang2019attention} have introduced a new attention-based structure containing one more level of attention. The structure named "Attention on Attention (AoA)" generates an "Information Vector" and an "Attention Gate" with two linear transformations. 
An attention module $f_{att}(\textbf{Q},\textbf{K},\textbf{V})$ operates on some queries, keys and values denoted by $\textbf{Q}$, $\textbf{K}$, and $\textbf{V}$ respectively and generates some weighted average vectors denoted by $\bm{\hat{V}}$. The attention module measures the similarity between $\textbf{Q}$ and $\textbf{K}$ and uses this similarity score to calculate weighted average vectors over $\textbf{V}$, which is formulated as:

\begin{equation}
    a_{i,j}=f_{sim}(\bm{q_i}, \bm{k_j}), \alpha_{i,j}=\frac{e^{a_{i,j}}}{\sum_{j}e^{a_{i,j}}}
\end{equation}

\begin{equation}
    \bm{\hat{v_i}}=\sum_{j}\alpha_{i,j}\bm{v_j}
\end{equation}

where $\bm{q_i} \in \bm{Q}$ is the $i^{th}$ query, $\bm{k_j} \in \bm{K}$ and $\bm{v_j}\in \bm{V}$ are the $j^{th}$ key/value pair. $f_{sim}$ is a function that computes the similarity score of each $\bm{k_j}$ and $\bm{q_i}$, and $\bm{\hat{v_i}}$ is the attended vector for the query $\bm{q_i}$. Since the attention module produces a weighted average for each query regardless of the relation between $\bm{Q}$ and $\bm{K/V}$, the weighted average vector can be irrelevant or misleading information. The AoA module measures the relevance between the attention results and the query.
The information vector $\bm{i}$ is generated with a linear transformation on current content (caption) and results from the attention component, and stores both parts’ data. The attention gate $\bm{g}$ is generated from the content and the result from the attention component using sigmoid activation. The value inside each part (also called a channel) of this attention gate determines the level of importance of the channel in the information vector. Both the information vector and the attention gate is conditioned on the attention result and the current context (\emph{i.e.} the query) $\bm{q}$. The AoA structure adds another level of attention with element-wise multiplication of the attention gate and the information vector and finally produces the attended information, which contains useful data. The AoA structure is applied to both the encoder and decoder (termed AoANet): AoA is applied to the encoder after extracting image features to obtain the relation between objects present inside the image. AoA is also applied to the decoder to remove the attention results which are unrelated to the actual output or are ambiguous and leave the essential and useful results. The AoA structure has been introduced as an addition to the attention-based methods, and it can be applied to any attention method. In the experiments conducted by the authors, a Faster R-CNN \cite{ren2015faster} pre-trained on the ImageNet \cite{deng2009imagenet} and Visual Genome \cite{krishna2017visual} datasets is used to extract feature vectors from the image. 

Jiang et al. propose a novel recurrent fusion network (RFNet) in \cite{jiang2018recurrent} for the image captioning task, which uses multiple CNNs as encoders, and a recurrent fusion process is inserted after the encoders to produce better representations for the decoder. Each representation extracted from an individual image CNN can be regarded as an individual view depicting the input image. The fusion procedure consists of two stages; the first stage produces multiple sets of "thought vectors" by exploiting the interactions among the representations from multiple CNNs. The second stage performs multi-attention on the sets of thought vectors and generates a new set of thought vectors for the decoder. For the experiments, they use ResNet \cite{he2016deep}, DenseNet \cite{huang2017densely}, Inception-V3 \cite{szegedy2016rethinking}, Inception-V4 \cite{szegedy2017inception} and Inception-ResNet-V2 \cite{szegedy2017inception} as encoders to extract 5 groups of representations. Having considered reinforcement learning (RL) as a method to improve image captioning performance, they have trained their model with cross-entropy loss and fine-tuned the trained model with CIDEr optimization using reinforcement learning.

Incorporating attention in image captioning has transformed the field considerably, enabling more accurate and natural caption generation. However, they do not come without flaws. One problem with classic attention-based image captioning is that they do not consider the relations between the objects detected inside the image.

\subsubsection{Injecting Spatial and Semantic Relation Information into Attention-Based Methods}\label{section_attention_spatial_semantic}
A group of attention-based methods have included the spatial and semantic relations in an image to describe the content more appropriately.


In the method proposed by Pan et al. \cite{pan2020x},  
a new type of attention in the form of a unified block called "X-Linear Attention Block" is introduced, which uses bilinear pooling to emphasize salient image features and multimodal reasoning. This structure uses spatial and channel-wise bilinear attention to extract second-order interactions. The second-order interaction is obtained by calculating the outer product of the key (mapped image features) and the query (internal state of the sentence decoder) using bilinear pooling to consider all second-order interactions between keys and queries. After bilinear pooling, two embedding layers are used to predict the attention weights belonging to each region, and a softmax layer is then used to normalize the spatial attention vector. Also, a "squeeze-excitation" operation is performed on the embedded outer product (feature map). The squeezing process aggregates the feature map across spatial regions to produce a channel descriptor. The excitation process performs the self-gating mechanism with a sigmoid on the channel descriptor to obtain the channel-wise attention vector. Finally, the outer product of the key and query and the value from bilinear pooling is weighted summated with the spatial attention vector. Then, the channel-wise multiplication of this weighted sum and the channel attention vector is calculated and taken as the attended features.
Higher-order interactions can be computed by combining the X-linear attention blocks. This research work has used Faster R-CNN \cite{ren2015faster} to detect a set of regions. A stack of X-linear attention blocks is then used to encode region-level features of the image and the higher-order interactions between them to produce a set of enhanced region-level and image-level features. These attention blocks are used in the sentence decoder for multimodal reasoning. A new type of attention in the form of a unified block called "X-Linear Attention Block" is introduced, which uses bilinear pooling to emphasize salient image features and multimodal reasoning. This structure uses spatial and channel-wise bilinear attention to extract second-order interactions. The second-order interaction is obtained by calculating the outer product of the key (mapped image features) and the query (internal state of the sentence decoder) using bilinear pooling to consider all second-order interactions between keys and queries. After bilinear pooling, two embedding layers are used to predict the attention weights belonging to each region, and a softmax layer is then used to normalize the spatial attention vector. Also, a "squeeze-excitation" operation is performed on the embedded outer product (feature map). The squeezing process aggregates the feature map across spatial regions to produce a channel descriptor. The excitation process performs the self-gating mechanism with a sigmoid on the channel descriptor to obtain the channel-wise attention vector. Finally, the outer product of the key and query and the value from bilinear pooling is weighted summated with the spatial attention vector. Then, the channel-wise multiplication of this weighted sum and the channel attention vector is calculated and taken as the attended features. Higher-order interactions can be computed by combining the X-linear attention blocks. This research work has used Faster R-CNN \cite{ren2015faster} to detect a set of regions. A stack of X-linear attention blocks is then used to encode region-level features of the image and the higher-order interactions between them to produce a set of enhanced region-level and image-level features. These attention blocks are used in the sentence decoder for multimodal reasoning.

Cornia et al. \cite{cornia2019show} presented a method capable of describing an image by focusing on different regions in different orders following a given conditioning. By means of analyzing the syntactic dependencies between words, a higher level of abstraction can be recovered in which words can be organized into a tree-like structure. In a dependency tree, each word is linked together with its modifiers. Given a dependency tree, nouns can be grouped with their modifiers, thus building \emph{noun chunks}. The proposed model is built on a recurrent architecture which considers the decomposition of a sentence into noun chunks and models the relationship between image regions and textual chunks to ground the generation process on image regions explicitly. The model is conditioned on the input image $I$, and an ordered sequence of region sets $R$, which acts as a control signal and jointly predicts two output distributions corresponding to the word-level and chunk-level representation of the sentence. During the generation, the model keeps a pointer to the current region and can shift to the next element in $R$ using a boolean chunk-shifting gate $g_t$. To generate the output caption, a recurrent neural network with adaptive attention is used. The probability of switching to another chunk $p(g_t|R)$ is calculated in an adaptive mechanism in which an LSTM \cite{hochreiter1997long} computes a compatibility function between its internal state and a latent representation modeling the state of memory at the end of a chunk. The compatibility score is compared to that of attending one of the regions $r_t$, and the result is used as an indicator to switch to the next region set in $R$.

The addition of spatial and semantic relations to the attention-based framework has significantly improved the quality of the captions generated by the models. Despite the improvements achieved by this addition, some problems still remain, including the ambiguity of the captions, the lack of grounding, heavy computations associated with the object detectors, and the requirement of bounding-box annotations. In order to resolve some of these issues, other approaches to the image captioning problem have been introduced, which are explained and discussed in the following sections.

\subsection{Graph-Based Methods for Spatial and Semantic Relations between Image Elements}\label{section_graph_and_spatial_semantic}
The methods discussed in this section utilize scene graphs to better model the spatial and semantic relations between image elements. 

Due to their ability to represent relations between elements, graphs are used in applications in which the relations between elements are important \cite{bondy1976graph,wang2019role}. Studies have shown the effectiveness of incorporating semantic information and object attributes in generating captions of higher quality \cite{you2016image, wu2016value, gan2017semantic, yao2017boosting, zhou2016image}. Some research works on image captioning have used graphs to incorporate the spatial and semantic relations between the elements inside an image. In order to utilize graphs in caption generation, two types of graph extraction are usually used: scene graph extraction from images \cite{xu2017scene, yang2018graph, zhang2017visual, li2017scene, tang2019learning, gu2019scene, dai2017detecting} and scene graph extraction from textual data \cite{wang2018scene, anderson2016spice}.
Once a scene is abstracted into symbols, the language generation is almost independent of visual perception.\cite{yang2019auto} Given scene abstractions "helmet-on-human" and "road dirty," humans can infer "a man with a helmet in the countryside" by using common sense knowledge like "countryside road dirty." This can be considered as the inductive bias that enables humans to perform better than machines.

Yang et al. \cite{yang2019auto} have integrated the inductive bias of language generation into the encoder-decoder framework commonly used in image captioning. The proposed method uses scene graphs to connect the image and text modalities. A scene graph $G$ is a unified representation that connects the objects, their attributes, and their relationships in an image $I$ or a sentence $S$ by directed edges. To encode the language prior, Yang et al. \cite{yang2019auto} proposed the Scene Graph Auto-Encoder (SGAE), which is a sentence self-reconstruction network used in the $I \rightarrow G \rightarrow D \rightarrow S$ training pipeline. The $I \rightarrow G$ module is a visual scene graph detector. A multi-modal GCN is introduced and used in the $G \rightarrow D$ module to complement the visual cues that may be ignored due to imperfect visual detection. $D$ can be considered as a working memory \cite{vinyals2016matching} that assists in re-keying the encoded nodes from $I$ to $S$ to a more generic representation with smaller domain gaps. The proposed SGAE-based image captioning model is implemented using Faster R-CNN \cite{ren2015faster}, and the language decoder proposed by \cite{anderson2018bottom} with RL-based training strategy \cite{rennie2017self}. The proposed framework is formulated as follows:

  \begin{equation} \label{eq_autoencoding}
 \begin{gathered}
     \bm{Encoder:}  V \leftarrow I,  \\
     \bm{Map}: \hat{V} \leftarrow R(V,G;D), G \leftarrow V,\\
     \bm{Decoder:} S \leftarrow \hat{V}.
 \end{gathered}
 \end{equation}
 Where $V$ denotes the extracted image features (usually extracted by a Convolutional Neural Network (CNN)). The mapping module frequently used in the encoder-decoder framework for image captioning is the module that encodes the visual features from the image into a representation which is later taken as input by the language decoder. This mapping module has been modified according to the formulation in \ref{eq_autoencoding} by introducing the scene graph $G$ into a re-encoder $R$ parametrized by a shared dictionary $D$. The Scene Graph Auto-Encoder (SGAE) learns the dictionary $D$, which embeds the language inductive bias from sentence-to-sentence reconstruction. Next, the encoder-decoder framework is equipped with SGAE to form the overall image captioner.

Gu et al. \cite{gu2019unpaired} have introduced a particular framework for training an image captioning model in an unsupervised manner and without image-caption pairs. The framework uses a scene graph to generate an intermediate representation of images and captions and maps these scene graphs to their feature space using "Cycle-Consistent Adversarial Training" \cite{zhu2017unpaired}. This paper has used an image scene graph generator, a sentence scene graph generator, and a feature mapping module in charge of mapping image features and captions modalities together. To align scene graphs and captions, CycleGAN \cite{zhu2017unpaired} is used. The unrelated image and sentence scene graphs are first encoded using the scene graph encoder trained on the sentence corpus. Next, unsupervised cross-modal mapping is performed for feature alignment with CycleGAN. This work is closely related to \cite{yang2019auto}. The main difference is that the framework in \cite{yang2019auto} is based on paired settings. CycleGAN is generally used to transform two images together, and one of its applications is transforming two image elements together (For example, transforming an apple into an orange or a horse into a zebra.).

Gao et al. \cite{gao2018image} proposed a scene-graph-based semantic representation method by embedding the scene graph as an intermediate state. The task of image captioning is divided into two phases termed: concept cognition and sentence reconstruction. In the first phase, a vocabulary of semantic concepts is built, and a novel CNN-RNN-SVM framework is used to generate a scene-graph-based semantic representation, which is used as the input for an RNN generating captions in the second phase. The CNN part extracts visual features, the RNN part models image/concept relationships and concept dependency, and the SVM part classifies the semantic concepts and outputs the relevant concepts for the scene-graph-based sequence.

\begin{figure}[H]
  \centering
  \includegraphics[width=0.8\textwidth]{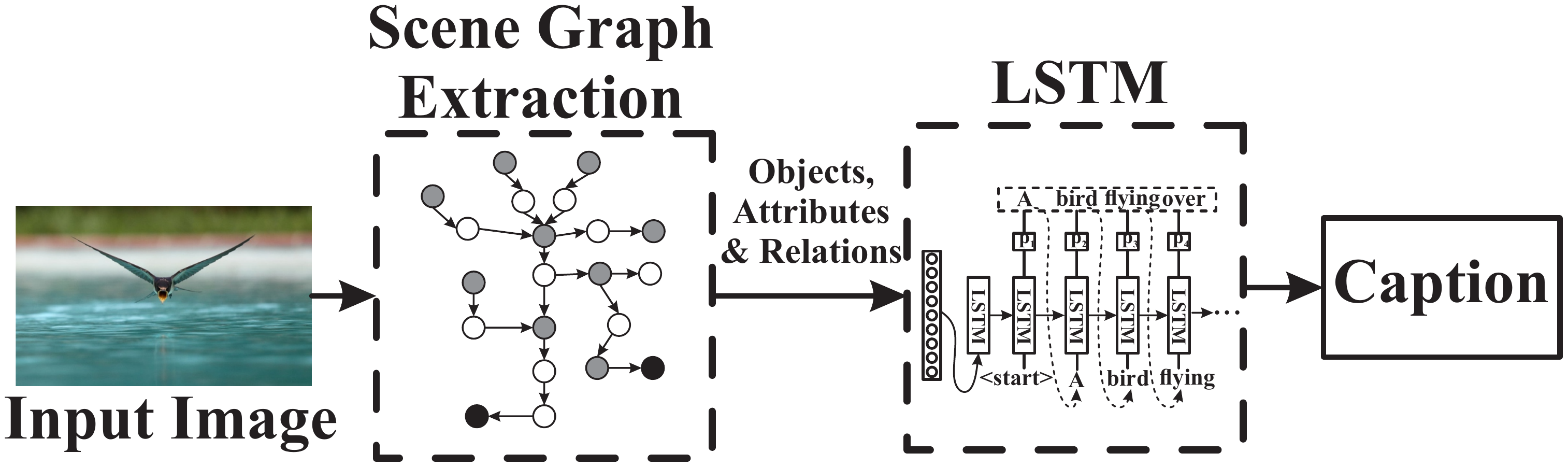}
  \caption{The general workflow of graph-based methods.}

  \label{fig_graph_based}
\end{figure}

The general workflow of the graph-based methods is displayed in Figure \ref{fig_graph_based}. Usually, a Convolutional Neural Network is used to extract visual features from the image, and the semantic and spatial graph is built on the detected regions. The vertices denote regions, and the edges denote the relationships between the regions. Next, Graph Convolutional Networks (GCNs) \cite{kipf2016semi} encode the regions and relationships in the scene graph. The obtained feature vector is then passed to LSTM \cite{hochreiter1997long} decoders to generate captions.

\subsection{Combining Attention-Based Methods and Graph-Based Methods} \label{section_combining_attention_graph}
In order to solve some of the issues revolving around image captioning problems and the problems regarding attention-based and graph-based methods, some recent research works have introduced structures that combine the two methodologies. 

As previously mentioned, the visual relations between image elements give insight into their relative positions or interactions. To detect the visual relations between image elements, one not only needs to detect object locations inside the image, but also needs to detect all sorts of interaction between pairs of elements. Using these visual relations will allow for a more in-depth comprehension of images. However, the considerable diversity in object sizes and their locations will make the interaction detection task more difficult. 

Yao et al. \cite{yao2018exploring} use a combination of Graph Convolutional Networks \cite{kipf2016semi} and LSTMs \cite{hochreiter1997long} to incorporate the relations between image elements while also taking the attention-based encoder-decoder framework into account. Spatial and semantic relations have been integrated to enrich image representations in the image encoder, and learning the relationships has been considered a classification problem. Faster R-CNN \cite{ren2015faster} has been used for region proposals. Two spatial and semantic graphs are built to represent spatial and semantic relations between image contents. These two graphs are generated from the detected image regions, with regions being graph nodes and the relations between them as the edges of the graph. In the spatial graph, spatial relations are considered as edges, and in the semantic graph, the semantic relations are considered as edges. The semantic graph is trained using the Visual Genome \cite{krishna2017visual} dataset. To represent the image, Graph Convolutional Networks \cite{kipf2016semi} are used which incorporate the semantic and spatial relations obtained from their corresponding graphs. The combination of the enhanced image region representations and their semantic and spatial relations are then fed into an LSTM \cite{hochreiter1997long} decoder to generate the caption sentences. During inference, to combine the output of the two spatial and semantic decoders, the distribution over the words generated by the two decoders is linear weight summated at each time step, and the word with the highest probability is extracted.

The proposed model by Zhong et al. \cite{zhong2020comprehensive} decomposes the image scene graph into a set of sub-graphs. Each sub-graph captures a semantic component of the input image. Zhong et al. \cite{zhong2020comprehensive} designed a sub-graph proposal network (sGPN) that learns to detect meaningful sub-graphs. An attention-based LSTM then decodes the selected sub-graphs for generating sentences. Given an input image $I$, a scene graph $G=(V,E)$ is extracted from $I$ using MotifNet \cite{zellers2018neural}, where $V$ represents the nodes corresponding to the detected objects in $I$ and $E$ represents the set of edges corresponding to the relationships between object pairs. The goal is to generate a set of sentences $S=\{S_j\}$ to describe different components of the image using the scene graph $G$. Sub-graphs are defined as $\{G_i^c=(V_i^c,E_i^c)\}$ where $V_i^c \subseteq V$ and $E_i^c\subseteq E$. The method aims to model the joint probability $P(C_{ij}=(G,G_i^c,C_j)|I)$, where $P(C_{ij}|I)=1$ when the sub-graph $G_i^c$ can be used to decode the sentence $S_j$ and $P(C_{ij}|I)=0$ otherwise. $P(C_{ij}|I)$ can be decomposed into three parts:

\begin{equation}
    P(C_{ij}|I)=P(G|I)P(G_i^c|G,I)P(S_j|G_i^c,G,I)
\end{equation}

$P(G|I)$ can be interpreted as the scene graph extraction phase, $P(G_i^c|G,I)$ as the scene graph decomposition phase and the selection of important sub-graphs for sentence generation, and $P(S_j|G_i^c,G,I)$ as the decoding phase in which a selected sub-graph $G_i^c$ is decoded into its corresponding sentence $S_j$, and the tokens in $S_j$ are associated to the nodes $V_i^c$ of the sub-graph $G_i^c$(the image regions in $I$).

Wang et al. \cite{wang2020learning} have used a Graph Neural Network \cite{scarselli2008graph} to represent the relation between image elements and have used a novel content-based attention framework to store image regions previously attended by the attention module as well. A ResNet-101 \cite{he2016deep} network trained on the ImageNet \cite{deng2009imagenet} dataset is first used to extract image features. The non-linear activations of the last convolutional layer of this network are used as the image representation and are denoted as:

\begin{equation}
V=\left\{v_{1}, v_{2}, \ldots, v_{n} \mid v_{i} \in \mathbb{R}^{m}\right\}
\end{equation}

Where $v_i$ represents each of the non-linear activations of the last convolutional layer. A Graph Convolutional Network \cite{kipf2016semi} $f_{gnn}$ is initialized using the image features belonging to different image regions to explore relations between the visual objects in the image. This graph initializes each node inside the graph with a spatial representation and to derive the implicit relation-aware representation $R=\left\{r_{1}, r_{2}, \ldots, r_{n} \mid r_{i} \in \mathbb{R}^{m}\right\}$ (where $r_i$ represents the nodes inside the graph), updates the value of the nodes with hidden representation from other nodes recursively. The visual representations $R$ are forwarded into context-aware attention model $f_{att}$. Unlike some other attention-based models, this novel attention framework uses LSTM \cite{hochreiter1997long} to store the previously attended regions. Storing these regions will aid the attention module in its future region selections. Next, a language model based on LSTM,  $f_{lstm}$, uses the previous hidden state $h_{(t-1)}$, the previously generated word embedding $X_t$ and the output $\bar{v}_{t}$ from the attention model as input and produces the current hidden state $h_t$ as the output to predict the next word.

Chen et al. \cite{chen2020say} have proposed a model to generate controllable image captions which actively consider user intentions. The paper introduces a more fine-grained control signal called Abstract Scene Graph (ASG), a directed graph composed of three types of abstract nodes grounded in the image: object, attribute, and relationship. The caption generation model is based on the encoder-decoder framework, consisting of a role-aware graph encoder and a language decoder that considers both the context and structure of nodes for attention. The decoder utilizes a two-layer LSTM \cite{hochreiter1997long} structure, including an attention LSTM and a language LSTM. The model gradually updates the graph representation during decoding to fully cover information in ASG without omission or repetition and keep track of graph access status. The role-aware graph encoder contains a role-aware node embedding to distinguish node intentions and a multi-relational Graph Convolutional Network for contextual encoding. 

Aiming to employ knowledge in scene graphs for image captioning explicitly, Li et al. \cite{li2019know} introduce a framework based on scene graphs. First, the scene graph for the input image is generated using the method proposed in \cite{xu2017scene}. A set of initial bounding boxes should be produced to generate the scene graph. Li et al. have used the region proposal network (RPN) proposed by Girshick et al. \cite{girshick2015fast} to produce a set of object proposals for the image. To capture the visual features, the VGG-16 network is used to extract CNN features from the corresponding regions of object entities. Semantic features are also obtained by extracting triplets, which are lexeme sequences that describe object relationships from the graph and embed them into fixed-length vectors. To utilize both types of information, a hierarchical attention-based fusion module is introduced which determines when and what to attend to during sentence generation.

Xu et al. \cite{xu2019scene} proposed a framework to embed the scene graph into a compact representation capable of capturing explicit semantic concepts and graph topology information. An input image $I$ is processed by a CNN to generate the image features. A set of modules detect the objects, attributes, and related components to infer the scene graph. Next, an external vocabulary compiles the scene graph into the vector $V_{con}(I)$. An adjacent matrix is presented where the objects and relationships of the graphs are used as vertices and edges. A fixed-length vector $V_{topo}(I)$ is extracted to capture the topological information from the adjacent matrix. Xu et al. proposed an attention extraction mechanism that extracts sub-graphs and selects an attention graph with the corresponding region by computing cluster nodes in the adjacency matrix. The attention region is denoted as $V_{att}(I)$. The four vectors are combined into a single representation for the scene graph, which is fed into the LSTM-based \cite{hochreiter1997long} language model. 

Lee et al. \cite{lee2019learning} have extended the top-down captioner introduced in \cite{anderson2018bottom} and have added an attention component for relation features. No graph convolutions are used in the proposed model. The authors state that using visual relations from scene graphs directly is an alternative to Graph Convolutional Networks (GCNs), and avoids expensive graph convolutions.

There is a different set of challenges associated with the use of scene-graphs. Scene graph extraction is a difficult task on its own, and the relations between the elements are not always as simple as pairwise relationships. Graph parsers still need improvement as well.

\subsection{Convolutional Network-Based Methods}\label{convolutional}
Convolutional network-based methods utilize convolutional neural networks to extract image features and generate captions using output from a language model.
Thanks to the recent advances in convolutional architectures on other sequence-to-sequence tasks such as convolutional image generation \cite{oord2016conditional} and machine translation \cite{gehring2016convolutional, gehring2017convolutional}, it is possible to consider CNNs as an effective solution to many vision-language tasks. The methods discussed in this section have incorporated CNNs into their proposed systems. 

Inspired by the advances of CNNs in vision-language tasks, Aneja et al. \cite{aneja2018convolutional} have presented a convolutional model containing three main components. The first and the last components are input/output word embeddings, respectively. While the middle component contains LSTM \cite{hochreiter1997long} or GRU \cite{cho2014learning} units in other methods, masked convolutions are used in the proposed approach. This component is feed-forward without any recurrent functions, unlike the RNN approaches. 

Wang et al. \cite{wang2018cnn} proposed a framework relying on Convolutional Neural Networks only to generate captions. 
The framework consists of four modules: a vision module, a language module, an attention module, and a prediction module. The vision module is a CNN without the fully connected layer, for which VGG-16 has been used. The language module is based on a CNN without pooling. RNNs use a recurrent path to memorize context, whereas CNNs use kernels and stack multiple layers to model the context. The prediction module is a one-hidden layer neural network as well. Since different levels of the language CNN represent different levels of concept, a hierarchical attention module has been employed where attention vectors are calculated at each level of the language model and fed into the next level. Since the attention maps are computed in a bottom-up manner as opposed to the RNN-based model, it is possible to train the model in parallel over all words in the sentence. The authors observed the effect of several hyper-parameters, such as the number of layers and the width of the kernel belonging to the language CNN. The receptive field of the language CNN can be increased by stacking more layers or increasing the width of the kernel. The experiments showed that increasing the kernel width is a better choice.

Less attention has been paid to convolutional network-based methods compared to the categories discussed above. Convolutional network-based models help generate more entropy and as a result, more caption diversity. Also, they perform better in classification tasks and do not suffer from vanishing gradients. However, these methods still need improvement in terms of performance according to the evaluation metrics.
\subsection{Transformer-Based Methods}
Many current works have utilized Transformers to build more robust solutions for the captioning problem. 
RNNs and LSTMs have been criticized due to their inflexibility, limitations regarding expression ability, and other complexities. Due to their recurrent nature, RNNs have difficulty memorizing inputs many steps ago, which leads to high-frequency phrase fragments without regard to the visual cues \cite{li2019entangled}. 
The limitations posed by LSTMs and RNNs as language models have led researchers to use alternatives such as Transformers. 

Some recent works have studied the application of Transformers \cite{vaswani2017attention} -mainly as the language model. Herdade et al. \cite{herdade2019image} utilized Transformers in the proposed "Object Relation Transformer" model, which incorporates spatial relations between detected objects using geometric attention. This encoder-decoder-based structure implements spatial relationships between detected objects inside an image using geometric attention. The object relation module presented by Hu et al. \cite{hu2018relation} represents the spatial relations in the encoder. The combination of Faster R-CNN \cite{ren2015faster}, and ResNet-101 \cite{he2016deep} as the base Convolutional Neural Network is used for object detection and feature extraction. Every image feature vector is processed through an input embedding layer consisting of a fully connected layer to reduce the dimension, followed by a ReLU and a dropout layer. The first encoder layer of the Transformer model uses the embedded feature vectors as input, and the subsequent layers use the output tokens of the previous encoder layers. Each encoder layer is composed of a multi-head self-attention layer followed by a small feed-forward neural network. 




Using the intermediate feature maps obtained from ResNet-101 \cite{he2016deep} as input, a "Region Proposal Network (RPN)" generates bounding boxes for the objects proposed by the network. Multiple neural network layers are added to predict the corresponding class for each region and correct the bounding box for each of the proposed regions. Also, to implement geometric attention, the value of attention weight matrices changes: bounding box properties (such as center, width, and height) are combined with their corresponding attention weights using a high-dimensional embedding \cite{vaswani2017attention}. 



Liu et al. \cite{liu2020exploring} 
introduce the "Global-and-Local Information Exploring-and-Distilling (GLIED)" approach that explores and distills the cross-modal source information. The Transformer-based structure globally captures the inherent spatial and relational groupings of the individual image regions and attribute words for an aspect-based image representation. Afterward, it extracts fine-grained source information locally for precise and accurate word selection. They used the RCNN-based visual features provided by Anderson et al. \cite{anderson2018bottom} for image regions extracted by Faster R-CNN \cite{ren2015faster}. 




Cornia et al. \cite{cornia2020meshed} introduce a fully attentive model called $M^2$- a Meshed Transformer with Memory for Image Captioning. The architecture is inspired by the Transformer model for machine translation and learns a multi-level representation of the relationships between image regions integrating learned a priori knowledge. The model incorporates two novelties: (1) image regions and their relationships are encoded in a multi-level fashion, in which both low-level and high-level relations are considered. The model learns and encodes a priori knowledge using persistent memory vectors. (2) The sentence generation -done with a multi-layer architecture- exploits both low- and high-level visual relationships via a learned gating mechanism, which weights multi-level contributions at each stage. This creates a mesh-like connection between the encoder and decoder layers. The encoder is in charge of processing regions in the input image and the relationships between them. Simultaneously, the decoder reads the output of each encoding layer and generates the caption word by word. All interactions between word and image-level features are modeled via scaled dot-product attention without using recurrence.

Huang et al. \cite{huang2019attention} used a Transformer-like encoder paired with an LSTM decoder. Li et al. \cite{li2019entangled} investigated a Transformer-based sequence modeling framework named "ETA-Transformer ."They have proposed EnTangled Attention (ETA) that enables the Transformer to benefit from both semantic and visual information simultaneously. Liu et al. \cite{liu2021cptr} introduce CaPtion TransformeR (CPTR), which takes sequentialized raw images as input to the Transformer. As an encoder-decoder framework, CPRT is a full Transformer network that replaces the commonly used CNN in the encoder part with the Transformer encoder. A purely Transformer-based architecture, PureT, is designed by Wang et al. \cite{wang2022end}. In PureT, SwinTransformer \cite{liu2021swin} replaces Faster-RCNN, and the architecture features a refining encoder and decoder. 
Fang et al. \cite{fang2022injecting} introduce a fully VIsion Transformer-based image CAPtioning mode (ViTCAP) along with a lightweight Concept Token Network (CTN), which is used to produce concept tokens. The structure uses a vision transformer backbone as the stem image encoder, which produces grid features. CTN is then applied to predict semantic concepts. A multi-modal module uses grid representations and Top-K concept tokens as input to perform the decoding process. Pseudo ground-truth concepts are extracted from the image captions using a simple classification task, and CTN is optimized to predict them during training. Li et al. \cite{li2022comprehending} designed a Transformer-style encoder-decoder structure called Comprehending and Ordering Semantics Networks (COS-Net). A CLIP model (image encoder and text encoder) \cite{radford2021learning} is used as a cross-modal retrieval model which retrieves sentences semantically similar to the input image. The semantic words in retrieved sentences are treated as the primary semantic cues. A novel semantic \textit{comprehender} is also introduced by the authors, which removes the irrelevant semantic words in those primary cues and simultaneously infers missing words visually grounded in the image. Afterward, a semantic ranker sorts the semantic words in linguistic order. Zeng et al. \cite{zeng2022s2} propose a Spatial-aware Pseudo-supervised (SP) module which uses a number of learnable semantic clusters to quantize grid features with multiple centroids without direct supervision. These centroids aim to integrate grid features of similar semantic information together. In addition to the SP module, a simple weighted residual connection is introduced, named Scale-wise Reinforcement (SR) module. This module explores both low and high-level encoded features concurrently. 

Nguyen et al. \cite{nguyen2022grit} present a Transformer-only neural architecture titled GRIT (Grid- and Region-based Image captioning Transformer) which uses DETR-based detector along with grid and region based features.
Hu et al. \cite{hu2022expansionnet} have proposed ExpansionNet v2 which utilizes a novel technique titled Block Static Expansion layer. This technique processes the input by distributing it over a collection of sequences with different lengths, which helps to explore the possibility of performance bottlenecks in the input length in Deep Learning methods. This layer is designed to improve the quality of features refinement and ultimately increase the effectiveness of the static expansion. The architecture of ExpansionNet v2 follows the standard encoder-decoder structure and is implemented on top of Swin-Transformer \cite{liu2021swin}.

\subsection{Combining Transformers and Scene Graphs}
A number of the works have experimented with model designs that incorporate both Transformers and scene graphs.

He et al. \cite{he2020image} aimed to employ the spatial relations between detected regions inside an image. In their proposed model, each Transformer layer implements multiple sub-transformers to encode relations between regions and decode information. The encoding method combines a visual semantic graph and a spatial graph. In another architecture introduced by Chen et al. \cite{chen2021captioning}, the encoder consists of two sub-encoders for visual and semantic information. Faster-RCNN proposes image regions, and a scene graph is built using the detected regions. GCN is then used to enrich the graph representation. A semantic matrix is learned from the scene graph and fed into a multi-modal attention module in the decoder. This module is used to leverage multi-modal representation in caption generation. Yang et al. \cite{yang2022reformer} have proposed an architecture called ReFormer, which generates features with relation information embedded. ReFormer explicitly expresses the pair-wise relationships between objects present inside an image. ReFormer combines scene graph generation and image captioning using one modified Transformer model.

\subsection{Vision Language Pre-Training Methods for Image Captioning}
Some recent works have attempted pre-training paradigms to lessen the reliance of the models on fully-supervised learning. A large-scale model is pre-trained on a dataset with an enormous amount of data by self-supervised learning. The pre-trained model is then generalized to various downstream tasks. 

One widely used pre-trained model is CLIP (Contrastive Language-Image Pre-Training). CLIP is designed to provide a shared representation for both image and text prompts \cite{mokady2021clipcap}. It has been trained on numerous images and captions using a contrastive loss, allowing for more consistency and correlation between its visual and textual representations. One of the recent works utilizing CLIP in the proposed method is ClipCap by Mokady et al. \cite{mokady2021clipcap}. The authors introduce a model that produces a prefix for each caption by applying a mapping network over the CLIP embeddings. Next, a pre-trained language model (GPT-2 \cite{radford2019language}) is fine-tuned to generate captions. This approach is inspired by Li et al. \cite{li2021prefix}, who discussed the possibility of adapting a language model for new tasks by concatenating a learned prefix. Barraco et al. \cite{barraco2022unreasonable} investigate the role of CLIP features in image captioning by devising an architecture composed of an encoder-decoder Transformer architecture. Hu et al. \cite{hu2021vivo} present the Visual VOcabulary pre-training (VIVO), which aims to learn a joint presentation of visual and text input. Unlike existing VLP models, which use image-caption pairs to pre-train, VIVO uses image-tag pairs for pre-training. In the pre-training stage, an image captioning model first learns to label image regions using image-tag pairs as training data. In the fine-tuning stage, the model learns to map an image to a sentence conditioned on the detected objects using image-caption pairs and their corresponding object tags. The sentences are learned from image-caption pairs, while object tags may refer to novel objects that do not exist in image-caption pairs. The addition of object tags allows for zero-shot generalization to novel visual objects for image captioning. Xia et al. \cite{xia2021xgpt} highlight that while recent pre-training methods for vision-language (VL) understanding tasks have achieved state-of-the-art performance, they cannot be directly applied to generation tasks. Xia et al. present Cross-modal Generative Pre-Training for Image Captioning (XGPT), which uses a cross-modal encoder-decoder architecture and is directly optimized for generation tasks. 

Li et al. \cite{li2020oscar} have proposed a pre-training method that leverages salient objects, which are usually present in both image and caption as anchor points. The method uses object tags as anchor points to align image and language modalities in a shared semantic space. The training samples are defined as triplets, each consisting of a word sequence, a set of object tags, and a set of image region features. This pre-training method can be applied to many vision-language tasks, including image-text retrieval, Visual Question Answering (VQA), and image captioning.
Many vision-language pre-training methods, including \cite{li2020oscar}, are built upon Bidirectional Encoder Representations from Transformers (BERT) \cite{devlin2018bert}. These models use a two-stage training scheme in which the model first learns the contextualized vision-language representations by predicting the masked words or image regions based on their intra-modality or cross-modality relationships on large amounts of image-text pairs. To counteract the problem of pre-training a single, unified model that is applicable to a wide range of vision-language tasks via fine-tuning, Zhou et al. \cite{zhou2020unified} have introduced a new pre-training method for a unified representation for both encoding and decoding. The unified encoder-decoder model, called the Vision-Language Pre-training (VLP) model, can be fine-tuned for both vision-language generation (e.g., image captioning) and understanding tasks (e.g., visual question answering). This model uses a shared multi-layer Transformer network for encoding and decoding, which is pre-trained on large amounts of image-caption pairs. The VLP model is optimized for two unsupervised vision-language prediction tasks: bidirectional and sequence-to-sequence (seq2seq) masked language prediction. These two tasks only differ in what context the prediction conditions are on, which is controlled by specific self-attention masks for the shared Transformer network. The context of the masked caption word, which is the target of prediction, consists of all the image regions and all words on its right and left in the caption in bidirectional prediction. In contrast, in the seq2seq task, the context consists of all the image regions and the words on the left of the to-be-predicted word in the caption. 

Li et al. \cite{li2022mplug} present mPLUG; a novel vision-language foundation model designed for both cross-modal understanding and generation. mPLUG aims to counteract some of the problems commonly witnessed in pre-training models, such as low computational efficiency and information asymmetry by novel cross-modal skip-connections. These skip-connections generate inter-layer shortcuts that skip a specific number of layers. This method is used to improve the slow full self-attention on the vision side.
Liu et al. \cite{liu2023prismer} present Prismer, a vision-language model that uses a group of domain experts to combine their knowledge and apply it to different vision-language reasoning tasks. Prismer performs well in fine-tuned and few-shot learning, while requiring significantly less training data compared to other models.
\subsection{Unsupervised Methods and Reinforcement Learning}
There has been a recent trend toward relaxing the reliance on paired image-caption datasets for image captioning. Many of the current research works employ reinforcement learning methods due to their unsupervised nature. The interest in unsupervised methods stem from the problem of the models relying almost entirely on the quality and volume of the image-caption pairs in datasets.

One early work by Gu et al. \cite{gu2018unpaired} involved generating captions in a pivot language and translating the caption to a target language. This method requires a paired image-caption dataset for the pivot language but does not use a paired dataset with captions being in the target language. Another research paper in this field used reinforcement learning with gradient policy along with RNNs in 2016 \cite{ranzato2015sequence}. Shetty et al. \cite{shetty2017speaking} proposed the first study that explored using conditional GANs \cite{mirza2014conditional} to generate human-like and diverse descriptions. 

Feng et al. \cite{feng2019unsupervised} use a set of images, a sentence corpus, and a visual concept detector for unsupervised training. The images and the sentence corpus are projected into a common latent space such that they can reconstruct each other. The sentence corpus is prepared using the captions available on Shutterstock \cite{shutterstock}, which is a photo-sharing platform. On this platform, each image is uploaded with a caption. This corpus is not related to the images and is independent. The proposed structure comprises an image encoder, a sentence generator, and a discriminator. The Inception-V4 \cite{szegedy2017inception} is used as the image encoder, and the sentence generator and discriminator are both LSTMs \cite{hochreiter1997long}.
Since no image-caption pairs exist, three new metrics have been introduced as three discriminators to evaluate the model's performance. The discriminator first distinguishes a real sentence from the sentence corpus from a sentence generated by the model, and the generator is rewarded at each time step. By maximizing this reward, the generator tries to produce plausible sentences. However, more than this discriminator is needed since the quality of the generated sentence, and its relevance to the image must also be evaluated. To do so, the model must learn the visual contents of the image. The generated words are rewarded if the generated caption contains words whose corresponding visual concept is detected inside the image. This reward is called a "concept reward." Finally, since the performance of the model is much dependent on the performance of the visual concept detector and these detectors only detect a limited number of objects, images and captions are projected into a common latent space such that they can reconstruct each other.

Chen et al. \cite{chen2019improving} proposed an image captioning framework based on conditional generative adversarial nets as an extension of the reinforcement learning-based encoder-decoder architecture. Highlighting that the conventional encoder-decoder structures directly optimize one metric, which cannot guarantee improvement in all metrics, the paper designed a discriminator network to decide if a caption is human-described or machine-generated based on the idea of GANs. Two discriminator models have been designed and tested: a CNN-based discriminator model that uses the conditional CNN for real or fake sentence classification, and an RNN-based discriminator model that consists of the standard LSTM \cite{hochreiter1997long}, a fully connected linear layer, and a softmax output layer. The CNN-based framework was shown to improve the performance more than the RNN-based framework, while the RNN-based framework can save 30\% training time. It was finally concluded that the ensemble results of 4 CNN-based (denoted as CNN-GAN) and 4 RNN-based (denoted as RNN-GAN) models could noticeably improve the performance of a single model.

Liu et al. \cite{liu2018show} have introduced an image captioning module and a self-retrieval module. A Convolutional Neural Network extracts image features, and an LSTM \cite{hochreiter1997long} decodes a sequence of words based on these features. 
The self-retrieval module evaluates the similarity between the generated captions, the input image, and some "distractors." If the caption generator module generates distinct and proper captions, the relevance between these captions and their corresponding images must be more than the relevance between the generated captions and unrelated, distracting images. This condition is represented as the text-to-image retrieval error and improves the performance of the image captioning module with back-propagation and the REINFORCE algorithm. 

Guo et al. \cite{guo2019mscap} used a discriminator structure similar to that of \cite{feng2019unsupervised}. The discriminator distinguishes whether the generated sentence is real and rewards the learner based on how real the sentences seem. Another discriminator distinguishes the style of the generated captions. Also, the LSTM \cite{hochreiter1997long} decoder used in \cite{gu2018stack} has been used as a reinforcement learning agent making an action (prediction of the next word). After a sentence is completed, the agent will observe a sentence-level reward and update its internal state.

\begin{figure}[H]
  \centering
  \includegraphics[width=0.95\textwidth]{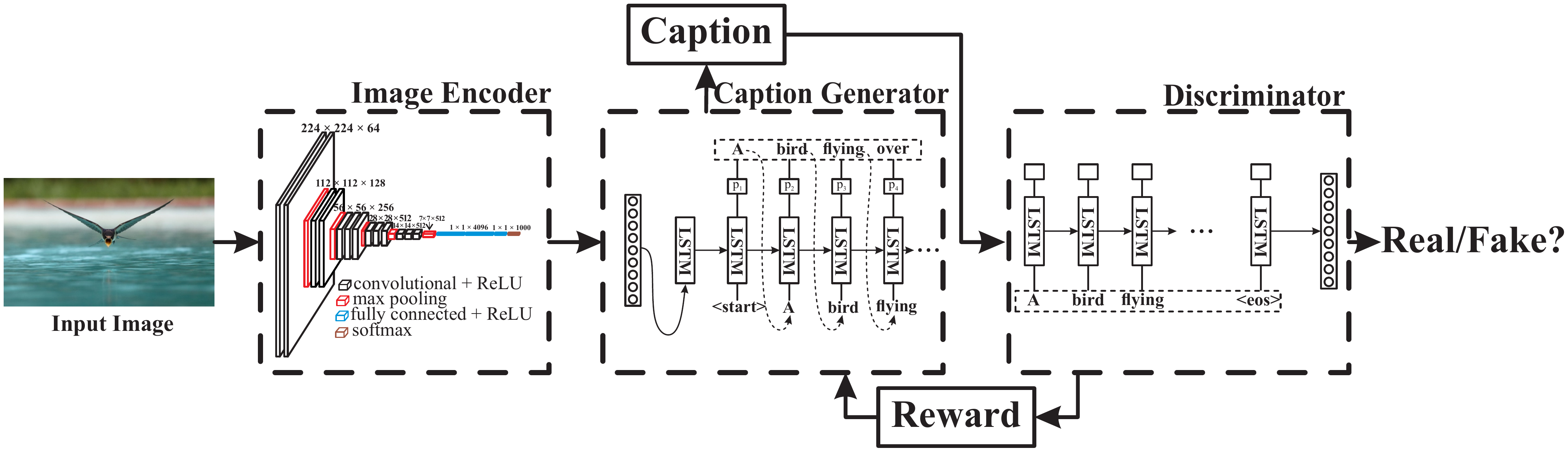}
  \caption{The general workflow of unsupervised methods.}
  
  \label{fig_unsupervised}
\end{figure}

A block diagram of the general workflow of the unsupervised methods is shown in Figure \ref{fig_unsupervised}. VGGNet \cite{simonyan2014very} has been used as the image encoder, and the caption generator is an LSTM network \cite{hochreiter1997long}. Therefore, the overall design follows the typical encoder-decoder structure. The discriminator is also an LSTM network, which determines if the given caption is real (from the sentence corpus) or generated by the model. The generator is rewarded accordingly by the discriminator.

Taking the issues related to supervised settings into account, such as the tedious process of dataset preparation and the difficult training process, the unsupervised setting has been the focus of many recent works and is expected to become more favored in the future as well.

\subsection{Generating Multi-Style Captions}\label{section_multistyle}
The papers discussed so far generate captions with a neutral tone. These generated captions usually describe factual data about image contents. Meanwhile, humans use many styles and tones in their daily speech to communicate with one another. Some of these styles and tones are humorous, hostile, and poetic. Incorporating these styles can help humans interact with the caption more and make the captions more attractive. Stylized captions can also be used in applications such as photo-sharing and Chatbots. 

Shuster et al. \cite{shuster2019engaging} have added tone and style as a feature to their dataset, as well as images and their appropriate captions. This paper has introduced a novel structure called TransResNet, which projects images, captions, and their corresponding personality traits into a shared space using an encoder-decoder framework. Two classes of models have been considered: retrieval models and generative models. The retrieval model considers any caption in the entire dataset as a possible candidate response, whereas the generative model produces captions word by word via the aforementioned structure. The retrieval model has given better results.

A structure consisting of five modules for caption generation in different styles has been introduced \cite{guo2019mscap} by Guo et al. . The first module is a plain image encoder. Next is a caption generator module that outputs a sentence conditioned on a specific style. The following module is a caption discriminator that distinguishes a real sentence from a generated sentence. This discriminator is trained in an adversarial manner to encourage the learner to generate more convincing captions closer to the human language. Afterward, a style discriminator module that determines the style of the generated caption is used. Inspired by the fact that there is some content consistency between neutral captions and stylized captions, another module called "The Back-Translation Module" is also used. This module translates a stylized caption into a neutral one. (If a stylized caption is generated and translated to a factual and neutral caption, we should arrive at the real factual caption.) This process is implemented using multi-lingual neural machine translation (NMT), in which the stylized captions are considered input and neutral captions are considered output.

\begin{figure}[H]
  \centering
  \includegraphics[width=0.99\textwidth]{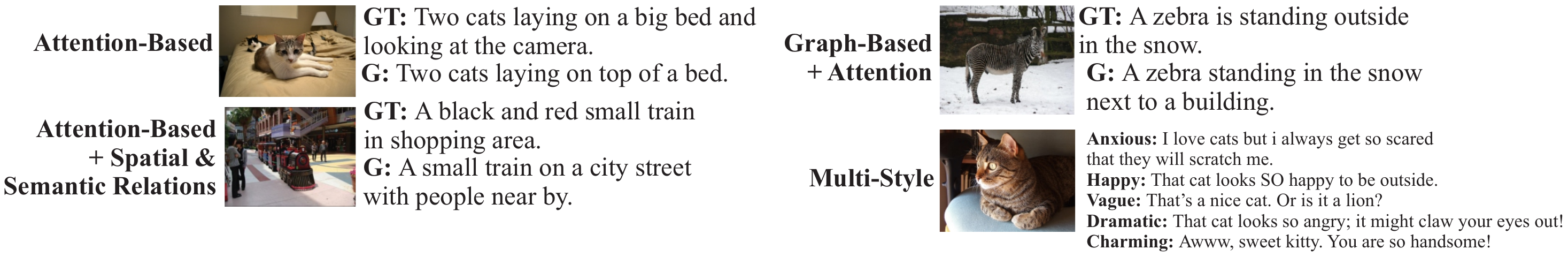}
  \caption{Sample captions generated by multiple methods in different categories. "GT" indicates "Ground Truth Caption," and "G" indicates "Generated Caption." The captions are generated by Huang et al. \cite{huang2019attention} (top-left), Wang et al. \cite{wang2020learning} (top-right), Li et al. \cite{li2020oscar} (bottom-left), and Shuster et al. \cite{shuster2019engaging} (bottom-right)}
  
  \label{fig_sample_all}
\end{figure}

A figure consisting of some example captions from sections \ref{section_attention}, \ref{section_attention_spatial_semantic}, \ref{section_combining_attention_graph}, and \ref{section_multistyle} in this survey is shown in Figure \ref{fig_sample_all}. Each row belongs to a specific category in which two images are displayed, along with the captions describing them. For each image, the ground-truth caption and the generated caption are shown.

\section{Problems in Image Captioning}
In image captioning, researchers are usually confronted with a set of problems, some of which commonly experienced in many artificial intelligence tasks such as the exposure bias problem \cite{ranzato2015sequence}, the loss-evaluation mismatch problem \cite{gu2018stack, liu2017improved, rennie2017self, xu2019multi}, the vanishing gradient problem \cite{hochreiter2001gradient}, and the exploding gradient problem \cite{pascanu2013difficulty,goodfellow2017deep}. In addition, image captioning poses certain challenges unique to the task. These challenges include object hallucination, illumination conditions, contextual understanding and referring expressions.
In this section, we first review some of the common problems in many intelligent tasks, followed by continuing problems in image captioning which come, in fact, a part of of the nature of the task.

\subsection{The Exposure Bias Problem}
This problem occurs in language models when the model is only exposed to the training data and not its own predictions. The standard RNN models are trained to predict the next word according to the previous words in the ground truth sequence. In contrast, in testing time, the ground truth data is no longer available, and the model uses its own previous predictions for its following predictions. This problem will gradually produce more errors in the model’s output \cite{ranzato2015sequence}.
The structure proposed by Gu et al. \cite{gu2018stack} has employed reinforcement learning for optimization. This structure uses each intermediate decoder’s output in testing time and also the output of the previous decoder to normalize the reward, which solves the exposure bias and the loss-evaluation mismatch problem at the same time.

\subsection{The Loss-Evaluation Mismatch Problem} The language models are usually trained to minimize the cross-entropy loss at each time step. Meanwhile, during testing, the generated captions are evaluated using sentence-level metrics (discussed in "Evaluation Metrics for Image Captioning Methods"). These metrics are non-differentiable and cannot be used directly as a test-time error \cite{gu2018stack}. Multiple efforts have been made to optimize these metrics using reinforcement learning \cite{liu2017improved, rennie2017self, xu2019multi}.

\subsection{The Vanishing Gradient Problem}
The vanishing gradient problem happens in neural networks that train with gradient methods and back-propagation. In these methods, each of the neural network's weights is updated according to the partial derivative of the error function based on the current value inside the weight at each iteration. In some cases, the value of the gradient is so minuscule that there is no change in weights. In the worst case, this might halt the training process completely \cite{hochreiter2001gradient}.

\subsection{The Exploding Gradient Problem}
Another problem associated with gradients is the "exploding gradient" problem. In deep or recurrent neural networks, error gradients can build up and accumulate during an update, resulting in enormous values of gradients. Consequently, the network's weights receive large updates, and as a result, the network will become very unstable. In the best case, the deep multilayer Perceptron network cannot learn from the training data and results in NaN (Not a Number) weight values that can no longer be updated. Also, in recurrent networks, exploding gradients can result in an unstable network that is unable to learn from training data. In the best case, it will result in a network that cannot learn over long input sequences of data \cite{pascanu2013difficulty,goodfellow2017deep}.

\subsection{Object Hallucination}
Object hallucination \cite{rohrbach2018object} is a persistent problem for image captioning models, wherein the model detects objects that are not present in the input image. This can lead to poor performance in visually-impaired users, who require accurate and concise captions. 
According to a study by MacLeod et al. \cite{10.1145/3025453.3025814}, for many visually impaired people who prefer correctness over coverage, hallucination is a severe disadvantage for a captioning model and an obvious concern. Furthermore, object hallucination indicates an internal issue of the model. Rohrbach et al. have proposed a new metric to measure object hallucination, CHAIR (Caption Hallucination Assessment with Image Relevance), which measures the proportion of generated words that correspond to objects in the input image, according to ground truth sentences and object segmentations. The CHAIR metric has both per-instance and per-sentence variants, denoted as $CHAIR_i$ (equation \ref{eq_chair_i}) and $CHAIR_s$ (equation \ref{eq_chair_s}), respectively. 

\begin{equation}\label{eq_chair_i}
    CHAIR_i=\frac{|\text{Hallucinated  objects}|}{|\text{All  objects  mentioned}|} 
\end{equation}

\begin{equation}\label{eq_chair_s}
    CHAIR_s=\frac{|\text{Sentences with hallucinated objects}|}{|\text{All sentences}|} 
\end{equation}
According to the study performed by Rohrbach et al., models that perform better on standard evaluation metrics (Such as BLEU \cite{papineni2002bleu} and SPICE \cite{anderson2016spice}) perform better on CHAIR. However, this is not always true. It was found that the models which were optimized for CIDEr frequently hallucinated more. Also, models with attention tended to perform better on the CHAIR metric than models that did not incorporate attention. However, this gain was primarily due to these models' access to the underlying convolutional features and not the actual attention mechanism. Also, GAN-based models decreased hallucination, implying that GAN loss is beneficial in decreasing hallucination. This is due to the fact that the GAN loss encourages sentences to resemble human-generated captions. The presence of a hallucinated object likely suggests that a sentence is generated, and the discriminator dismisses the caption containing the hallucinated object.
\subsection{Illumination Conditions} \label{problem_illumination_conditions} 
Illumination conditions are a critical factor that can impact the accuracy and reliability of the generated captions, particularly when the image is captured in low-light conditions or indoors. Poor lighting can result in images with reduced contrast, making it difficult for the captioning model to discern fine details and recognize objects, people, or scenes. Moreover, the presence of shadows or uneven illumination can further hinder the model's ability to accurately analyze the visual content. Shadows and uneven illumination can also further complicate the model's analysis of visual content. For example, an image of a black cat in a dimly lit room with uneven illumination may be difficult for the captioning model to recognize as a cat. To overcome these challenges, researchers have been actively exploring various techniques to improve the visual quality of the images \cite{al2022comparing, al2022low}, including contrast enhancement, color correction, and low-light image enhancement. These techniques aim to mitigate the challenges posed by poor illumination and improve the accuracy of generated captions.

\subsection{Contextual Understanding}\label{problem_contextual_understanding} 
Image captioning models also require the ability to understand the context of the scene, including the relationships between objects, the spatial arrangement, and the overall atmosphere \cite{you2016image, wu2016value, gan2017semantic, yao2017boosting, zhou2016image}. This contextual understanding can be difficult to achieve, as it requires the caption generation model to have a deep understanding of the visual content and the ability to perform reasoning given the visual content.

\subsection{Referring Expressions}\label{problem_referring_expressions} 
Another problem in image captioning is the use of referring expressions, such as "the girl with the red hair" or "the dog in the corner." These expressions require the captioning models to identify and link the appropriate objects in the image. This can pose a challenge, especially if the objects are partially obscured or if there are multiple similar objects in the scene, and requires a combination of visual and linguistic understanding \cite{coppock2020informativity}. Referring expressions are important for improving caption quality, as they provide more detailed and informative descriptions of the objects in the image, allowing the model to generate more accurate and nuanced captions.

\section{Discussion}\label{discussion_section}
This section provides a comprehensive critical analysis of the methods falling in the different categories overviewed in section \ref{deep_learning_based_image_captioning}. Each method-inevitably-possesses advantages and disadvantages. Nevertheless, considering these characteristics aids researchers in adopting a suitable solution. The technical details of the structures and methods discussed in this section have been explained in section \ref{deep_learning_based_image_captioning}.
\subsection{Using Attention}
Attention-based methods attempt to imitate the human attention mechanism by showing the model "where to look at" during the training process. Attention is widely used in encoder-decoder architectures where CNNs are typically used in combination with LSTMs to produce a representation for the given image and generate captions, respectively. 
Some of the papers focusing on attention-based methods have mentioned low precision in region selection for attention as a flaw of the attention-based methods. They claim that most of the attention-based methods presented choose regions of the same size and shape without considering image contents. They have also mentioned that determining the optimal number of region proposals will bring about an unresolvable trade-off between small or large amounts of detail (or, representing the image coarsely or finely). One solution to this problem was proposed by Anderson et al. in \cite{anderson2018bottom} as the "bottom-up and top-down" method. Another problem of the attention-based methods is the "single-stage" structure. Most of these methods are only a single encoder-decoder attention structure, which cannot provide rich captions for the images. In the multi-stage coarse-to-fine structure proposed by Gu et al. \cite{gu2018stack}, at each stage, attention weights and hidden vectors generated by the previous stage decoder are used as input to the next stage decoder, reducing ambiguity in the captions. This structure allows for a richer caption at each stage. Another problem associated with attention-based methods for image captioning is that a proper correlation between the vectors obtained from attention and caption is not guaranteed, and it might lead to improper results. If feature vectors do not contain valuable information, the attention model still generates a vector that is a weighted sum over candidate vectors and is unrelated to the correct caption. To solve this issue, Huang et al. \cite{huang2019attention} have introduced an attention-based structure (Attention on Attention- or AoA) which contains one more level of attention. The authors have compared AoA with LSTM \cite{hochreiter1997long} and GRU networks \cite{cho2014learning}: internal states, memories, and gates are used in LSTMs and GRUs to implement the attention mechanism. AoA only performs two linear transformations and does not require hidden states, making it computationally reasonable while outperforming LSTM. The combination of LSTM and AoA has been reported to be unstable since it can reach a sub-optimal point. This means that increasing the volume of the stack and the number of gates to improve the performance is futile. Jiang et al. \cite{jiang2018recurrent} state that the existing encoder-decoder models employ only one kind of CNN to describe image content. Consequently, the image contents will be described from only one specific viewpoint, and the semantic meaning of the input image cannot be comprehensively understood, which will restrict the performance. In order to improve the image captioning model, the model introduced by Jiang et al. \cite{jiang2018recurrent} extracts diverse representations from multiple encoders. The novel recurrent fusion network (RFNet) proposed in the paper uses multiple CNNs as encoders. Each representation extracted from an individual CNN can act as an individual view of the image content.
\subsubsection{Injecting Spatial and Semantic Relation Information into Attention-Based Methods}
One of the significant downsides of the methods that only use the attention mechanism as their main solution for image captioning is that these methods fail to consider the spatial and semantic relations between image elements. Spatial and semantic relations in an image are integral to comprehension of the image contents \cite{you2016image, wu2016value, gan2017semantic, yao2017boosting, zhou2016image}. For example, spatial relations in an image could help differentiate between "a person riding a horse" and "a person standing on a horse's back." Also, relative size can help differentiate between objects with their most significant difference being their size, like violins and cellos. In addition to that, incorporating these relations makes the object detection task more precise. As a possible solution, Herdade et al. \cite{herdade2019image} have introduced the "Object Relation Transformer". Pan et al. \cite{pan2020x} have mentioned another problem about the attention-based image captioning methods: in most of these methods, only the first-order interactions between objects inside the image are observed. Their paper has claimed that since the image captioning problem involves multi-modal data (image and text), multi-modal reasoning is needed, and observing the first-order interaction between features only will render more in-depth reasoning impossible. The structure proposed by Pan et al. uses spatial and channel-wise bilinear attention to extract second-order interactions.
Liu et al. \cite{liu2020exploring} claim that there is still great difficulty in deep image understanding; because the systems tend to view one image as unrelated individual segments and are not guided to comprehend the relationships between the objects inside the image. They argue that such understanding requires adequate attention to correlated image regions and coherent attributes of interest. To do so, they have presented the "Global-and-Local Information Exploring-and-Distilling (GLIED) approach.\newline
To pre-train the models and methods discussed so far, existing methods mostly concatenate detected regions and textual features and use self-attention to learn the semantic alignments between the two modalities. These methods suffer from two main issues: ambiguity and a need for more grounding \cite{li2020oscar}.

\textbf{Ambiguity: }The methods that utilize the spatial and semantic relations between objects in images use object detectors to locate salient objects. These detectors usually generate redundant regions, and the visual features extracted using these detectors are extracted from overly sampled regions. The regions belonging to multiple objects might overlap heavily and cause ambiguity in the extracted visual embeddings.

\textbf{Lack of grounding: }the object tags used in previous methods are not associated with both object regions and word embeddings, resulting in a lack of grounding. Also, the attention models so far do not focus on the same regions as a human would when looking at an image \cite{das2017human}. However, salient objects are usually present in both image and the corresponding caption, which can be used as anchor points to ease the process of training in the vision-language tasks. The pre-training method proposed by Li et al. \cite{li2020oscar} leverages these anchor points to tackle the mentioned issues.
In many vision-language pre-training methods, the pre-trained model is fine-tuned for downstream tasks. However, it is challenging to pre-train a single, unified model that is universally applicable to a wide range of vision-language tasks via fine-tuning. Zhou et al. presented a pre-training method for a unified representation for encoding and decoding in \cite{zhou2020unified}. The Vision-Language Pre-training (VLP) model proposed in this paper has the advantage of unifying the encoder and decoder and learning a more universal contextualized vision-language representation, which can be fine-tuned for generation and understanding tasks easily. This unified procedure results in a single model architecture for the two distinct vision-language prediction tasks (bidirectional and seq2seq). This alleviates the need to train multiple pre-training models for different tasks without significant performance loss. In order to fine-tune for the image captioning task, the VLP model is fine-tuned on the target dataset using the seq2seq objective. Cornia et al. \cite{cornia2019show} claim that an attention-based architecture implicitly selects which regions to focus on, but it does not provide a way of controlling which regions are described and what importance is given to each region. The model suggested in their paper is able to focus on different regions in different orders following a given condition. Words can be organized into a tree-like structure, and a higher level of abstraction can be recovered considering the syntactic dependencies between words.
\subsection{Using Graphs for Spatial and Semantic Relations}
Graphs have been used extensively in many image captioning methods due to their ability to cohesively represent the relation between multiple elements. These methods have utilized graphs in two ways: scene graphs extracted from images and scene graphs extracted from textual data. Scene graphs have been used as a component inside encoder-decoder-based or unsupervised frameworks, and some have employed scene graphs along with Transformers. Graph-based methods pose challenges of their own. 
Yang et al. \cite{yang2019auto} rightfully state that an ever-present problem has never been substantially resolved: the different variants of the encoder-decoder-based framework, when fed an unseen image, usually produce a simple and trivial caption about the salient objects in the image, which is no better than a list of object detection. The model presented by Yang et al. adds the inductive bias of language generation to the encoder-decoder framework and uses scene graphs to connect the image and text modalities. 

Gu et al. \cite{gu2019unpaired} argue that the majority of image captioning studies are conducted in English, and preparing image-caption paired datasets in other languages requires human expertise and is time-consuming. The method introduced in their paper uses scene graphs as an intermediate representation of the image and sentence and maps the scene graphs in their feature space using cycle-consistent adversarial training.
\subsection{Using Attention and Graphs}
Considering how mutual correlations or interactions between objects are the natural basis for image description, Yao et al. \cite{yao2018exploring} study the visual relationships between objects and how they can be utilized for this matter. They have built semantic and spatial correlations on image regions and used Graph Convolutions to learn richer representations. One major challenge of image captioning is the problem of grounded captioning. Most models do not focus on the same image regions as a human would while observing an image, which may lead to object hallucination \cite{rohrbach2018object}. Zhong et al. \cite{zhong2020comprehensive} addressed this problem by revisiting the representation of image scene graphs. The key idea is to select essential sub-graphs and only decode a single target sentence from a chosen sub-graph. The model can link the decoded tokens back into the image regions, demonstrating noticeable results for caption grounding. Another downside of the attention-based methods is that they do not incorporate the regions previously attended by the attention model. These regions can be used in the module’s following region selections. Wang et al. \cite{wang2020learning} have integrated this point as well as the semantic relations between image elements in their proposed structure, which uses a novel content-based attention framework to store previously attended image regions. Chen et al. have discussed in \cite{chen2020say} that even though some methods focus on controlling expressive styles or attempt to control the description contents (discussed in section \ref{section_multistyle}), they can only handle a coarse-level signal. Their method uses a directed graph consisting of three node types grounded in the image, which allows for incorporating user intentions. Li et al. \cite{li2019know} argue that most methods that devise semantic concepts treat entities in images individually and lack helpful, structured information. Therefore, they have utilized scene graphs along with CNN features from the bounding box offsets of object entities. Another work by Xu et al. \cite{xu2019scene} also addresses the lack of structured information in current systems. The authors propose the Scene Graph Captioner (SGC), which is divided into three major components: The graph embedding model, the attention extraction model, and the language model. The attention extraction model is inspired by the concept of \textit{small world} in the human brain.
The work proposed by Lee et al. \cite{lee2019learning} uses visual relations from scene graphs directly instead of GCNs, claiming that it will avoid expensive graph convolutions. While the performance of some GCN-based models is slightly better, evading graph convolutions may be reasonable in some frameworks. 
\subsection{Using Convolutional Network-Based Methods}
LSTM networks \cite{hochreiter1997long} have been considered the standard for vision-language tasks such as image captioning and visual question answering due to their impressive ability to memorize long-term dependencies through a memory cell. However, training such networks can be considerably challenging due to the complex addressing and overwriting mechanism combined with the required processing being inherently sequential, and the significant storage required in the process. LSTMs \cite{hochreiter1997long} also require more careful engineering when considering a novel task \cite{aneja2018convolutional}. Earlier, CNNs could not perform as well as LSTMs on vision-language tasks. The recent advances in convolutional structures on other sequence-to-sequence tasks have enabled researchers to use CNNs in many other vision-language tasks. Also, CNNs produce more entropy \cite{aneja2018convolutional}, which can be helpful for diverse predictions, have better classification accuracy, and do not suffer from vanishing gradients. Aneja et al. \cite{aneja2018convolutional} proposed a convolutional model which uses masked convolutions instead of LSTM or GRU units. This work also experimented with attention by forming an attended image vector and adding it to the word embedding at every layer. Doing so, the model has outperformed the attention baseline \cite{xu2015show}. With attention, the model could identify salient objects for the given image. Arguing that RNNs or LSTMs \cite{hochreiter1997long}, which are widely used in image captioning, cannot be computed in parallel and also ignore the underlying hierarchical structure of the sentences, Wang et al. \cite{wang2018cnn} designed a framework entirely relying on CNNs. The proposed model can be computed in parallel and is faster to train. However, convolutional-network based methods still need improvement in terms of performance.
\subsection{Using Transformers}
The encoder-decoder framework continues to dominate the image captioning world, with the models only varying in details and sub-modules. The recent success of Transformers in natural language processing tasks has inspired many researchers to replace the RNN model with Transformer in the decoders, aiming to benefit from its excellent performance and the possibility of parallel training. Transformers have been the center of attention in the computer vision field as well, with models such as DETR \cite{carion2020end},  ViT \cite{dosovitskiy2020image}, SETR \cite{zheng2021rethinking}, and IPT \cite{chen2021pre}.

Liu et al. \cite{liu2021cptr} have proposed the CaPtion TransformeR (CPTR), a full Transformer network to replace the widely used CNN in the encoder part of the encoder-decoder framework. 
Fang et al. \cite{fang2022injecting} criticize the use of object detectors as a tool to provide visual representation, stating that it may lead to heavy computational load and that they require box annotations. Fang et al. \cite{fang2022injecting} have introduced the detector-free ViTCAP model with a fully Transformer architecture, which uses grid representations without regional operations.

Nguyen et al. \cite{nguyen2022grit} mention another issue with CNN-based detectors. CNN-based detectors use non-maximum suppression (NMS) at the last stage of computation to remove redundant bounding boxes. As a consequence, end-to-end training of an entire model consisting of detector and decoder modules becomes difficult. To overcome this problem and to reduce their high computation cost, Nguyen et al. employ the Deformable DETR \cite{zhu2020deformable} and replace the CNN backbone in the original design with Swin Transformer.
The COS-Net model (Comprehending and Ordering Semantics Network) proposed by Li et al. \cite{li2022comprehending} aims to unify semantic comprehending and ordering. COS-Net uses a CLIP model (image encoder and text encoder) \cite{radford2021learning} is utilized as a cross-modal retrieval model which retrieves sentences semantically similar to the input image.

Zeng et al. \cite{zeng2022s2} argue that directly operating at grid features may lead to the loss of spatial information caused by the flattening operation. The objective of the Pseudo-supervised (SP) module designed by the authors is to resolve this issue. Also, the Scale-wise Reinforcement (SR) module has been introduced to maintain the model size and improve performance.
Wang et al. \cite{wang2022end} argue that using a network such as Faster-RCNN as the encoder divides the captioning task into two stages and thus limits it. The PureT model built by the authors is a pure Transformer-based structure that integrates the captioning task into one stage and enables end-to-end training.

ExpansionNet v2 introduced by Hu et al. \cite{hu2022expansionnet} aims to solve the problem of performance bottlenecks in the input length in Deep Learning methods for image captioning. To address this issue, the authors introduce a new technique called Block Static Expansion, which distributes and processes the input over a collection of sequences with different lengths. This method helps to improve the quality of features refinement and ultimately increase the effectiveness of the static expansion.
\subsubsection{Using Graphs and Transformers}
Some current captioning encoders use a GCN to represent the relation information. Yang et al. \cite{yang2022reformer} highlight that these encoders are ineffective in image captioning due to the use of methods such as Maximum Likelihood Estimation rather than a relation-centric loss and the use of pre-trained models to obtain relationships instead of the encoder to improve model explainability. Yang et al. propose the ReFormer architecture, which applies the objective of scene graph generation and image captioning by means of one modified Transformer.
Chen et al. \cite{chen2021captioning} use the Transformer as their base architecture in the model SGGC (Scene Graph Guiding Captioning). The encoder is composed of two sub-encoders named visual encoder and semantic encoder. In the visual encoder sub-component, a Transformer encoder consisting of $N$ identical encoding layers has been used instead of the general CNN-based encoder to capture the relationships between visual regions better. Scene graphs have been used as additional guidance for decoder generation.
While Transformers are suitable for self-supervised pretext tasks on large-scale data, training can become expensive and burdensome. There is a need for more economic Transformer-based large-scale multi-modal models which can be achieved by means of incorporating more inductive bias about vision and language data \cite{xu2022image}.
\subsection{Using Vision-Language Pre-Training for Image Captioning}
Vision-language pre-training (VLP) has remarkably contributed to the recent advances in image captioning and is currently the dominant training method for vision-language (VL) tasks. In VLP approaches, a large-scale model is usually pre-trained on massive amounts of data using self-supervised learning, and then generalized to adapt to downstream tasks. Studies by \cite{li2020oscar, zhang2021vinvl, zhou2020unified} have extensively observed the effect of pre-training objective methods and architectures. The scale of the pre-training dataset is also believed to be a crucial factor in outstanding performance. VLP helps alleviate some of the problems experienced in conventional image captioning methods. The conventional methods typically need to minimize the gap between the visual and textual modals and are therefore resource-hungry \cite{mokady2021clipcap}. Excessive training time and numerous trainable parameters are also required, reducing their practicality. On the other hand, given new samples, the models need to be updated to adapt to new inputs. This brings about the need for lightweight models with faster training times and fewer parameters. 
It has recently been observed that powerful vision-language pre-trained models improve zero-shot performance dramatically and reduce training time. One such pre-trained model is CLIP (Contrastive Language-Image Pre-Training). 
Mokady et al. \cite{mokady2021clipcap} have leveraged CLIP encoding as prefix to the captions in their ClipCap model. A lightweight Transformer-based mapping network is trained from the CLIP embedding space and a learned constant. The GPT-2 network is used as the language model to generate captions given the prefix embeddings.
Hu et al. \cite{hu2021vivo} point out that while many VLP methods have been introduced that learn vision-language representations through training large-scale Transformer models, most are designed for understanding tasks. The few solutions that can be applied to image captioning \cite{li2020oscar, zhou2020unified} use paired image-caption data for pre-training, which cannot improve zero-shot performance. VIVO (VIsual VOcabulary pre-training), proposed by Hu et al., learns vision-language alignment on image-tag pairs. Since caption annotations are not needed, many existing vision datasets originally prepared for tasks such as image tagging or object detection can be used. 
Xia et al. \cite{xia2021xgpt} emphasize that VL generation tasks necessitate the ability to learn generation capabilities as well as the ability to understand cross-modal representations. Also, Xia et al. explain that the pre-trained models developed for understanding tasks only provide the encoder, and separate decoders need to be trained to enable generation. In addition to this deficiency, none of the pre-training tasks are designed for the whole sentence generation. The XGPT (Cross-modal Generative Pre-Training for Image Captioning) takes advantage of a cross-modal encoder-decoder architecture and is directly optimized for VL generation tasks. Three generative pre-training tasks have been designed to countervail the lack of pre-training objectives for generation tasks, namely: Image-conditioned Masked Language Modeling (IMLM), Image-conditioned Denoising Autoencoding (IDA), and Text-conditioned Image Feature Generation (TIFG).

Li et al. \cite{li2022mplug} mention computational inefficiency and information asymmetry as some of the shortcomings of existing pre-trained models. Li et al. \cite{li2022mplug} have proposed the mPLUG model, which incorporates a novel cross-modal fusion mechanism with cross-modal skip-connections to alleviate these problems. 
Liu et al. \cite{liu2023prismer} point out the data insensitivity problem and heavy computations associated with current vision-language problem, and take a different approach in their model Prismer to learn domain knowledge via distinct and separate sub-networks referred to as experts. Prismer includes modality-specific experts that encode multiple types of visual information directly from their corresponding network outputs. The expert models are pre-trained and frozen individually and are connected through lightweight trainable components. This approach results in a significant reduction in total network parameters.

\subsection{Using Unsupervised Methods and Reinforcement Learning}
The research works discussed in the aforementioned categories used a combination of images and their corresponding captions to train the structures they introduced and generated captions for new images while optimizing metrics. Training these supervised methods is challenging and involves some problems. One problem is that most of the research on image captioning has only worked on generating captions in the English language, and a proper dataset consisting of captions in multiple languages is not available. Preparing such a dataset requires the skills of human experts and is very time-consuming. Preparing a dataset of images and their corresponding captions is generally a difficult task. The Microsoft COCO dataset \cite{lin2014microsoft}, which is widely used in image captioning is much smaller than other datasets specifically designed for the object detection task, such as ImageNet and Open Images \cite{cornia2020meshed}. Microsoft COCO dataset \cite{lin2014microsoft} has 100 object classes only; consequently, the models trained on this dataset fail to generalize for new images that were not covered in the dataset. A considerable part of image captioning research is moving towards unsupervised methods to solve these issues. The early works improved the diversity of the captions; however, they sacrificed overall performance. Feng et al. \cite{feng2019unsupervised} have used a sentence corpus, a visual concept detector, and a set of images for unsupervised training. The model is composed of an image encoder, a sentence generator, and a discriminator. The results obtained from this research work have been criticized by Gu et al. \cite{gu2019unpaired} (discussed in section \ref{section_graph_and_spatial_semantic}). It has been explained that considering the limitations imposed by supervised learning, this research work has not achieved significant results, and the performance of the proposed model is not satisfactory. Gu et al. \cite{gu2019unpaired} use an unsupervised method (CycleGAN) to align the scene graph and the captions. Chen et al. \cite{chen2019improving} point out one issue with conventional encoder-decoder structures: many directly optimize one or a combination of metrics. This can not guarantee consistent improvement over all metrics. As a solution, Chen et al. have designed a discriminator network based on the idea of GANs, which judges if a caption is human-generated or produced by a machine. Liu et al. \cite{liu2018show} have introduced a system consisting of a captioning module and a self-retrieval module. The notable part of this work is the self-retrieval module (which uses the REINFORCE algorithm) that improves the performance of the aforementioned structure while only training on partially labeled data. 
\subsection{Captioning in Multiple Styles}
Some of the papers covered in this survey generate captions in multiple styles, with some of these styles being humorous or hostile. The structure called "TransResNet" presented by Shuster et al. \cite{shuster2019engaging} considers two classes of models: retrieval and generative. While the retrieval model has given better results, a disadvantage of the retrieval models for caption generation is that these models do not produce a new caption and only choose a caption from a massive dataset. The retrieval models usually generate general and repetitive captions. This pushes many researchers to use unsupervised methods. Guo et al. \cite{guo2019mscap} have stated that incorporating appropriate styles into captions will enrich their clarity and appeal and allows for user engagement and social interactions. The structure presented by Guo et al. is composed of five modules for caption generation in different styles. 

Stylized captions can help improve user interaction. However, since neutral captions that report factual data are more appropriate for visually impaired individuals, stylized captions may not be the best choice to utilize in assistive technologies.

\section{Datasets and Performance Comparison}
The methods discussed in previous sections use various datasets and are evaluated with multiple evaluation metrics. In this section, we review the datasets and metrics widely used in recent research works in depth. The available datasets for the image captioning task are still small compared to that of object detection, and the evaluation metrics have many limitations. Considering the increasing importance of the image captioning task, preparing richer datasets and more accurate metrics can be vital to the growth and improvement of the task.

\subsection{Datasets Used by Recent Works}

\subsubsection{Microsoft COCO}
The MS COCO dataset \cite{lin2014microsoft} is a vast dataset for object detection, image segmentation, and image captioning. This dataset contains many features, such as image segmentation, 328,000 images, 91 object classes, and five captions for each image.
\subsubsection{Flickr30K}
This dataset \cite{young2014image} is introduced for the automatic image captioning and grounded language understanding task. This dataset contains 31,000 images collected from the Ficker website, along with 158 thousand captions written by humans. This dataset contains a detector for everyday objects, a color classifier, and a bias toward selecting larger objects. This dataset has not specified any split settings for training, testing, and evaluation, and researchers use any split settings they desire when using this dataset.
\subsubsection{Flickr30K Entities}
The Flickr30K Entities dataset \cite{plummer2015flickr30k} is based on the Flickr30K dataset and contains 158k captions from Flickr30K with 244k coreference chains which link mentions of the same entities in images. The dataset also contains 276k manually annotated bounding boxes corresponding to each entity.

\subsubsection{Visual Genome}
Unlike the other dataset discussed which only had one caption for the entire image, this dataset \cite{krishna2017visual} presents a separate caption for each image region. This dataset comprises seven parts: region descriptions, attributes, relationships, region graphs, scene graphs, and question-answer pairs. The Visual Genome dataset contains more than 108 thousand images, with each image having an average of 35 objects, 26 attributes, and 21 pairwise relationships between objects.
\subsubsection{FlickrStyle10k}
This dataset \cite{gan2017stylenet} contains 10 thousand images with captions of varying styles. Training data consists of 7 thousand images, and the testing and evaluation data consist of 2 thousand and 1 thousand images, respectively. Each image has captions in different styles, such as poetic, humorous, and neutral (factual).

In addition to the datasets commonly used by many research works, some have collected and prepared their own dataset.

\subsubsection{TextCaps}
This dataset  \cite{sidorov2020textcaps} aims to help train visual assistants for visually impaired individuals, focusing on presenting captions for images with written text inside them. According to the data reported in \cite{bigham2010vizwiz}, 21 percent of the questions asked by visually impaired individuals involved written texts inside images. This dataset presents 145 thousand captions for 28 thousand images.
\subsubsection{VizWiz-Captions}
This dataset \cite{gurari2020captioning} has been introduced as a dataset appropriate for image captioning for visually impaired individuals. This dataset consists of 23,431 training images and 117,155 training captions, 7,750 validation images, 38,750 evaluation captions, 8,000 images, and 40,000 testing captions. The images have been taken directly by visually impaired individuals.

There are some datasets recently introduced that are not used by many works yet are potential choices for future research works.
\subsubsection{Google's Conceptual Captions}
This dataset \cite{sharma2018conceptual} consists of approximately 3.3 million images and captions. The images have been collected from the Internet first, along with the "alt-text" associated with them. These image-caption pairs have been then filtered and processed to extract appropriate captions for the images that describe image contents. This dataset is split into training and evaluation splits. There are 3,318,333 image-caption pairs in the training split and 15,840 image-caption pairs in the evaluation split.
\subsubsection{nocaps}
The "Novel Object Captioning at Scale (nocaps)" dataset \cite{agrawal2019nocaps} has been presented to encourage the development of captioning models that can surpass the limitation of visual concepts in existing datasets. The introduced benchmark is composed of 166,100 human-generated captions describing 15,100 images from the Open Images validation and test sets. The training data consists of Open Images image-level labels and object bounding boxes in addition to COCO image-caption pairs. Considering that Open Images contains many more object classes not present in COCO, about 400 object classes in test images have almost no associated training captions. 
\subsubsection{Open Images V6: Visual Relationships}
The Open Images dataset \cite{OpenImagesv4} contains various sections for object detection, image segmentation, object relationships, and more. The dataset includes approximately 9 million images in 600 different classes. Each image contains an average of 8.4 objects. One section of this dataset is the Visual Relationships section which contains 329 tertiary relationships for 375 thousand images. These relationships are in the forms of human-object (for example, "a person holding a microphone"), object-object (for example, "a dog inside a car"), and object-attribute (for example, "bench is wooden," "handbag is made of leather"). The most recent version of this dataset is available at \cite{openimagesv6}.
\subsubsection{Open Images V6: Localized Narratives}
In 2020 and the sixth version of the Open Images dataset, a new section under the name of "Localized Narratives" was added \cite{PontTuset_eccv2020}. This new section contains 1 million and 671 thousand images from the Open Images Dataset. A human describer has described each image in the dataset via a voice recording while moving their computer mouse on the regions they were describing. Since the words of the caption are in sync with the mouse movements, the location associated with each word is available.
\subsubsection{SentiCap}
SentiCap \cite{mathews2016senticap} is a sentiment captioning dataset based on the MS COCO dataset \cite{lin2014microsoft}. There are three positive and three negative sentiment captions for each image. The positive sentiment subset consists of 2,873 sentences and 998 images for training and another 2,019 sentences over 673 images for testing. The negative sentiment subset consists of 2,468 sentences and 997 images for training and another 1,509 sentences over 503 images for testing.
\subsubsection{SBU Captions Dataset}
The SBU dataset \cite{ordonez2011im2text} consists of 1 million images and their corresponding descriptions given by individuals as they upload an image to Flickr. The captions are not guaranteed to be visual or unbiased. Therefore this dataset has more noise compared to other datasets.
\subsubsection{The Karpathy Split }
In order to train machine learning learners, the dataset is usually split into training, evaluation, and testing split. One typical split for the datasets widely used in recent works is a split called Karpathy \cite{karpathy2015deep}, which splits the dataset into 5000 images for offline testing, and 5000 images for offline evaluation and uses the rest for training. Most research works use this splitting method to be more consistent with other works.

The details regarding datasets discussed in this section have been summarized in table \ref{table:1}.

\begin{table*}[hbt!] 
  \caption{Most common datasets: details (R.D. indicates "Region Descriptions", L.N. indicates "Localized Narratives", and V.R. indicates "Visual Relationships)}
  \centering
  \label{tab:datasetsAndEvals}
  \begin{tabular}{c|c|c|c|c}
    \toprule
    \textbf{Dataset} &\textbf{Total Images} & \textbf{Objects/Image} & \textbf{Object Classes} & \textbf{Captions/Image}\\
    \midrule
    Visual Genome \cite{krishna2017visual} & 108,077 & 36.17 & 80,138 & 5.4m R.D.\\
    MS COCO\cite{lin2014microsoft} & 330,000 & 7.57 &91 & 5 \\
    Flickr30K Entities\cite{plummer2015flickr30k} & 31,783 & 8.7 & 44,518 & 5\\
    OpenImagesV6:V.R.\cite{openimagesv6} & 375,000 & 8.4 & - & 1\\
    Flickr30K\cite{young2014image}& 31,000 & - & - & 5 \\
    FlickrStyle10K\cite{gan2017stylenet} & 10,000 & - & - & 2 \\
    OpenImagesV6:L.N.\cite{PontTuset_eccv2020} & 849,000 & - & - & 1\\
    SBU Captions\cite{ordonez2011im2text} & 1 mil & - & - & 1\\
    SentiCap\cite{mathews2016senticap} & 3171 & - & - & 6\\
    TextCaps\cite{sidorov2020textcaps} & 28,408 & - & - & 5\\
    VizWiz-Captions\cite{gurari2020captioning} & 39,181 & - & - & 5\\
    nocaps\cite{agrawal2019nocaps} & 15,100 & - & 680 & 11\\
    Conceptual Captions\cite{sharma2018conceptual}&3 mil< & - & - & 1\\
    \bottomrule
  \end{tabular}
  
\label{table:1}
\end{table*}

\subsection{Evaluation Metrics for Image Captioning Methods}
The metrics discussed below fall into two categories: the \emph{text evaluation} metrics and the \emph{caption evaluation} metrics. The text evaluation metrics evaluate machine-generated text portions independently. Most of these metrics were introduced for evaluating the text generated by machine translation models. The caption evaluation metrics evaluate the captions generated by the models and have been designed specifically for the image captioning task.

\subsubsection{BLEU (Bilingual Evaluation Understudy)}
BLEU \cite{papineni2002bleu} is an evaluation metric for machine-generated texts. Separate parts of a text are compared against a set of reference texts, and each part receives a score. The overall score is an average over these scores; however, the syntactical correctness is not evaluated. The performance of this metric varies based on the references used and the size of the generated text. The BLEU metric is a widely used metric due to being a pioneer in evaluating machine-generated texts, being independent of language, their simplicity, high speed, low cost, and being quite comparable with human judgment. BLEU counts the consistent $n$-grams in the machine-generated text and the reference text. $n$-grams are a contiguous sequence of n items in a text in the field of computational linguistics and probability. These items can be phonemes, syllables, letters, words, or base pairs. The number "n" determines the number of grams that will be compared against each other. Usually, BLEU-1, BLEU-2, BLEU-3, and BLEU-4 are computed using the BLEU metric. To compute BLEU-n, the $n$-grams of 1 to "n" are computed, and each is assigned one single weight. Next, the geometric mean of these $n$-grams is calculated according to these weights. For example, when computing BLEU-4, the $n$-grams of 1 to 4 are calculated, and each is given the value 0.25 as their weight, followed by the geometric mean being computed over these values. This metric does have some disadvantages, such as the fact that the computed scores are only high when the generated text is short. Also, in some cases, a high score achieved using this metric is unreliable and does not mean a higher-quality text.

\subsubsection{ROUGE (Recall-Oriented Understudy for Gisting Evaluation)}
ROUGE \cite{lin2004rouge} is a set of metrics that evaluate the quality of text summarization. This metric determines the quality of a summary by comparing it to other ideal human-created summaries: the number of overlapping units, like $n$-grams, word sequences, and word pairs between the machine-generated summary and the ideal summaries are counted. Multiple measures are introduced: ROUGE-N (which counts the overlap of $n$-grams between the machine-generated summary and the ideal summary. ROUGE-1 and ROUGE-2 are subsets of ROUGE-N), ROUGE-L (which is essentially longest common subsequence based statistics, and considers sentence level structure similarity naturally and automatically finds the longest co-occurring in sequence $n$-grams), ROUGE-W (which is based on weighted longest common subsequence that prefers consecutive longest common subsequences), ROUGE-S (which is based on skip-bigram co-occurrence, with skip-bigram being any pair of words in their sentence order), and ROUGE-SU (which is based on skip-bigram plus unigram co-occurrence), with each being used in a specific application. This metric does not perform well for evaluating summaries in more than one text.
\subsubsection{METEOR (Metric for Evaluation of Translation with Explicit Ordering)}
This metric \cite{banerjee2005meteor} compares word segments against reference texts. This method is based on the harmonic mean of unigram precision and recall (recall is weighted higher than precision). METEOR has features such as stemming and synonymy matching in addition to the standard exact word matching. This metric makes a better correlation at the sentence level or segment level.
\subsubsection{CIDEr (Consensus-based Image Description Evaluation)}
This metric \cite{vedantam2015cider} is explicitly designed for evaluating image captions and descriptions. In contrast to other metrics working with only five captions per image-which makes them unsuitable for evaluating the consensus between the generated captions and human judgments- CIDEr reaches this level of consensus using term-frequency inverse document frequency (TF-IDF). CIDER is technically an annotation modality for automatically computing consensus. A measure of consensus encodes how often $n$-grams in the candidate sentence are present in the reference sentences. Also, $n$-grams not present in the reference sentences must not exist in the candidate sentences.
Furthermore, lower weight must be given to $n$-grams frequently appearing across all images in the dataset since they are likely to contain less information. To encode this, Vedantam et al. \cite{vedantam2015cider} performed a Term-Frequency Inverse Document Frequency (TF-IDF) weighting for each $n$-gram. A version of CIDEr called CIDEr-D exists as a part of the Microsoft COCO evaluation server.
\subsubsection{SPICE (Semantic Propositional Image Caption Evaluation)}
The SPICE metric \cite{anderson2016spice} is a metric for evaluating image captions based on semantic context. This metric measures how well objects, attributes, and the relations between them are covered in image captions. A scene graph is used to extract the names of different objects, attributes, and the relationships between them from image captions. The metric utilizes semantic representations produced by this graph.

The discussed methods are far from human judgment in terms of quality due to various factors. Using external knowledge databases along with evaluation metrics can help improve evaluation quality.

\subsection{Performance Comparison Based on MSCOCO Test Servers}
Microsoft COCO has presented an online server for testing and evaluation purposes to enable a more fair and uniform testing platform. In Tables \ref{table:2} and \ref{table:3}, the results from the research works which have used the Microsoft COCO test servers have been listed. The Microsoft COCO test servers report two numbers for each evaluation metric: c5 and c40. c5 is computed using five reference captions, and c40 is computed using 40.
The best performances are highlighted with boldface font.
\begin{table*}[hbt!]
  \caption{The reported results obtained from Microsoft COCO servers - Top 10 methods (B:BLEU \cite{papineni2002bleu})}
  \centering
  {\begin{tabular}{c|cc|cc|cc|cc}
    \toprule
    \textbf{Reference}&\multicolumn{2}{|c|}{\textbf{B-1}}&\multicolumn{2}{|c|}{\textbf{B-2}}&\multicolumn{2}{|c|}{\textbf{B-3}}&\multicolumn{2}{|c}{\textbf{B-4}}\\
    &\multicolumn{2}{|c|}{c5\;\;\;\;\;\;c40}
    &\multicolumn{2}{|c|}{c5\;\;\;\;\;\;c40}
    &\multicolumn{2}{|c|}{c5\;\;\;\;\;\;c40}
    &\multicolumn{2}{|c}{c5\;\;\;\;\;\;c40}\\
    \midrule

Nguyen et al. \cite{nguyen2022grit}&
\textbf{84.1}&  \textbf{97.6}&   \textbf{69.4}&   \textbf{93.5}&   \textbf{54.9}&   \textbf{86.3}&   \textbf{42.5}&   \textbf{76.8} \\

Hu et al. \cite{hu2022expansionnet}&
83.3&   96.9&   68.8&   92.6&   54.4&   85.0&   42.1&   75.3\\

Anderson et al. \cite{anderson2018bottom}&
80.2&	95.2&	64.1&	88.8&	49.1&	79.4&	36.9&	68.5\\


Chen et al. \cite{chen2019improving}&
81.9&	95.6&	66.3&	90.1&	51.7&	81.7&	39.6&	71.5\\

Cornia et al. \cite{cornia2020meshed}&
81.6&	96.0&	66.4&	90.8&	51.8&	82.7&	39.7&	72.8\\




Jiang et al. \cite{jiang2018recurrent}&
80.4&	95.0&	64.9&	89.3&	50.1&	80.1&	38.0&	69.2\\


Li et al. \cite{li2022comprehending} & 83.3 & 96.8 & 68.6 & 92.3 & 54.2 & 84.5 & 42 & 74.7\\

Liu et al. \cite{liu2020exploring}&
80.1&	94.6&	64.7&	88.9&	50.2&	80.4&	38.5&	70.3\\

Pan et al. \cite{pan2020x} (X-LAN)&	81.4&	95.7&	66.5&	90.5&	52.0&	82.4&	40.0&	72.4\\

Pan et al. \cite{pan2020x} (X-Transformer)&	81.9&	95.7&	66.9&	90.5 & 52.4&	82.5&	40.3&	72.4\\


Yang et al. \cite{yang2019auto}& -& -&  -&  -&  -&  -&  38.5&   69.7\\   

Yao et al. \cite{yao2018exploring}&
-&	-&	65.5&	89.3&	50.8&	80.3&	38.7&	69.7\\

Zeng et al. \cite{zeng2022s2} & 82.2 & 96.5 & 67 & 91.4 & 52.4 & 83.3 & 40.1 & 73.5\\

    \bottomrule
  \end{tabular}
  }
  \label{table:2}
\end{table*}

\begin{table*}[hbt!]
  \caption{The reported results obtained from Microsoft COCO servers (M:METEOR \cite{banerjee2005meteor}, R:ROUGE \cite{lin2004rouge}, C:CIDEr \cite{vedantam2015cider}, S:SPICE \cite{anderson2016spice})}
  \centering
  {\begin{tabular}{c|cc|cc|cc|cc}
    \toprule
    \textbf{Reference}&\multicolumn{2}{|c|}{\textbf{R}}&\multicolumn{2}{|c|}{\textbf{M}}&\multicolumn{2}{|c|}{\textbf{C}}&\multicolumn{2}{|c}{\textbf{S}}\\
    &\multicolumn{2}{|c|}{c5\;\;\;\;\;\;c40}
    &\multicolumn{2}{|c|}{c5\;\;\;\;\;\;c40}
    &\multicolumn{2}{|c|}{c5\;\;\;\;\;\;\;c40}
    &\multicolumn{2}{|c}{c5\;\;\;\;\;\;c40}\\
    \midrule

Nguyen et al. \cite{nguyen2022grit}&
\textbf{61.2}&   \textbf{77.1}&   \textbf{30.9}&   \textbf{41.0}&   \textbf{141.3}&  \textbf{143.8}&  - & - \\

Hu et al. \cite{hu2022expansionnet}&
60.8&   76.4&   30.4&   40.1&   138.5&  140.8& -& -\\

Anderson et al. \cite{anderson2018bottom}&
57.1&	72.4&	27.6&	36.7&	117.9&	120.5&	\textbf{21.5}&	\textbf{71.5}\\


Chen et al. \cite{chen2019improving}&
59.0&	74.4&	28.7&	38.2&	123.1&	124.3&	-&	-\\

Cornia et al. \cite{cornia2020meshed}&
59.2&	74.8&	29.4&	39.0&	129.3&	132.1&	-&	-\\



Huang et al. \cite{huang2019attention}&
58.9&	74.5&	29.1&	38.5&	126.9&	129.6&	-&	-\\

Jiang et al. \cite{jiang2018recurrent}&
58.2&	73.1&	28.2&	37.2&	122.9&	125.1&	-&	-\\


Li et al. \cite{li2022comprehending} & 60.6 & 76.4 & 30.4 & 40.1 & 136.7 & 138.3 & - & -\\

Liu et al. \cite{liu2020exploring}&
58.3&	73.8&	28.6&	37.9&	123.3&	125.6&	-&	-\\

Pan et al. \cite{pan2020x} (X-LAN)&	59.5&	75.2&	29.7&	39.3&	130.2&	132.8&	-&	-\\

Pan et al. \cite{pan2020x} (X-Transformer)& 59.5&	75.0&	29.6&	39.2&	131.1&	133.5&	-&	-\\



Yang et al. \cite{yang2019auto}&    58.6&   73.6&   28.2&   37.2&   123.8&    126.5&  -&  -\\

Yao et al. \cite{yao2018exploring}&
58.5&	73.4&	28.5&	37.6&	125.3&	126.5&	-&	-\\

Zeng et al. \cite{zeng2022s2} & 59.5 & 75 & 29.6 & 39.3 & 132.6 & 135 & - & -\\
    \bottomrule
  \end{tabular}
  }
  \label{table:3}
\end{table*}
According to the results collected in Tables \ref{table:2} and \ref{table:3}, the model proposed by Li et al. \cite{li2022comprehending} has achieved the best c5 and c40 results in most metrics. The model proposed by Aneja et al. \cite{aneja2018convolutional}-a convolutional-network based model- has achieved the lowest results in both c5 and c40 for almost all metrics. The model designed by Guo et al. \cite{guo2019mscap} has achieved the worst results in BLEU-2 c5 and BLEU-3 c5. Achieving worse results is natural, considering that the model operates with unpaired text, as there is no total consistency between image and text. The model proposed by Wang et al. \cite{wang2018cnn} falls at the lower part of the performance list in most metrics as well, which shows that convolutional-network based methods still need much more improvement to reach the performance of the other methods, such as the attention-based ones. Still, the benefits and advantages of convolutional-network based methods, such as a more straightforward training process (discussed more thoroughly in "Convolutional-network based methods," section \ref{convolutional}), encourage further research in this field.

\subsection{Comparing Independent Results}\label{comparing_independent_results}
Many research works have reported their results independently, as well as the results reported by the Microsoft COCO servers. A large number of these research works have used the code publicly available in \cite{cocoevaltool} to evaluate their performance. This codebase evaluates a models performance using BLEU \cite{papineni2002bleu}, METEOR \cite{banerjee2005meteor}, ROUGE \cite{lin2004rouge}, CIDEr \cite{vedantam2015cider} and SPICE \cite{anderson2016spice} metrics. 

In this section, we list the results reported independently (not obtained by Microsoft COCO servers) by the works covered in this survey in Tables \ref{table:4}, \ref{table:5}, \ref{table:6}, and \ref{table:7}. We have listed the best results for the research works that reported results under different settings (for example, optimization using different loss functions). The best performances are highlighted with boldface font.
\begin{table*}[hbt!]
  \caption{The independent results - Top 10 methods - BLEU-1 and BLEU-2 (B:BLEU \cite{papineni2002bleu}, Ref: Reference)}
  \centering
  {\begin{tabular}{cc|cc}
    \toprule
    \multicolumn{2}{c|}{\textbf{B-1}}&\multicolumn{2}{c}{\textbf{B-2}}\\
    \multicolumn{2}{c|}{ \quad \quad Ref \quad \;\quad \quad \quad Score}
    &\multicolumn{2}{c}{ \quad \quad Ref \quad \quad \quad \; Score}
   \\
    \midrule

Zhong et al.\cite{zhong2020comprehensive}& \textbf{90.7}&
Li et al.\cite{li2022comprehending} &\textbf{ 69.1}\\

Nguyen et al.\cite{nguyen2022grit}&84.2&
Pan et al.\cite{pan2020x}&66.8\\

Hu et al. \cite{hu2022expansionnet}&83.5&
Liu et al.\cite{liu2021cptr}&66.6\\

Li et al.\cite{li2022comprehending}&83.5&
Jiang et al.\cite{jiang2018recurrent}&64.7\\

Yang et al.\cite{yang2022reformer}&82.3&
Li et al.\cite{li2019know}&63.2\\

Cornia et al.\cite{cornia2020meshed}&82.0&
Liu et al.\cite{liu2018show}&63.1\\

Liu et al.\cite{liu2021cptr}&81.7&
Gu et al.\cite{gu2018stack}&62.5\\

Pan et al.\cite{pan2020x}&81.7&
Chen et al.\cite{chen2021captioning}&60.7\\

Huang et al.\cite{huang2019attention}&81.6&
Wang et al.\cite{wang2020learning}&60.3\\

Li et al.\cite{li2019entangled}&81.5&
Aneja et al.\cite{aneja2018convolutional}&55.3\\

    \bottomrule
  \end{tabular}
  }
  \label{table:4}
\end{table*}

\begin{table*}[hbt!]
  \caption{The independent results - Top 10 methods (B:BLEU \cite{papineni2002bleu}, Ref: Reference)}
  \centering
  {\begin{tabular}{cc|cc}
    \toprule
    \multicolumn{2}{c|}{\textbf{B-3}}&\multicolumn{2}{c}{\textbf{B-4}}\\

       \multicolumn{2}{c|}{ \quad \quad Ref \quad \;\quad \quad \quad Score}
    &\multicolumn{2}{c}{ \quad \quad Ref \quad \quad \quad \; Score}
   \\
    \midrule

Li et al.\cite{li2022comprehending}&\textbf{54.9}&
Zhong et al.\cite{zhong2020comprehensive}&\textbf{59.3}\\

Pan et al.\cite{pan2020x}&52.6&
Li et al. \cite{li2022mplug}&46.5\\

Liu et al.\cite{liu2021cptr}&52.2&
Li et al.\cite{li2022comprehending}&42.9\\

Jiang et al.\cite{jiang2018recurrent}&50.0&
Hu et al.\cite{hu2022expansionnet}&42.7\\

Li et al.\cite{li2019know}&48.3&
Nguyen et al. \cite{nguyen2022grit}& 42.4\\

Liu et al.\cite{liu2018show}&48.0&
Li et al.\cite{li2020oscar}&41.7\\

Gu et al.\cite{gu2018stack}&47.9&
Pan et al.\cite{pan2020x}&40.7\\

Wang et al.\cite{wang2020learning}&46.5&
Cornia et al.\cite{cornia2020meshed}&40.5\\

Chen et al.\cite{chen2021captioning}&46.2&
Liu et al. \cite{liu2023prismer}&40.4\\

Aneja et al.\cite{aneja2018convolutional}&41.8&
Huang et al.\cite{huang2019attention}&40.2\\

    \bottomrule
  \end{tabular}
  }
  \label{table:5}
\end{table*}

\begin{table*}[hbt!]
  \caption{The independent results - Top 10 methods ( M:METEOR \cite{banerjee2005meteor}, R:ROUGE \cite{lin2004rouge},  Ref: Reference)}
  \centering
  {\begin{tabular}{cc|cc}
    \toprule
    \multicolumn{2}{c|}{\textbf{M}}&\multicolumn{2}{c}{\textbf{R}}\\
    \multicolumn{2}{c|}{ \quad \quad Ref \quad \;\quad \quad  \;\;\;\;\;\;Score}
    &\multicolumn{2}{c}{ \quad \quad Ref \quad \;\quad \quad \;\;\;\;\;\; Score}\\
    \midrule

Zhong et al.\cite{zhong2020comprehensive}&\textbf{40.1}&
Zhong et al.\cite{zhong2020comprehensive}& \textbf{71.5}\\

Li et al.\cite{li2022mplug}&32.0&
Hu et al.\cite{hu2022expansionnet}&61.1\\

Liu et al.\cite{liu2023prismer}&31.4&
Li et al.\cite{li2022comprehending}&61.0\\

Hu et al.\cite{hu2022scaling}&31.4&
Nguyen et al.\cite{nguyen2022grit}&60.7\\

Li et al.\cite{li2022comprehending}&30.8&
Fang et al.\cite{fang2022injecting}&60.1\\

Nguyen et al.\cite{nguyen2022grit}&30.6&
Barraco et al.\cite{barraco2022unreasonable}&59.9\\

Hu et al.\cite{hu2022expansionnet}&30.6&
Yang et al.\cite{yang2022reformer}&59.8\\

Li et al.\cite{li2020oscar}&30.6&
Pan et al.\cite{pan2020x}&59.7\\

Fang et al.\cite{fang2022injecting}&30.1&
Cornia et al.\cite{cornia2020meshed}&59.5\\

Barraco et al.\cite{barraco2022unreasonable}&30.0&
Liu et al.\cite{liu2021cptr}&59.4\\


	



    
    
    
    
    
    
    
    
    \bottomrule
  \end{tabular}
  }
  \label{table:6}
\end{table*}

\begin{table*}[hbt!]
  \caption{The independent results - Top 10 methods (C:CIDEr \cite{vedantam2015cider}, S:SPICE \cite{anderson2016spice}, Ref: Reference)}
  \centering
  {\begin{tabular}{cc|cc}
    \toprule
    \multicolumn{2}{c|}{\textbf{C}}&\multicolumn{2}{c}{\textbf{S}}\\
    \multicolumn{2}{c|}{ \quad \quad \quad Ref \quad \;\quad  \;\;\;\;\;\; Score}
    &\multicolumn{2}{c}{ \quad \quad \quad Ref \quad \;\quad \quad \; Score}\\
    \midrule

Cornia et al.\cite{cornia2019show}&\textbf{209.7}&
Cornia et al.\cite{cornia2019show}&\textbf{48.5}\\

Chen et al.\cite{chen2020say}&204.2&	
Chen et al.\cite{chen2020say}&42.1\\

Zhong et al.\cite{zhong2020comprehensive}&166.7&
Zhong et al.\cite{zhong2020comprehensive}&30.1\\

Li et al.\cite{li2022mplug}&155.1&
Li et al.\cite{li2022mplug}&26.0\\

Hu et al.\cite{hu2022scaling}&145.5&
Hu et al.\cite{hu2022scaling}&25.5\\

Nguyen et al.\cite{nguyen2022grit}&144.2&
Hu et al.\cite{hu2022expansionnet}&24.7\\

Hu et al.\cite{hu2022expansionnet}&143.7&
Li et al.\cite{li2022comprehending}&24.7\\

Li et al.\cite{li2022comprehending}&143.0&
Li et al.\cite{li2020oscar}&24.5\\

Li et al.\cite{li2020oscar}&140.0&	
Liu et al. \cite{liu2023prismer}&24.4\\

Barraco et al.\cite{barraco2022unreasonable}&139.4&
Nguyen et al.\cite{nguyen2022grit}&24.3\\

    

	




    

    
    
    
    
    
    
    \bottomrule
  \end{tabular}
  }
  \label{table:7}
\end{table*}

Among the research works covered in this survey paper, \cite{zhong2020comprehensive} (which introduces a sub-graph proposal network along with an attention-based LSTM decoder) has had the best results in BLEU-1 (90.7) and BLEU-4 (59.3), as well as METEOR \cite{banerjee2005meteor} (40.1), and ROUGE \cite{lin2004rouge} (71.5), while \cite{li2022comprehending} (COS-Net, a model which uses CLIP image and text encoder as a cross-modal retrieval model) has had the best results in BLEU-2 (69.1) and BLEU-3 (54.9). Also, \cite{cornia2019show} (Show, Control and Tell) has achieved the best CIDEr and SPICE results (209.7 and 48.5, respectively).

A recurring pattern among the best-performing methods presented in Tables \ref{table:2} and \ref{table:3} is the application of Transformers, scene graphs, and vision language pre-training methods \cite{barraco2022unreasonable, chen2021captioning, chen2020say, cornia2020meshed, fang2022injecting, he2020image, hu2022scaling, huang2019attention,  li2019entangled,  li2019know,  li2020oscar, li2022comprehending, liu2021cptr, pan2020x, yang2022reformer, yang2019auto, zeng2022s2, zhong2020comprehensive}. These methods owe their performance to the capabilities of Transformers, scene graphs, and vision language pre-training methods. 
Transformers are capable of capturing complex relationships between objects and their surroundings, making them particularly effective in handling long-range dependencies in image sequences. Scene graphs, on the other hand, represent the relationships between objects within an image and allow for efficient inference of the visual content. Another desirable feature of graphs is their ability to represent composite and unstructured data types, as they provide a flexible and efficient way to model the complex relationships and interconnections between various entities within a system.
In addition to Transformers and scene graphs, some of the high-performing image captioning methods in Tables \ref{table:2} and \ref{table:3} also utilize vision language pre-training techniques \cite{barraco2022unreasonable, hu2022scaling, li2020oscar, li2022comprehending}. These methods involve training a model on large datasets that consist of both visual and textual information, allowing the model to learn a joint embedding space. By pre-training on such datasets and acquiring knowledge from multiple modalities, the model can effectively learn to understand visual content and generate natural language descriptions. The integration of these techniques in captioning models has led to a notable improvement in their overall performance, as evidenced by the results presented in Tables \ref{table:2} and \ref{table:3}.

\section{Challenges and the Future Directions}\label{challenges_future_directions}
Despite the abundance of solutions and methods presented to solve the image captioning problem, some challenges and open problems remain. The performance of the supervised methods relies significantly on the quality of the datasets. However, datasets can not cover the real world regardless of how massive they are, and the applicability of supervised methods is limited to the set of objects the detector is trained to distinguish. On the other hand, datasets with image-caption pairs inevitably contain more examples of a specific situation (one example being: "man riding a skateboard"). These examples in the training data falsely bias the model towards generating more captions similar to those examples rather than including actual detected objects \cite{li2019know}. The supervised paradigm overly relies on the language priors, which can lead to the object-hallucination phenomenon as well \cite{li2022comprehending}. 

The problems associated with the supervised methods encourage researchers to devise unsupervised techniques. On the other hand, due to the different properties of image and text modalities, the encoders of image and sentence cannot be shared. Therefore, the critical challenge in an unpaired setting is the gap of information misalignment in images, and sentences \cite{gu2019unpaired}. The current unsupervised image captioning methods still need to catch up in performance rankings. 

One promising direction of research is using scene graphs for image captioning. However, despite the many possibilities unveiled by scene graphs, discussed extensively in the previous sections, utilizing them comes with challenges. Constructing scene graphs is a complicated task in itself, and due to the interactions between objects being beyond simple pairwise relations, integrating scene graphs is quite tedious \cite{xu2019scene}. Also, scene graph parsers are still not as powerful \cite{yang2019auto, wang2019role}. According to some of the works which studied the impact of scene graphs on the quality of the captions, scene graphs are effective only if pre-training of the scene graph generators is done with visually relevant relation data \cite{lee2019learning}.

VLP methods have been used to resolve some of the flaws with supervised methods and object detector-based designs. However, most VLP approaches are catered to understanding tasks, and generation tasks such as image captioning demand more capabilities. A number of the recent works covered in this paper have aimed to fulfill this need. However, this field needs more investigation and analysis.
Moreover, detector-free designs have a rising popularity. In these designs, the detector is removed for the vision-language pre-training in an end-to-end fashion \cite{fang2022injecting}. Also, a general visual encoder replaces the detector and is used to produce grid features for later cross-modal fusion. However, the construction of a stronger detector-free image captioning model still needs investigation.
Despite the challenges faced when working with scene graphs, vision language pre-training methods and Transformers, almost each one of the best-performing models according to evaluation metrics use one or a combination of these techniques, as shown in section \ref{comparing_independent_results}. This further proves the potential of these techniques in solving the image captioning problem, and are promising tools for the future of generative tasks. Specifically, graphs are valuable in representing complex relationships and interconnections between different entities, particularly for composite, semi-structured, and unstructured data that may not be easily handled by other types of data models. Considering the recent advancements in generative artificial intelligence such as large language models (LLMs) \cite{openai_chatgpt} and multimodal language models (MLLMs) \cite{openai_gpt4, huang2023language, alayrac2022flamingo, li2023blip}, the need for representation methods capable of handling such data types will become more and more visible and felt in near future.

Another gap in the literature is the lack of focus on the application of image captioning for the visually impaired. Describing images can be the core of a vision assistant designed to aid the visually impaired in their daily lives: one can be informed of potential dangers in their environment and have a general understanding of what is happening around them.
Considering the issues mentioned earlier and gaps, unsupervised learning, and unpaired setting are of great potential. Also, the graph-based approach is expected to become even more popular in the near future. LLMs, MLLMs and Transformers in combination with vision-language pre-training methods are also very likely to become standard practice.

\section{Conclusion}
This paper has covered recent image captioning methods, provided a taxonomy of the approaches, and mentioned their features and properties. We also discussed the common problems in image captioning, reviewed datasets and evaluation metrics, compared the performance of the covered methods and algorithms in terms of experimental results, and highlighted the challenges and future directions in image captioning.
Despite the numerous methods and solutions presented for the image captioning problem, there are still some major problems and challenges for which few solutions have been suggested. On the other hand, the generated captions still need to be higher in quality and are far from human-generated captions. Also, the datasets cannot cover the infinite real world. The evaluation metrics still need to be improved and are still not ideal for evaluating the precise performance of the models. However, Vision-Language Pre-Training (VLP) methods are frequently used in recent works and have shown promising performance. VLP methods and Transformers are likely to be inseparable components of models in the future of image captioning.

Moreover, more research needs to be done on visual assistants for visually impaired individuals. Preparing such an assistant requires certain features to be implemented, making it different from the other applications of image captioning. The best models presented by the research works do not perform well as visual assistants and do not consider the specific demands and needs of visually impaired people. A proper caption for a visually impaired person includes the most important aspects of the image first and the other noticeable details afterward. The surroundings and finer details must also be described, such as details about the textures and the position of objects relative to each other. Therefore, a caption appropriate for the needs of visually impaired individuals is denser and contains much more detail compared to the captions generated by conventional methods and models. Also, the caption generation process may be altered in a way that the initial caption provided to the user can be more general and shorter. The caption may become denser and more detailed upon the user asking more questions about the image. 
Considering the importance of the aforementioned issues and the growing number of visually impaired individuals, a noticeable lack of an efficient solution remains. Valuable research work in this field would be automatic image captioning with a particular focus on creating a visual assistant for visually impaired individuals.

\setcitestyle{numbers}
\bibliographystyle{abbrvnat}
\bibliography{references}  

\begin{thebibliography}{161}
\providecommand{\natexlab}[1]{#1}
\providecommand{\url}[1]{\texttt{#1}}
\expandafter\ifx\csname urlstyle\endcsname\relax
  \providecommand{\doi}[1]{doi: #1}\else
  \providecommand{\doi}{doi: \begingroup \urlstyle{rm}\Url}\fi

\bibitem[Agrawal et~al.(2019)Agrawal, Desai, Wang, Chen, Jain, Johnson, Batra,
  Parikh, Lee, and Anderson]{agrawal2019nocaps}
H.~Agrawal, K.~Desai, Y.~Wang, X.~Chen, R.~Jain, M.~Johnson, D.~Batra,
  D.~Parikh, S.~Lee, and P.~Anderson.
\newblock nocaps: novel object captioning at scale.
\newblock In \emph{2019 IEEE/CVF International Conference on Computer Vision
  (ICCV)}, pages 8948--8957, Manhattan, New York, U.S., 2019. IEEE.
\newblock \doi{10.1109/ICCV.2019.00904}.
\newblock URL \url{http://doi.org/10.1109/ICCV.2019.00904}.

\bibitem[Ahsan et~al.(2021)Ahsan, Bhatt, Shah, and Bhalla]{ahsan2021multi}
H.~Ahsan, D.~Bhatt, K.~Shah, and N.~Bhalla.
\newblock Multi-modal image captioning for the visually impaired.
\newblock In \emph{Proceedings of the 2021 Conference of the North American
  Chapter of the Association for Computational Linguistics: Student Research
  Workshop}, pages 53--60, Stroudsburg, PA, USA, jun 2021. Association for
  Computational Linguistics.
\newblock \doi{10.18653/v1/2021.naacl-srw.8}.
\newblock URL \url{http://doi.org/10.18653/v1/2021.naacl-srw.8}.

\bibitem[Al~Sobbahi and Tekli(2022{\natexlab{a}})]{al2022comparing}
R.~Al~Sobbahi and J.~Tekli.
\newblock Comparing deep learning models for low-light natural scene image
  enhancement and their impact on object detection and classification:
  Overview, empirical evaluation, and challenges.
\newblock \emph{Signal Processing: Image Communication}, page 116848,
  2022{\natexlab{a}}.

\bibitem[Al~Sobbahi and Tekli(2022{\natexlab{b}})]{al2022low}
R.~Al~Sobbahi and J.~Tekli.
\newblock Low-light image enhancement using image-to-frequency filter learning.
\newblock In \emph{Image Analysis and Processing--ICIAP 2022: 21st
  International Conference, Lecce, Italy, May 23--27, 2022, Proceedings, Part
  II}, pages 693--705. Springer, 2022{\natexlab{b}}.

\bibitem[Alayrac et~al.(2022)Alayrac, Donahue, Luc, Miech, Barr, Hasson, Lenc,
  Mensch, Millican, Reynolds, et~al.]{alayrac2022flamingo}
J.-B. Alayrac, J.~Donahue, P.~Luc, A.~Miech, I.~Barr, Y.~Hasson, K.~Lenc,
  A.~Mensch, K.~Millican, M.~Reynolds, et~al.
\newblock Flamingo: a visual language model for few-shot learning.
\newblock \emph{Advances in Neural Information Processing Systems},
  35:\penalty0 23716--23736, 2022.

\bibitem[Anderson et~al.(2016)Anderson, Fernando, Johnson, and
  Gould]{anderson2016spice}
P.~Anderson, B.~Fernando, M.~Johnson, and S.~Gould.
\newblock Spice: Semantic propositional image caption evaluation.
\newblock In B.~Leibe, J.~Matas, N.~Sebe, and M.~Welling, editors,
  \emph{Computer Vision -- ECCV 2016}, pages 382--398, Manhattan, New York,
  USA, 2016. Springer International Publishing.
\newblock ISBN 978-3-319-46454-1.
\newblock \doi{10.1007/978-3-319-46454-1_24}.
\newblock URL \url{https://doi.org/10.1007/978-3-319-46454-1_24}.

\bibitem[Anderson et~al.(2018)Anderson, He, Buehler, Teney, Johnson, Gould, and
  Zhang]{anderson2018bottom}
P.~Anderson, X.~He, C.~Buehler, D.~Teney, M.~Johnson, S.~Gould, and L.~Zhang.
\newblock Bottom-up and top-down attention for image captioning and visual
  question answering.
\newblock In \emph{2018 IEEE/CVF Conference on Computer Vision and Pattern
  Recognition (CVPR)}, pages 6077--6086, Los Alamitos, CA, USA, jun 2018. IEEE
  Computer Society.
\newblock \doi{10.1109/CVPR.2018.00636}.
\newblock URL
  \url{https://doi.ieeecomputersociety.org/10.1109/CVPR.2018.00636}.

\bibitem[Aneja et~al.(2018)Aneja, Deshpande, and
  Schwing]{aneja2018convolutional}
J.~Aneja, A.~Deshpande, and A.~G. Schwing.
\newblock Convolutional image captioning.
\newblock In \emph{Proceedings of the IEEE conference on computer vision and
  pattern recognition}, pages 5561--5570, Manhattan, New York, U.S., June 2018.
  IEEE.
\newblock \doi{10.1109/CVPR.2018.00583}.
\newblock URL \url{http://doi.org/10.1109/CVPR.2018.00583}.

\bibitem[Ayesha et~al.(2021)Ayesha, Iqbal, Tariq, Abrar, Sanaullah, Abbas,
  Rehman, Niazi, and Hussain]{ayesha2021automatic}
H.~Ayesha, S.~Iqbal, M.~Tariq, M.~Abrar, M.~Sanaullah, I.~Abbas, A.~Rehman,
  M.~F.~K. Niazi, and S.~Hussain.
\newblock Automatic medical image interpretation: State of the art and future
  directions.
\newblock \emph{Pattern Recognition}, 114, June 2021.
\newblock ISSN 0031-3203.
\newblock \doi{10.1016/j.patcog.2021.107856}.
\newblock URL \url{https://doi.org/10.1016/j.patcog.2021.107856}.

\bibitem[Bahdanau et~al.(2015)Bahdanau, Cho, and Bengio]{bahdanau2014neural}
D.~Bahdanau, K.~Cho, and Y.~Bengio.
\newblock Neural machine translation by jointly learning to align and
  translate.
\newblock In Y.~Bengio and Y.~LeCun, editors, \emph{3rd International
  Conference on Learning Representations, {ICLR} 2015, San Diego, CA, USA, May
  7-9, 2015, Conference Track Proceedings}, 2015.

\bibitem[Banerjee and Lavie(2005)]{banerjee2005meteor}
S.~Banerjee and A.~Lavie.
\newblock Meteor: An automatic metric for mt evaluation with improved
  correlation with human judgments.
\newblock In J.~Goldstein, A.~Lavie, C.-Y. Lin, and C.~Voss, editors,
  \emph{Proceedings of the acl workshop on intrinsic and extrinsic evaluation
  measures for machine translation and/or summarization}, volume~29, pages
  65--72, Stroudsburg, PA, USA, 2005. Association for Computational
  Linguistics.
\newblock URL \url{https://aclanthology.org/W05-09}.

\bibitem[Barraco et~al.(2022)Barraco, Cornia, Cascianelli, Baraldi, and
  Cucchiara]{barraco2022unreasonable}
M.~Barraco, M.~Cornia, S.~Cascianelli, L.~Baraldi, and R.~Cucchiara.
\newblock The unreasonable effectiveness of clip features for image captioning:
  An experimental analysis.
\newblock In \emph{Proceedings of the IEEE/CVF Conference on Computer Vision
  and Pattern Recognition}, pages 4662--4670, Los Alamitos, CA, USA, jun 2022.
  IEEE Computer Society.
\newblock \doi{10.1109/CVPRW56347.2022.00512}.
\newblock URL
  \url{https://doi.ieeecomputersociety.org/10.1109/CVPRW56347.2022.00512}.

\bibitem[Bigham et~al.(2010)Bigham, Jayant, Ji, Little, Miller, Miller, Miller,
  Tatarowicz, White, White, et~al.]{bigham2010vizwiz}
J.~P. Bigham, C.~Jayant, H.~Ji, G.~Little, A.~Miller, R.~C. Miller, R.~Miller,
  A.~Tatarowicz, B.~White, S.~White, et~al.
\newblock Vizwiz: nearly real-time answers to visual questions.
\newblock In \emph{Proceedings of the 23nd Annual ACM Symposium on User
  Interface Software and Technology}, UIST '10, pages 333--342, New York, NY,
  USA, 2010. Association for Computing Machinery.
\newblock ISBN 9781450302715.
\newblock \doi{10.1145/1866029.1866080}.
\newblock URL \url{https://doi.org/10.1145/1866029.1866080}.

\bibitem[Bondy and Murty(1976)]{bondy1976graph}
J.~A. Bondy and U.~S.~R. Murty.
\newblock \emph{Graph theory with applications}, volume 290.
\newblock North-Holland, Amsterdam, Netherlands, 1976.
\newblock ISBN 0-444-19451-7.

\bibitem[Carion et~al.(2020)Carion, Massa, Synnaeve, Usunier, Kirillov, and
  Zagoruyko]{carion2020end}
N.~Carion, F.~Massa, G.~Synnaeve, N.~Usunier, A.~Kirillov, and S.~Zagoruyko.
\newblock End-to-end object detection with transformers.
\newblock In \emph{European conference on computer vision}, pages 213--229,
  Cham, 2020. Springer, Springer International Publishing.

\bibitem[Chen et~al.(2019)Chen, Mu, Xiao, Ye, Wu, and Ju]{chen2019improving}
C.~Chen, S.~Mu, W.~Xiao, Z.~Ye, L.~Wu, and Q.~Ju.
\newblock Improving image captioning with conditional generative adversarial
  nets.
\newblock \emph{33rd AAAI Conference on Artificial Intelligence, AAAI 2019,
  31st Innovative Applications of Artificial Intelligence Conference, IAAI 2019
  and the 9th AAAI Symposium on Educational Advances in Artificial
  Intelligence, EAAI 2019}, 33\penalty0 (01):\penalty0 8142--8150, July 2019.
\newblock ISSN 2159-5399.
\newblock \doi{10.1609/aaai.v33i01.33018142}.
\newblock URL \url{https://ojs.aaai.org/index.php/AAAI/article/view/4823}.

\bibitem[Chen et~al.(2021{\natexlab{a}})Chen, Wang, Guo, Xu, Deng, Liu, Ma, Xu,
  Xu, and Gao]{chen2021pre}
H.~Chen, Y.~Wang, T.~Guo, C.~Xu, Y.~Deng, Z.~Liu, S.~Ma, C.~Xu, C.~Xu, and
  W.~Gao.
\newblock Pre-trained image processing transformer.
\newblock In \emph{2021 IEEE/CVF Conference on Computer Vision and Pattern
  Recognition (CVPR)}, pages 12294--12305, Los Alamitos, CA, USA, jun
  2021{\natexlab{a}}. IEEE Computer Society.
\newblock \doi{10.1109/CVPR46437.2021.01212}.
\newblock URL
  \url{https://doi.ieeecomputersociety.org/10.1109/CVPR46437.2021.01212}.

\bibitem[Chen et~al.(2021{\natexlab{b}})Chen, Wang, Yang, and
  Li]{chen2021captioning}
H.~Chen, Y.~Wang, X.~Yang, and J.~Li.
\newblock Captioning transformer with scene graph guiding.
\newblock In \emph{2021 IEEE international conference on image processing
  (ICIP)}, pages 2538--2542. IEEE, 2021{\natexlab{b}}.
\newblock \doi{10.1109/ICIP42928.2021.9506193}.
\newblock URL \url{https://doi.org/10.1109/ICIP42928.2021.9506193}.

\bibitem[Chen et~al.(2020)Chen, Jin, Wang, and Wu]{chen2020say}
S.~Chen, Q.~Jin, P.~Wang, and Q.~Wu.
\newblock Say as you wish: Fine-grained control of image caption generation
  with abstract scene graphs.
\newblock In \emph{Proceedings of the IEEE/CVF Conference on Computer Vision
  and Pattern Recognition}, pages 9962--9971, Manhattan, New York, U.S., 2020.
  IEEE.
\newblock \doi{10.1109/CVPR42600.2020.00998}.
\newblock URL \url{http://doi.org/10.1109/CVPR42600.2020.00998}.

\bibitem[Cho et~al.(2014)Cho, Van~Merri{\"e}nboer, Gulcehre, Bahdanau,
  Bougares, Schwenk, and Bengio]{cho2014learning}
K.~Cho, B.~Van~Merri{\"e}nboer, C.~Gulcehre, D.~Bahdanau, F.~Bougares,
  H.~Schwenk, and Y.~Bengio.
\newblock Learning phrase representations using rnn encoder-decoder for
  statistical machine translation.
\newblock In \emph{Proceedings of the 2014 Conference on Empirical Methods in
  Natural Language Processing (EMNLP)}, page 1724–1734, Stroudsburg, PA, USA,
  2014. Association for Computational Linguistics.
\newblock \doi{10.3115/v1/d14-1179}.
\newblock URL \url{https://aclanthology.org/D14-1179}.

\bibitem[Chohan et~al.(2020)Chohan, Khan, Mahar, Hassan, Ghafoor, and
  Khan]{chohan2020image}
M.~Chohan, A.~Khan, M.~S. Mahar, S.~Hassan, A.~Ghafoor, and M.~Khan.
\newblock Image captioning using deep learning: A systematic literature review.
\newblock \emph{International Journal of Advanced Computer Science and
  Applications}, 11\penalty0 (5), 2020.
\newblock \doi{10.14569/IJACSA.2020.0110537}.
\newblock URL \url{http://dx.doi.org/10.14569/IJACSA.2020.0110537}.

\bibitem[Coppock et~al.(2020)Coppock, Dionne, Graham, Ganem, Zhao, Lin, Liu,
  and Wijaya]{coppock2020informativity}
E.~Coppock, D.~Dionne, N.~Graham, E.~Ganem, S.~Zhao, S.~Lin, W.~Liu, and
  D.~Wijaya.
\newblock Informativity in image captions vs. referring expressions.
\newblock In \emph{Proceedings of the Probability and Meaning Conference (PaM
  2020)}, pages 104--108, 2020.

\bibitem[Cornia et~al.(2019)Cornia, Baraldi, and Cucchiara]{cornia2019show}
M.~Cornia, L.~Baraldi, and R.~Cucchiara.
\newblock Show, control and tell: A framework for generating controllable and
  grounded captions.
\newblock In \emph{2019 IEEE/CVF Conference on Computer Vision and Pattern
  Recognition (CVPR)}, pages 8299--8308, Manhattan, New York, U.S., 2019. IEEE.
\newblock \doi{10.1109/CVPR.2019.00850}.
\newblock URL \url{http://doi.org/10.1109/CVPR.2019.00850}.

\bibitem[Cornia et~al.(2020)Cornia, Stefanini, Baraldi, and
  Cucchiara]{cornia2020meshed}
M.~Cornia, M.~Stefanini, L.~Baraldi, and R.~Cucchiara.
\newblock Meshed-memory transformer for image captioning.
\newblock In \emph{Proceedings of the IEEE/CVF Conference on Computer Vision
  and Pattern Recognition}, pages 10578--10587, Manhattan, New York, U.S.,
  2020. IEEE.
\newblock \doi{10.1109/CVPR42600.2020.01059}.
\newblock URL \url{http://doi.org/10.1109/CVPR42600.2020.01059}.

\bibitem[Dai et~al.(2017)Dai, Zhang, and Lin]{dai2017detecting}
B.~Dai, Y.~Zhang, and D.~Lin.
\newblock Detecting visual relationships with deep relational networks.
\newblock In \emph{Proceedings of the IEEE conference on computer vision and
  Pattern recognition}, pages 3298--3308, Manhattan, New York, U.S., 2017.
  IEEE.
\newblock \doi{10.1109/CVPR.2017.352}.
\newblock URL \url{http://doi.org/10.1109/CVPR.2017.352}.

\bibitem[Das et~al.(2017)Das, Agrawal, Zitnick, Parikh, and
  Batra]{das2017human}
A.~Das, H.~Agrawal, L.~Zitnick, D.~Parikh, and D.~Batra.
\newblock Human attention in visual question answering: Do humans and deep
  networks look at the same regions?
\newblock \emph{Computer Vision and Image Understanding}, 163:\penalty0
  90--100, Oct. 2017.
\newblock ISSN 1077-3142.
\newblock \doi{10.1016/j.cviu.2017.10.001}.
\newblock URL \url{http://doi.org/10.1016/j.cviu.2017.10.001}.
\newblock Language in Vision.

\bibitem[Deng et~al.(2009)Deng, Dong, Socher, Li, Li, and
  Fei-Fei]{deng2009imagenet}
J.~Deng, W.~Dong, R.~Socher, L.-J. Li, K.~Li, and L.~Fei-Fei.
\newblock Imagenet: A large-scale hierarchical image database.
\newblock In \emph{2009 IEEE Conference on Computer Vision and Pattern
  Recognition}, pages 248--255, Manhattan, New York, U.S., 2009. IEEE.
\newblock \doi{10.1109/CVPR.2009.5206848}.
\newblock URL \url{http://doi.org/10.1109/CVPR.2009.5206848}.

\bibitem[Devlin et~al.(2019)Devlin, Chang, Lee, and Toutanova]{devlin2018bert}
J.~Devlin, M.-W. Chang, K.~Lee, and K.~Toutanova.
\newblock Bert: Pre-training of deep bidirectional transformers for language
  understanding.
\newblock In \emph{Proceedings of the 2019 Conference of the North {A}merican
  Chapter of the Association for Computational Linguistics: Human Language
  Technologies, Volume 1 (Long and Short Papers)}, pages 4171--4186,
  Stroudsburg, PA, USA, 2019. Association for Computational Linguistics.
\newblock \doi{10.18653/v1/N19-1423}.
\newblock URL \url{http://doi.org/10.18653/v1/N19-1423}.

\bibitem[Dognin et~al.(2021)Dognin, Melnyk, Mroueh, Padhi, Rigotti, Ross,
  Schiff, Young, and Belgodere]{dognin2020image}
P.~Dognin, I.~Melnyk, Y.~Mroueh, I.~Padhi, M.~Rigotti, J.~Ross, Y.~Schiff,
  R.~A. Young, and B.~Belgodere.
\newblock Image captioning as an assistive technology: Lessons learned from
  vizwiz 2020 challenge.
\newblock \emph{arXiv preprint arXiv:2012.11696}, 2021.

\bibitem[Dosovitskiy et~al.(2021)Dosovitskiy, Beyer, Kolesnikov, Weissenborn,
  Zhai, Unterthiner, Dehghani, Minderer, Heigold, Gelly, Uszkoreit, and
  Houlsby]{dosovitskiy2020image}
A.~Dosovitskiy, L.~Beyer, A.~Kolesnikov, D.~Weissenborn, X.~Zhai,
  T.~Unterthiner, M.~Dehghani, M.~Minderer, G.~Heigold, S.~Gelly, J.~Uszkoreit,
  and N.~Houlsby.
\newblock An image is worth 16x16 words: Transformers for image recognition at
  scale.
\newblock In \emph{International Conference on Learning Representations},
  September 2021.
\newblock URL \url{https://openreview.net/forum?id=YicbFdNTTy}.

\bibitem[Elhagry and Kadaoui(2021)]{elhagry2021a}
A.~Elhagry and K.~Kadaoui.
\newblock A thorough review on recent deep learning methodologies for image
  captioning.
\newblock 2021.
\newblock \doi{10.48550/ARXIV.2107.13114}.
\newblock URL \url{https://arxiv.org/abs/2107.13114v1}.

\bibitem[Fang et~al.(2022)Fang, Wang, Hu, Liang, Gan, Wang, Yang, and
  Liu]{fang2022injecting}
Z.~Fang, J.~Wang, X.~Hu, L.~Liang, Z.~Gan, L.~Wang, Y.~Yang, and Z.~Liu.
\newblock Injecting semantic concepts into end-to-end image captioning.
\newblock In \emph{2022 IEEE/CVF Conference on Computer Vision and Pattern
  Recognition (CVPR)}, pages 17988--17998, Los Alamitos, CA, USA, jun 2022.
  IEEE Computer Society.
\newblock \doi{10.1109/CVPR52688.2022.01748}.
\newblock URL
  \url{https://doi.ieeecomputersociety.org/10.1109/CVPR52688.2022.01748}.

\bibitem[Feng et~al.(2019)Feng, Ma, Liu, and Luo]{feng2019unsupervised}
Y.~Feng, L.~Ma, W.~Liu, and J.~Luo.
\newblock Unsupervised image captioning.
\newblock In \emph{Proceedings of the IEEE/CVF Conference on Computer Vision
  and Pattern Recognition (CVPR)}, volume 2019-June, pages 4125--4134,
  Manhattan, New York, U.S., June 2019. IEEE.
\newblock \doi{10.1109/CVPR.2019.00425}.
\newblock URL \url{http://doi.org/10.1109/CVPR.2019.00425}.

\bibitem[Fukui et~al.(2016)Fukui, Park, Yang, Rohrbach, Darrell, and
  Rohrbach]{fukui2016multimodal}
A.~Fukui, D.~H. Park, D.~Yang, A.~Rohrbach, T.~Darrell, and M.~Rohrbach.
\newblock Multimodal compact bilinear pooling for visual question answering and
  visual grounding.
\newblock In \emph{Proceedings of the 2016 Conference on Empirical Methods in
  Natural Language Processing}, pages 457–--468, Stroudsburg, PA, USA, 2016.
  Association for Computational Linguistics.
\newblock \doi{10.18653/v1/D16-1044}.
\newblock URL \url{http://doi.org/10.18653/v1/D16-1044}.

\bibitem[Gan et~al.(2017{\natexlab{a}})Gan, Gan, He, Gao, and
  Deng]{gan2017stylenet}
C.~Gan, Z.~Gan, X.~He, J.~Gao, and L.~Deng.
\newblock Stylenet: Generating attractive visual captions with styles.
\newblock In \emph{2017 IEEE Conference on Computer Vision and Pattern
  Recognition (CVPR)}, pages 955--964, Los Alamitos, CA, USA, July
  2017{\natexlab{a}}. IEEE Computer Society.
\newblock \doi{10.1109/CVPR.2017.108}.
\newblock URL \url{https://doi.ieeecomputersociety.org/10.1109/CVPR.2017.108}.

\bibitem[Gan et~al.(2017{\natexlab{b}})Gan, Gan, He, Pu, Tran, Gao, Carin, and
  Deng]{gan2017semantic}
Z.~Gan, C.~Gan, X.~He, Y.~Pu, K.~Tran, J.~Gao, L.~Carin, and L.~Deng.
\newblock Semantic compositional networks for visual captioning.
\newblock In \emph{2017 IEEE Conference on Computer Vision and Pattern
  Recognition (CVPR)}, pages 1141--1150, Manhattan, New York, USA,
  2017{\natexlab{b}}. IEEE.
\newblock \doi{10.1109/CVPR.2017.127}.
\newblock URL \url{http://doi.org/10.1109/CVPR.2017.127}.

\bibitem[Gao et~al.(2018)Gao, Wang, and Wang]{gao2018image}
L.~Gao, B.~Wang, and W.~Wang.
\newblock Image captioning with scene-graph based semantic concepts.
\newblock In \emph{Proceedings of the 2018 10th International Conference on
  Machine Learning and Computing}, ICMLC 2018, pages 225--229, New York, NY,
  USA, 2018. Association for Computing Machinery.
\newblock ISBN 9781450363532.
\newblock \doi{10.1145/3195106.3195114}.
\newblock URL \url{https://doi.org/10.1145/3195106.3195114}.

\bibitem[Gehring et~al.(2017{\natexlab{a}})Gehring, Auli, Grangier, and
  Dauphin]{gehring2016convolutional}
J.~Gehring, M.~Auli, D.~Grangier, and Y.~N. Dauphin.
\newblock A convolutional encoder model for neural machine translation.
\newblock In \emph{Proceedings of the 55th Annual Meeting of the Association
  for Computational Linguistics (Volume 1: Long Papers)}, pages 123--135,
  Stroudsburg, PA, USA, July 2017{\natexlab{a}}. Association for Computational
  Linguistics.
\newblock \doi{10.18653/v1/P17-1012}.
\newblock URL \url{https://doi.org/10.18653/v1/P17-1012}.

\bibitem[Gehring et~al.(2017{\natexlab{b}})Gehring, Auli, Grangier, Yarats, and
  Dauphin]{gehring2017convolutional}
J.~Gehring, M.~Auli, D.~Grangier, D.~Yarats, and Y.~N. Dauphin.
\newblock Convolutional sequence to sequence learning.
\newblock In \emph{International Conference on Machine Learning}, pages
  1243--1252. PMLR, 2017{\natexlab{b}}.

\bibitem[Gers et~al.(2000)Gers, Schmidhuber, and Cummins]{Gers2000learning}
F.~A. Gers, J.~Schmidhuber, and F.~Cummins.
\newblock {Learning to Forget: Continual Prediction with LSTM}.
\newblock \emph{Neural Computation}, 12\penalty0 (10):\penalty0 2451--2471, 10
  2000.
\newblock ISSN 0899-7667.
\newblock \doi{10.1162/089976600300015015}.
\newblock URL \url{https://doi.org/10.1162/089976600300015015}.

\bibitem[Girshick(2015)]{girshick2015fast}
R.~Girshick.
\newblock Fast r-cnn.
\newblock In \emph{2015 IEEE International Conference on Computer Vision
  (ICCV)}, volume~1, pages 1440--1448, Los Alamitos, CA, USA, dec 2015. IEEE
  Computer Society.
\newblock \doi{10.1109/ICCV.2015.169}.
\newblock URL \url{https://doi.ieeecomputersociety.org/10.1109/ICCV.2015.169}.

\bibitem[Girshick et~al.(2014)Girshick, Donahue, Darrell, and
  Malik]{girshick2014rich}
R.~Girshick, J.~Donahue, T.~Darrell, and J.~Malik.
\newblock Rich feature hierarchies for accurate object detection and semantic
  segmentation.
\newblock In \emph{2014 IEEE Conference on Computer Vision and Pattern
  Recognition (CVPR)}, volume~1, pages 580--587, Los Alamitos, CA, USA, jun
  2014. IEEE Computer Society.
\newblock \doi{10.1109/CVPR.2014.81}.
\newblock URL \url{https://doi.ieeecomputersociety.org/10.1109/CVPR.2014.81}.

\bibitem[Goodfellow et~al.(2016)Goodfellow, Bengio, and
  Courville]{goodfellow2017deep}
I.~Goodfellow, Y.~Bengio, and A.~Courville.
\newblock \emph{Deep learning (adaptive computation and machine learning
  series)}.
\newblock The MIT Press, Cambridge, MA, USA, 2016.
\newblock ISBN 978-0262035613.

\bibitem[Google(2021)]{openimagesv6}
Google.
\newblock Open images dataset v6 + extensions, August 2021.
\newblock URL \url{https://storage.googleapis.com/openimages/web/index.html}.

\bibitem[Greff et~al.(2017)Greff, Srivastava, Koutnik, Steunebrink, and
  Schmidhuber]{Greff_2017}
K.~Greff, R.~K. Srivastava, J.~Koutnik, B.~R. Steunebrink, and J.~Schmidhuber.
\newblock {LSTM}: A search space odyssey.
\newblock \emph{{IEEE} Transactions on Neural Networks and Learning Systems},
  28\penalty0 (10):\penalty0 2222--2232, oct 2017.
\newblock ISSN 2162-2388.
\newblock \doi{10.1109/tnnls.2016.2582924}.
\newblock URL \url{https://doi.org/10.1109/TNNLS.2016.2582924}.

\bibitem[Gruber and Jockisch(2020)]{gruber2020gru}
N.~Gruber and A.~Jockisch.
\newblock Are gru cells more specific and lstm cells more sensitive in motive
  classification of text?
\newblock \emph{Frontiers in artificial intelligence}, 3:\penalty0 40, 2020.
\newblock ISSN https://portal.issn.org/resource/ISSN/26248212.
\newblock \doi{10.3389/frai.2020.00040}.
\newblock URL \url{https://dx.doi.org/10.3389/frai.2020.00040}.

\bibitem[Gu et~al.(2018{\natexlab{a}})Gu, Cai, Wang, and Chen]{gu2018stack}
J.~Gu, J.~Cai, G.~Wang, and T.~Chen.
\newblock Stack-captioning: Coarse-to-fine learning for image captioning.
\newblock In \emph{32nd AAAI Conference on Artificial Intelligence, AAAI 2018},
  pages 6837--6845, Palo Alto, California USA, 2018{\natexlab{a}}. AAAI Press.

\bibitem[Gu et~al.(2018{\natexlab{b}})Gu, Joty, Cai, and Wang]{gu2018unpaired}
J.~Gu, S.~Joty, J.~Cai, and G.~Wang.
\newblock Unpaired image captioning by language pivoting.
\newblock In V.~Ferrari, M.~Hebert, C.~Sminchisescu, and Y.~Weiss, editors,
  \emph{Computer Vision -- ECCV 2018}, pages 519--535, Manhattan, New York,
  USA, 2018{\natexlab{b}}. Springer International Publishing.
\newblock ISBN 978-3-030-01246-5.
\newblock \doi{10.1007/978-3-030-01246-5_31}.
\newblock URL \url{http://doi.org/10.1007/978-3-030-01246-5_31}.

\bibitem[Gu et~al.(2019{\natexlab{a}})Gu, Joty, Cai, Zhao, Yang, and
  Wang]{gu2019unpaired}
J.~Gu, S.~Joty, J.~Cai, H.~Zhao, X.~Yang, and G.~Wang.
\newblock Unpaired image captioning via scene graph alignments.
\newblock In \emph{Proceedings of the IEEE/CVF International Conference on
  Computer Vision}, volume 2019-October, pages 10323--10332, Manhattan, New
  York, U.S., 2019{\natexlab{a}}. IEEE.
\newblock \doi{10.1109/ICCV.2019.01042}.
\newblock URL \url{http://doi.org/10.1109/ICCV.2019.01042}.

\bibitem[Gu et~al.(2019{\natexlab{b}})Gu, Zhao, Lin, Li, Cai, and
  Ling]{gu2019scene}
J.~Gu, H.~Zhao, Z.~Lin, S.~Li, J.~Cai, and M.~Ling.
\newblock Scene graph generation with external knowledge and image
  reconstruction.
\newblock In \emph{2019 IEEE/CVF Conference on Computer Vision and Pattern
  Recognition (CVPR)}, pages 1969--1978, Manhattan, New York, U.S.,
  2019{\natexlab{b}}. IEEE.
\newblock \doi{10.1109/CVPR.2019.00207}.
\newblock URL \url{http://doi.org/10.1109/CVPR.2019.00207}.

\bibitem[Guo et~al.(2019)Guo, Liu, Yao, Li, and Lu]{guo2019mscap}
L.~Guo, J.~Liu, P.~Yao, J.~Li, and H.~Lu.
\newblock Mscap: Multi-style image captioning with unpaired stylized text.
\newblock In \emph{Proceedings of the IEEE/CVF Conference on Computer Vision
  and Pattern Recognition}, volume 2019-June, pages 4204--4213, Manhattan, New
  York, U.S., 2019. IEEE.
\newblock \doi{10.1109/CVPR.2019.00433}.
\newblock URL \url{http://doi.org/10.1109/CVPR.2019.00433}.

\bibitem[Gurari et~al.(2020)Gurari, Zhao, Zhang, and
  Bhattacharya]{gurari2020captioning}
D.~Gurari, Y.~Zhao, M.~Zhang, and N.~Bhattacharya.
\newblock Captioning images taken by people who are blind.
\newblock In A.~Vedaldi, H.~Bischof, T.~Brox, and J.-M. Frahm, editors,
  \emph{Computer Vision -- ECCV 2020}, pages 417--434, Manhattan, New York,
  USA, 2020. Springer International Publishing.
\newblock ISBN 978-3-030-58520-4.
\newblock \doi{10.1007/978-3-030-58520-4_25}.
\newblock URL \url{http://doi.org/10.1007/978-3-030-58520-4_25}.

\bibitem[He et~al.(2016)He, Zhang, Ren, and Sun]{he2016deep}
K.~He, X.~Zhang, S.~Ren, and J.~Sun.
\newblock Deep residual learning for image recognition.
\newblock In \emph{2016 IEEE Conference on Computer Vision and Pattern
  Recognition (CVPR)}, pages 770--778, Manhattan, New York, U.S., 2016. IEEE.
\newblock \doi{10.1109/CVPR.2016.90}.
\newblock URL \url{http://doi.org/10.1109/CVPR.2016.90}.

\bibitem[He et~al.(2020)He, Liao, Tavakoli, Yang, Rosenhahn, and
  Pugeault]{he2020image}
S.~He, W.~Liao, H.~R. Tavakoli, M.~Yang, B.~Rosenhahn, and N.~Pugeault.
\newblock Image captioning through image transformer.
\newblock In \emph{Proceedings of the Asian Conference on Computer Vision
  (ACCV)}, page 153–169, Berlin, Heidelberg, November 2020. Springer-Verlag.
\newblock ISBN 978-3-030-69537-8.
\newblock \doi{10.1007/978-3-030-69538-5_10}.
\newblock URL \url{https://doi.org/10.1007/978-3-030-69538-5_10}.

\bibitem[Herdade et~al.(2019)Herdade, Kappeler, Boakye, and
  Soares]{herdade2019image}
S.~Herdade, A.~Kappeler, K.~Boakye, and J.~Soares.
\newblock Image captioning: Transforming objects into words.
\newblock In H.~Wallach, H.~Larochelle, A.~Beygelzimer, F.~d\textquotesingle
  Alch\'{e}-Buc, E.~Fox, and R.~Garnett, editors, \emph{Advances in Neural
  Information Processing Systems}, volume~32, pages 11137--11147, Red Hook, New
  York, USA, 2019. Curran Associates, Inc.
\newblock URL
  \url{https://proceedings.neurips.cc/paper/2019/file/680390c55bbd9ce416d1d69a9ab4760d-Paper.pdf}.

\bibitem[Hochreiter and Schmidhuber(1997)]{hochreiter1997long}
S.~Hochreiter and J.~Schmidhuber.
\newblock Long short-term memory.
\newblock \emph{Neural computation}, 9\penalty0 (8):\penalty0 1735--1780, nov
  1997.
\newblock ISSN 0899-7667.
\newblock \doi{10.1162/neco.1997.9.8.1735}.
\newblock URL \url{https://doi.org/10.1162/neco.1997.9.8.1735}.

\bibitem[Hochreiter et~al.(2001)Hochreiter, Bengio, Frasconi, Schmidhuber,
  et~al.]{hochreiter2001gradient}
S.~Hochreiter, Y.~Bengio, P.~Frasconi, J.~Schmidhuber, et~al.
\newblock \emph{Gradient flow in recurrent nets: the difficulty of learning
  long-term dependencies}.
\newblock Wiley-IEEE Press, Piscataway, New Jersey, USA, 2001.
\newblock ISBN 978-0-780-35369-5.
\newblock \doi{10.1109/9780470544037.ch14}.

\bibitem[Hossain et~al.(2019)Hossain, Sohel, Shiratuddin, and
  Laga]{hossain2019comprehensive}
M.~Z. Hossain, F.~Sohel, M.~F. Shiratuddin, and H.~Laga.
\newblock A comprehensive survey of deep learning for image captioning.
\newblock \emph{ACM Comput. Surv.}, 51\penalty0 (6), Feb. 2019.
\newblock ISSN 0360-0300.
\newblock \doi{10.1145/3295748}.
\newblock URL \url{https://doi.org/10.1145/3295748}.

\bibitem[Hu et~al.(2018)Hu, Gu, Zhang, Dai, and Wei]{hu2018relation}
H.~Hu, J.~Gu, Z.~Zhang, J.~Dai, and Y.~Wei.
\newblock Relation networks for object detection.
\newblock In \emph{Proceedings of the IEEE conference on computer vision and
  pattern recognition}, pages 3588--3597, Manhattan, New York, USA, 2018. IEEE.
\newblock \doi{10.1109/CVPR.2018.00378}.
\newblock URL \url{http://doi.org/10.1109/CVPR.2018.00378}.

\bibitem[Hu et~al.(2022{\natexlab{a}})Hu, Cavicchioli, and
  Capotondi]{hu2022expansionnet}
J.~C. Hu, R.~Cavicchioli, and A.~Capotondi.
\newblock Expansionnet v2: Block static expansion in fast end to end training
  for image captioning.
\newblock \emph{arXiv preprint arXiv:2208.06551}, 2022{\natexlab{a}}.

\bibitem[Hu et~al.(2021)Hu, Yin, Lin, Zhang, Gao, Wang, and Liu]{hu2021vivo}
X.~Hu, X.~Yin, K.~Lin, L.~Zhang, J.~Gao, L.~Wang, and Z.~Liu.
\newblock Vivo: Visual vocabulary pre-training for novel object captioning.
\newblock \emph{Proceedings of the AAAI Conference on Artificial Intelligence},
  35\penalty0 (2):\penalty0 1575--1583, May 2021.
\newblock \doi{10.1609/aaai.v35i2.16249}.
\newblock URL \url{https://ojs.aaai.org/index.php/AAAI/article/view/16249}.

\bibitem[Hu et~al.(2022{\natexlab{b}})Hu, Gan, Wang, Yang, Liu, Lu, and
  Wang]{hu2022scaling}
X.~Hu, Z.~Gan, J.~Wang, Z.~Yang, Z.~Liu, Y.~Lu, and L.~Wang.
\newblock Scaling up vision-language pretraining for image captioning.
\newblock In \emph{2022 IEEE/CVF Conference on Computer Vision and Pattern
  Recognition (CVPR)}, pages 17959--17968, Los Alamitos, CA, USA, jun
  2022{\natexlab{b}}. IEEE Computer Society.
\newblock \doi{10.1109/CVPR52688.2022.01745}.
\newblock URL
  \url{https://doi.ieeecomputersociety.org/10.1109/CVPR52688.2022.01745}.

\bibitem[Huang et~al.(2017)Huang, Liu, Van Der~Maaten, and
  Weinberger]{huang2017densely}
G.~Huang, Z.~Liu, L.~Van Der~Maaten, and K.~Q. Weinberger.
\newblock Densely connected convolutional networks.
\newblock In \emph{Proceedings of the IEEE Conference on Computer Vision and
  Pattern Recognition (CVPR)}, pages 4700--4708, Manhattan, New York, U.S.,
  2017. IEEE Computer Society.
\newblock \doi{10.1109/CVPR.2017.243}.
\newblock URL \url{http://doi.org/10.1109/CVPR.2017.243}.

\bibitem[Huang et~al.(2021)Huang, Wu, and Worring]{huang2021contextualized}
J.-H. Huang, T.-W. Wu, and M.~Worring.
\newblock Contextualized keyword representations for multi-modal retinal image
  captioning.
\newblock In \emph{Proceedings of the 2021 International Conference on
  Multimedia Retrieval}, ICMR '21, page 645–652, New York, NY, USA, 2021.
  Association for Computing Machinery.
\newblock ISBN 9781450384636.
\newblock \doi{10.1145/3460426.3463667}.
\newblock URL \url{https://doi.org/10.1145/3460426.3463667}.

\bibitem[Huang et~al.(2019)Huang, Wang, Chen, and Wei]{huang2019attention}
L.~Huang, W.~Wang, J.~Chen, and X.-Y. Wei.
\newblock Attention on attention for image captioning.
\newblock In \emph{Proceedings of the IEEE/CVF International Conference on
  Computer Vision}, volume 2019-October, pages 4634--4643, Manhattan, New York,
  U.S., 2019. IEEE.
\newblock \doi{10.1109/ICCV.2019.00473}.
\newblock URL \url{http://doi.org/10.1109/ICCV.2019.00473}.

\bibitem[Huang et~al.(2023)Huang, Dong, Wang, Hao, Singhal, Ma, Lv, Cui,
  Mohammed, Liu, et~al.]{huang2023language}
S.~Huang, L.~Dong, W.~Wang, Y.~Hao, S.~Singhal, S.~Ma, T.~Lv, L.~Cui, O.~K.
  Mohammed, Q.~Liu, et~al.
\newblock Language is not all you need: Aligning perception with language
  models.
\newblock \emph{arXiv preprint arXiv:2302.14045}, 2023.

\bibitem[Jiang et~al.(2018)Jiang, Ma, Jiang, Liu, and
  Zhang]{jiang2018recurrent}
W.~Jiang, L.~Ma, Y.-G. Jiang, W.~Liu, and T.~Zhang.
\newblock Recurrent fusion network for image captioning.
\newblock In \emph{Computer Vision -- ECCV 2018)}, pages 510--526, Manhattan,
  New York, USA, 2018. Springer International Publishing.
\newblock \doi{10.1007/978-3-030-01216-8_31}.
\newblock URL \url{http://doi.org/10.1007/978-3-030-01216-8_31}.

\bibitem[Johnson et~al.(2016)Johnson, Karpathy, and
  Fei-Fei]{johnson2016densecap}
J.~Johnson, A.~Karpathy, and L.~Fei-Fei.
\newblock Densecap: Fully convolutional localization networks for dense
  captioning.
\newblock In \emph{2016 IEEE Conference on Computer Vision and Pattern
  Recognition (CVPR)}, pages 4565--4574, Manhattan, New York, U.S., 2016. IEEE.
\newblock \doi{10.1109/CVPR.2016.494}.
\newblock URL \url{http://doi.org/10.1109/CVPR.2016.494}.

\bibitem[Jordan(1997)]{osti_6910294}
M.~I. Jordan.
\newblock \emph{Chapter 25 - Serial Order: A Parallel Distributed Processing
  Approach}, volume 121 of \emph{Advances in Psychology}, pages 471--495.
\newblock North-Holland, Amsterdam, Netherlands, 1997.
\newblock \doi{10.1016/S0166-4115(97)80111-2}.
\newblock URL \url{https://doi.org/10.1016/S0166-4115(97)80111-2}.

\bibitem[Karim(2019)]{attention_mechanism_figure}
R.~Karim.
\newblock Attn: Illustrated attention. attention in gifs and how it is used
  in…, Jan 2019.
\newblock URL
  \url{https://towardsdatascience.com/attn-illustrated-attention-5ec4ad276ee3}.

\bibitem[Karpathy and Fei-Fei(2017)]{karpathy2015deep}
A.~Karpathy and L.~Fei-Fei.
\newblock Deep visual-semantic alignments for generating image descriptions.
\newblock \emph{IEEE Transactions on Pattern Analysis and Machine
  Intelligence}, 39:\penalty0 664--676, Apr. 2017.
\newblock ISSN 01628828.
\newblock \doi{10.1109/TPAMI.2016.2598339}.
\newblock URL \url{http://doi.org/10.1109/TPAMI.2016.2598339}.

\bibitem[Kipf and Welling(2017)]{kipf2016semi}
T.~N. Kipf and M.~Welling.
\newblock Semi-supervised classification with graph convolutional networks.
\newblock In \emph{International Conference on Learning Representations
  (ICLR)}, 2017.

\bibitem[Krishna et~al.(2017)Krishna, Zhu, Groth, Johnson, Hata, Kravitz, Chen,
  Kalantidis, Li, Shamma, et~al.]{krishna2017visual}
R.~Krishna, Y.~Zhu, O.~Groth, J.~Johnson, K.~Hata, J.~Kravitz, S.~Chen,
  Y.~Kalantidis, L.-J. Li, D.~A. Shamma, et~al.
\newblock Visual genome: Connecting language and vision using crowdsourced
  dense image annotations.
\newblock \emph{International journal of computer vision}, 123\penalty0
  (1):\penalty0 32--73, May 2017.
\newblock ISSN 15731405.
\newblock \doi{10.1007/s11263-016-0981-7}.

\bibitem[Kuznetsova et~al.(2020)Kuznetsova, Rom, Alldrin, Uijlings, Krasin,
  Pont-Tuset, Kamali, Popov, Malloci, Kolesnikov, Duerig, and
  Ferrari]{OpenImagesv4}
A.~Kuznetsova, H.~Rom, N.~Alldrin, J.~Uijlings, I.~Krasin, J.~Pont-Tuset,
  S.~Kamali, S.~Popov, M.~Malloci, A.~Kolesnikov, T.~Duerig, and V.~Ferrari.
\newblock The open images dataset v4: Unified image classification, object
  detection, and visual relationship detection at scale.
\newblock \emph{International Journal of Computer Vision}, 128:\penalty0
  1956--1981, July 2020.
\newblock ISSN 15731405.
\newblock \doi{10.1007/s11263-020-01316-z}.
\newblock URL \url{http://doi.org/10.1007/s11263-020-01316-z}.

\bibitem[Lee et~al.(2019)Lee, Palangi, Chen, Hu, and Gao]{lee2019learning}
K.-H. Lee, H.~Palangi, X.~Chen, H.~Hu, and J.~Gao.
\newblock Learning visual relation priors for image-text matching and image
  captioning with neural scene graph generators.
\newblock \emph{arXiv preprint arXiv:1909.09953}, 2019.
\newblock URL \url{https://arxiv.org/abs/1909.09953}.

\bibitem[Li et~al.(2022{\natexlab{a}})Li, Xu, Tian, Wang, Yan, Bi, Ye, Chen,
  Xu, Cao, et~al.]{li2022mplug}
C.~Li, H.~Xu, J.~Tian, W.~Wang, M.~Yan, B.~Bi, J.~Ye, H.~Chen, G.~Xu, Z.~Cao,
  et~al.
\newblock mplug: Effective and efficient vision-language learning by
  cross-modal skip-connections.
\newblock \emph{arXiv preprint arXiv:2205.12005}, 2022{\natexlab{a}}.

\bibitem[Li et~al.(2019)Li, Zhu, Liu, and Yang]{li2019entangled}
G.~Li, L.~Zhu, P.~Liu, and Y.~Yang.
\newblock Entangled transformer for image captioning.
\newblock In \emph{Proceedings of the IEEE/CVF international conference on
  computer vision}, pages 8928--8937, Los Alamitos, CA, USA, nov 2019. IEEE
  Computer Society.
\newblock \doi{10.1109/ICCV.2019.00902}.
\newblock URL
  \url{https://doi.ieeecomputersociety.org/10.1109/ICCV.2019.00902}.

\bibitem[Li et~al.(2023)Li, Li, Savarese, and Hoi]{li2023blip}
J.~Li, D.~Li, S.~Savarese, and S.~Hoi.
\newblock Blip-2: Bootstrapping language-image pre-training with frozen image
  encoders and large language models.
\newblock \emph{arXiv preprint arXiv:2301.12597}, 2023.

\bibitem[Li et~al.(2020{\natexlab{a}})Li, Qu, Song, Wang, and
  Xue]{li2020traffic}
W.~Li, Z.~Qu, H.~Song, P.~Wang, and B.~Xue.
\newblock The traffic scene understanding and prediction based on image
  captioning.
\newblock \emph{IEEE Access}, 9:\penalty0 1420--1427, Dec. 2020{\natexlab{a}}.
\newblock ISSN 2169-3536.
\newblock \doi{10.1109/ACCESS.2020.3047091}.
\newblock URL \url{http://doi.org/10.1109/ACCESS.2020.3047091}.

\bibitem[Li and Jiang(2019)]{li2019know}
X.~Li and S.~Jiang.
\newblock Know more say less: Image captioning based on scene graphs.
\newblock \emph{IEEE Transactions on Multimedia}, 21\penalty0 (8):\penalty0
  2117--2130, Jan. 2019.
\newblock ISSN 1941-0077.
\newblock \doi{10.1109/TMM.2019.2896516}.
\newblock URL \url{http://doi.org/10.1109/TMM.2019.2896516}.

\bibitem[Li et~al.(2020{\natexlab{b}})Li, Yin, Li, Zhang, Hu, Zhang, Wang, Hu,
  Dong, Wei, et~al.]{li2020oscar}
X.~Li, X.~Yin, C.~Li, P.~Zhang, X.~Hu, L.~Zhang, L.~Wang, H.~Hu, L.~Dong,
  F.~Wei, et~al.
\newblock Oscar: Object-semantics aligned pre-training for vision-language
  tasks.
\newblock In A.~Vedaldi, H.~Bischof, T.~Brox, and J.-M. Frahm, editors,
  \emph{Computer Vision -- ECCV 2020}, pages 121--137, Manhattan, New York,
  USA, 2020{\natexlab{b}}. Springer, Springer International Publishing.
\newblock ISBN 978-3-030-58577-8.
\newblock \doi{10.1007/978-3-030-58577-8_8}.
\newblock URL \url{https://doi.org/10.1007/978-3-030-58577-8_8}.

\bibitem[Li and Liang(2021)]{li2021prefix}
X.~L. Li and P.~Liang.
\newblock Prefix-tuning: Optimizing continuous prompts for generation.
\newblock \emph{arXiv preprint arXiv:2101.00190}, 2021.
\newblock \doi{10.48550/ARXIV.2101.00190}.
\newblock URL \url{https://arxiv.org/abs/2101.00190}.

\bibitem[Li et~al.(2017)Li, Ouyang, Zhou, Wang, and Wang]{li2017scene}
Y.~Li, W.~Ouyang, B.~Zhou, K.~Wang, and X.~Wang.
\newblock Scene graph generation from objects, phrases and region captions.
\newblock In \emph{Proceedings of the IEEE international conference on computer
  vision}, pages 1270--1279, Manhattan, New York, U.S., 2017. IEEE.
\newblock \doi{10.1109/ICCV.2017.142}.
\newblock URL \url{http://doi.org/10.1109/ICCV.2017.142}.

\bibitem[Li et~al.(2022{\natexlab{b}})Li, Pan, Yao, and
  Mei]{li2022comprehending}
Y.~Li, Y.~Pan, T.~Yao, and T.~Mei.
\newblock Comprehending and ordering semantics for image captioning.
\newblock In \emph{2022 IEEE/CVF Conference on Computer Vision and Pattern
  Recognition (CVPR)}, pages 17969--17978, Los Alamitos, CA, USA, jun
  2022{\natexlab{b}}. IEEE Computer Society.
\newblock \doi{10.1109/CVPR52688.2022.01746}.
\newblock URL
  \url{https://doi.ieeecomputersociety.org/10.1109/CVPR52688.2022.01746}.

\bibitem[Lin(2004)]{lin2004rouge}
C.-Y. Lin.
\newblock Rouge: A package for automatic evaluation of summaries.
\newblock In \emph{Proceedings of the workshop on text summarization branches
  out (WAS 2004)}, pages 74--81, Stroudsburg, PA, USA, 2004. Association for
  Computational Linguistics.
\newblock URL \url{https://aclanthology.org/W04-1000}.

\bibitem[Lin(2021)]{cocoevaltool}
T.-Y. Lin.
\newblock Microsoft coco caption evaluation, August 2021.
\newblock URL \url{https://github.com/tylin/coco-caption}.

\bibitem[Lin et~al.(2015)Lin, Maire, Belongie, Hays, Perona, Ramanan,
  Doll{\'a}r, and Zitnick]{lin2014microsoft}
T.-Y. Lin, M.~Maire, S.~Belongie, J.~Hays, P.~Perona, D.~Ramanan,
  P.~Doll{\'a}r, and C.~L. Zitnick.
\newblock Microsoft coco: Common objects in context.
\newblock In D.~Fleet, T.~Pajdla, B.~Schiele, and T.~Tuytelaars, editors,
  \emph{Computer Vision -- ECCV 2014}, pages 740--755, Manhattan, New York,
  USA, 2015. Springer, Springer International Publishing.
\newblock ISBN 978-3-319-10602-1.

\bibitem[Liu et~al.(2020)Liu, Ren, Liu, Lei, and Sun]{liu2020exploring}
F.~Liu, X.~Ren, Y.~Liu, K.~Lei, and X.~Sun.
\newblock Exploring and distilling cross-modal information for image
  captioning.
\newblock In \emph{IJCAI International Joint Conference on Artificial
  Intelligence}, volume 2019-August, pages 5095--5101, California, USA, 7 2020.
  International Joint Conferences on Artificial Intelligence Organization.
\newblock \doi{10.24963/ijcai.2019/708}.
\newblock URL \url{https://doi.org/10.24963/ijcai.2019/708}.

\bibitem[Liu et~al.(2017)Liu, Zhu, Ye, Guadarrama, and Murphy]{liu2017improved}
S.~Liu, Z.~Zhu, N.~Ye, S.~Guadarrama, and K.~Murphy.
\newblock Improved image captioning via policy gradient optimization of spider.
\newblock In \emph{2017 IEEE International Conference on Computer Vision
  (ICCV)}, pages 873--881, Manhattan, New York, U.S., 2017. IEEE.
\newblock \doi{10.1109/ICCV.2017.100}.
\newblock URL \url{http://doi.org/10.1109/ICCV.2017.100}.

\bibitem[Liu et~al.(2023)Liu, Fan, Johns, Yu, Xiao, and
  Anandkumar]{liu2023prismer}
S.~Liu, L.~Fan, E.~Johns, Z.~Yu, C.~Xiao, and A.~Anandkumar.
\newblock Prismer: A vision-language model with an ensemble of experts.
\newblock \emph{arXiv preprint arXiv:2303.02506}, 2023.

\bibitem[Liu et~al.(2021{\natexlab{a}})Liu, Chen, Guo, Zhu, and
  Liu]{liu2021cptr}
W.~Liu, S.~Chen, L.~Guo, X.~Zhu, and J.~Liu.
\newblock Cptr: Full transformer network for image captioning.
\newblock \emph{arXiv preprint arXiv:2101.10804}, 2021{\natexlab{a}}.
\newblock \doi{10.48550/ARXIV.2101.10804}.
\newblock URL \url{https://arxiv.org/abs/2101.10804}.

\bibitem[Liu et~al.(2018)Liu, Li, Shao, Chen, and Wang]{liu2018show}
X.~Liu, H.~Li, J.~Shao, D.~Chen, and X.~Wang.
\newblock Show, tell and discriminate: Image captioning by self-retrieval with
  partially labeled data.
\newblock In \emph{Computer Vision -- ECCV 2018}, pages 353--369, Manhattan,
  New York, USA, September 2018. Springer International Publishing.
\newblock ISBN 978-3-030-01267-0.
\newblock \doi{10.1007/978-3-030-01267-0_21}.
\newblock URL \url{http://doi.org/10.1007/978-3-030-01267-0_21}.

\bibitem[Liu et~al.(2021{\natexlab{b}})Liu, Lin, Cao, Hu, Wei, Zhang, Lin, and
  Guo]{liu2021swin}
Z.~Liu, Y.~Lin, Y.~Cao, H.~Hu, Y.~Wei, Z.~Zhang, S.~Lin, and B.~Guo.
\newblock Swin transformer: Hierarchical vision transformer using shifted
  windows.
\newblock In \emph{Proceedings of the IEEE/CVF International Conference on
  Computer Vision}, pages 10012--10022, Los Alamitos, CA, USA, oct
  2021{\natexlab{b}}. IEEE Computer Society.
\newblock \doi{10.1109/ICCV48922.2021.00986}.
\newblock URL
  \url{https://doi.ieeecomputersociety.org/10.1109/ICCV48922.2021.00986}.

\bibitem[Luo et~al.(2019)Luo, Hsu, Wen, and Ye]{luo2019visual}
R.~C. Luo, Y.-T. Hsu, Y.-C. Wen, and H.-J. Ye.
\newblock Visual image caption generation for service robotics and industrial
  applications.
\newblock In \emph{2019 IEEE International Conference on Industrial Cyber
  Physical Systems (ICPS)}, pages 827--832, Manhattan, New York, U.S., 2019.
  IEEE.
\newblock \doi{10.1109/ICPHYS.2019.8780171}.
\newblock URL \url{http://doi.org/10.1109/ICPHYS.2019.8780171}.

\bibitem[MacLeod et~al.(2017)MacLeod, Bennett, Morris, and
  Cutrell]{10.1145/3025453.3025814}
H.~MacLeod, C.~L. Bennett, M.~R. Morris, and E.~Cutrell.
\newblock Understanding blind people's experiences with computer-generated
  captions of social media images.
\newblock In \emph{Proceedings of the 2017 CHI Conference on Human Factors in
  Computing Systems}, CHI '17, page 5988–5999, New York, NY, USA, 2017.
  Association for Computing Machinery.
\newblock ISBN 9781450346559.
\newblock \doi{10.1145/3025453.3025814}.
\newblock URL \url{https://doi.org/10.1145/3025453.3025814}.

\bibitem[Makav and K{\i}l{\i}{\c{c}}(2019)]{makav2019new}
B.~Makav and V.~K{\i}l{\i}{\c{c}}.
\newblock A new image captioning approach for visually impaired people.
\newblock In \emph{2019 11th International Conference on Electrical and
  Electronics Engineering (ELECO)}, pages 945--949, Manhattan, New York, U.S.,
  2019. IEEE.
\newblock \doi{10.23919/ELECO47770.2019.8990630}.
\newblock URL \url{http://doi.org/10.23919/ELECO47770.2019.8990630}.

\bibitem[Mathews et~al.(2016)Mathews, Xie, and He]{mathews2016senticap}
A.~Mathews, L.~Xie, and X.~He.
\newblock Senticap: Generating image descriptions with sentiments.
\newblock In \emph{Proceedings of the Thirtieth AAAI Conference on Artificial
  Intelligence}, AAAI'16, page 3574–3580, Palo Alto, California, USA, 2016.
  AAAI Press.

\bibitem[Mirza and Osindero(2014)]{mirza2014conditional}
M.~Mirza and S.~Osindero.
\newblock Conditional generative adversarial nets.
\newblock \emph{arXiv preprint arXiv:1411.1784}, 2014.

\bibitem[Mokady et~al.(2021)Mokady, Hertz, and Bermano]{mokady2021clipcap}
R.~Mokady, A.~Hertz, and A.~H. Bermano.
\newblock Clipcap: Clip prefix for image captioning.
\newblock \emph{arXiv preprint arXiv:2111.09734}, 2021.
\newblock \doi{10.48550/ARXIV.2111.09734}.
\newblock URL \url{https://arxiv.org/abs/2111.09734}.

\bibitem[Nguyen et~al.(2022)Nguyen, Suganuma, and Okatani]{nguyen2022grit}
V.-Q. Nguyen, M.~Suganuma, and T.~Okatani.
\newblock Grit: Faster and better image captioning transformer using dual
  visual features.
\newblock In \emph{European Conference on Computer Vision}, pages 167--184.
  Springer, 2022.

\bibitem[Oord et~al.(2016)Oord, Kalchbrenner, Vinyals, Espeholt, Graves, and
  Kavukcuoglu]{oord2016conditional}
A.~v.~d. Oord, N.~Kalchbrenner, O.~Vinyals, L.~Espeholt, A.~Graves, and
  K.~Kavukcuoglu.
\newblock Conditional image generation with pixelcnn decoders.
\newblock In \emph{Advances in Neural Information Processing Systems},
  volume~29, pages 4790--4798, Red Hook, New York, USA, 2016. Curran
  Associates, Inc.
\newblock URL
  \url{https://proceedings.neurips.cc/paper/2016/file/b1301141feffabac455e1f90a7de2054-Paper.pdf}.

\bibitem[OpenAI(202)]{openai_gpt4}
OpenAI.
\newblock Gpt-4, Mar 202.
\newblock URL \url{https://openai.com/research/gpt-4}.

\bibitem[OpenAI(2022)]{openai_chatgpt}
OpenAI.
\newblock Introducing chatgpt, Nov 2022.
\newblock URL \url{https://openai.com/blog/chatgpt}.

\bibitem[Ordonez et~al.(2011)Ordonez, Kulkarni, and Berg]{ordonez2011im2text}
V.~Ordonez, G.~Kulkarni, and T.~Berg.
\newblock Im2text: Describing images using 1 million captioned photographs.
\newblock In J.~Shawe-Taylor, R.~Zemel, P.~Bartlett, F.~Pereira, and K.~Q.
  Weinberger, editors, \emph{Advances in Neural Information Processing
  Systems}, volume~24, pages 1143--1151, Red Hook, New York, USA, 2011. Curran
  Associates, Inc.
\newblock URL
  \url{https://proceedings.neurips.cc/paper/2011/file/5dd9db5e033da9c6fb5ba83c7a7ebea9-Paper.pdf}.

\bibitem[Pan et~al.(2020)Pan, Yao, Li, and Mei]{pan2020x}
Y.~Pan, T.~Yao, Y.~Li, and T.~Mei.
\newblock X-linear attention networks for image captioning.
\newblock In \emph{Proceedings of the IEEE/CVF Conference on Computer Vision
  and Pattern Recognition}, pages 10971--10980, Manhattan, New York, U.S.,
  2020. IEEE.
\newblock \doi{10.1109/CVPR42600.2020.01098}.
\newblock URL \url{http://doi.org/10.1109/CVPR42600.2020.01098}.

\bibitem[Papineni et~al.(2002)Papineni, Roukos, Ward, and
  Zhu]{papineni2002bleu}
K.~Papineni, S.~Roukos, T.~Ward, and W.-J. Zhu.
\newblock Bleu: a method for automatic evaluation of machine translation.
\newblock In \emph{Proceedings of the 40th Annual Meeting on Association for
  Computational Linguistics}, ACL '02, pages 311--318, Stroudsburg, PA, USA,
  2002. Association for Computational Linguistics.
\newblock \doi{10.3115/1073083.1073135}.
\newblock URL \url{https://doi.org/10.3115/1073083.1073135}.

\bibitem[Pascanu et~al.(2013)Pascanu, Mikolov, and
  Bengio]{pascanu2013difficulty}
R.~Pascanu, T.~Mikolov, and Y.~Bengio.
\newblock On the difficulty of training recurrent neural networks.
\newblock In \emph{Proceedings of the 30th International Conference on
  International Conference on Machine Learning - Volume 28}, ICML'13, page
  III–1310–III–1318, 12 Mountain Rock Ln, Norfolk, Massachusetts, 02056,
  United States, 2013. JMLR.

\bibitem[Pavlopoulos et~al.(2019)Pavlopoulos, Kougia, and
  Androutsopoulos]{pavlopoulos2019survey}
J.~Pavlopoulos, V.~Kougia, and I.~Androutsopoulos.
\newblock A survey on biomedical image captioning.
\newblock In \emph{Proceedings of the Second Workshop on Shortcomings in Vision
  and Language}, pages 26--36, Stroudsburg, PA, USA, 2019. Association for
  Computational Linguistics.
\newblock \doi{10.18653/v1/w19-1803}.
\newblock URL \url{http://doi.org/10.18653/v1/w19-1803}.

\bibitem[Plummer et~al.(2017)Plummer, Wang, Cervantes, Caicedo, Hockenmaier,
  and Lazebnik]{plummer2015flickr30k}
B.~A. Plummer, L.~Wang, C.~M. Cervantes, J.~C. Caicedo, J.~Hockenmaier, and
  S.~Lazebnik.
\newblock Flickr30k entities: Collecting region-to-phrase correspondences for
  richer image-to-sentence models.
\newblock \emph{International Journal of Computer Vision}, 123\penalty0
  (1):\penalty0 74--93, May 2017.
\newblock ISSN 1573-1405.
\newblock \doi{10.1007/s11263-016-0965-7}.
\newblock URL \url{https://doi.org/10.1007/s11263-016-0965-7}.

\bibitem[Pont-Tuset et~al.(2020)Pont-Tuset, Uijlings, Changpinyo, Soricut, and
  Ferrari]{PontTuset_eccv2020}
J.~Pont-Tuset, J.~Uijlings, S.~Changpinyo, R.~Soricut, and V.~Ferrari.
\newblock Connecting vision and language with localized narratives.
\newblock In A.~Vedaldi, H.~Bischof, T.~Brox, and J.-M. Frahm, editors,
  \emph{Computer Vision -- ECCV 2020}, pages 647--664, Manhattan, New York,
  USA, 2020. Springer International Publishing.
\newblock ISBN 978-3-030-58558-7.
\newblock \doi{10.1007/978-3-030-58558-7_38}.
\newblock URL \url{http://doi.org/10.1007/978-3-030-58558-7_38}.

\bibitem[Radford et~al.(2019)Radford, Wu, Child, Luan, Amodei, Sutskever,
  et~al.]{radford2019language}
A.~Radford, J.~Wu, R.~Child, D.~Luan, D.~Amodei, I.~Sutskever, et~al.
\newblock Language models are unsupervised multitask learners.
\newblock \emph{OpenAI blog}, 1\penalty0 (8):\penalty0 9, 2019.

\bibitem[Radford et~al.(2021)Radford, Kim, Hallacy, Ramesh, Goh, Agarwal,
  Sastry, Askell, Mishkin, Clark, et~al.]{radford2021learning}
A.~Radford, J.~W. Kim, C.~Hallacy, A.~Ramesh, G.~Goh, S.~Agarwal, G.~Sastry,
  A.~Askell, P.~Mishkin, J.~Clark, et~al.
\newblock Learning transferable visual models from natural language
  supervision.
\newblock In M.~Meila and T.~Zhang, editors, \emph{Proceedings of the 38th
  International Conference on Machine Learning}, volume 139 of
  \emph{Proceedings of Machine Learning Research}, pages 8748--8763. PMLR,
  18--24 Jul 2021.
\newblock URL \url{https://proceedings.mlr.press/v139/radford21a.html}.

\bibitem[Ranzato et~al.(2016)Ranzato, Chopra, Auli, and
  Zaremba]{ranzato2015sequence}
M.~Ranzato, S.~Chopra, M.~Auli, and W.~Zaremba.
\newblock Sequence level training with recurrent neural networks.
\newblock In \emph{4th International Conference on Learning Representations,
  ICLR 2016 - Conference Track Proceedings}, 2016.

\bibitem[Ren et~al.(2015)Ren, He, Girshick, and Sun]{ren2015faster}
S.~Ren, K.~He, R.~Girshick, and J.~Sun.
\newblock Faster r-cnn: Towards real-time object detection with region proposal
  networks.
\newblock In \emph{Proceedings of the 28th International Conference on Neural
  Information Processing Systems - Volume 1}, NIPS'15, pages 91--99, Cambridge,
  MA, USA, 2015. MIT Press.

\bibitem[Rennie et~al.(2017)Rennie, Marcheret, Mroueh, Ross, and
  Goel]{rennie2017self}
S.~J. Rennie, E.~Marcheret, Y.~Mroueh, J.~Ross, and V.~Goel.
\newblock Self-critical sequence training for image captioning.
\newblock In \emph{2017 IEEE Conference on Computer Vision and Pattern
  Recognition (CVPR)}, pages 1179--1195, Manhattan, New York, U.S., 2017. IEEE.
\newblock \doi{10.1109/CVPR.2017.131}.
\newblock URL \url{http://doi.org/10.1109/CVPR.2017.131}.

\bibitem[Rohrbach et~al.(2018)Rohrbach, Hendricks, Burns, Darrell, and
  Saenko]{rohrbach2018object}
A.~Rohrbach, L.~A. Hendricks, K.~Burns, T.~Darrell, and K.~Saenko.
\newblock Object hallucination in image captioning.
\newblock In \emph{Proceedings of the 2018 Conference on Empirical Methods in
  Natural Language Processing}, page 4035–4045, Stroudsburg, PA, USA, 2018.
  Association for Computational Linguistics.
\newblock \doi{10.18653/v1/D18-1437}.
\newblock URL \url{http://doi.org/10.18653/v1/D18-1437}.

\bibitem[Rumelhart et~al.(1988)Rumelhart, Hinton, and
  Williams]{rumelhart1985learning}
D.~E. Rumelhart, G.~E. Hinton, and R.~J. Williams.
\newblock \emph{Learning internal representations by error propagation}.
\newblock Morgan Kaufmann, Burlington, Massachusetts, 1988.
\newblock ISBN 978-1-4832-1446-7.
\newblock \doi{https://doi.org/10.1016/B978-1-4832-1446-7.50035-2}.
\newblock URL
  \url{https://www.sciencedirect.com/science/article/pii/B9781483214467500352}.

\bibitem[Scarselli et~al.(2008)Scarselli, Gori, Tsoi, Hagenbuchner, and
  Monfardini]{scarselli2008graph}
F.~Scarselli, M.~Gori, A.~C. Tsoi, M.~Hagenbuchner, and G.~Monfardini.
\newblock The graph neural network model.
\newblock \emph{IEEE Transactions on Neural Networks}, 20\penalty0
  (1):\penalty0 61--80, Dec. 2008.
\newblock ISSN 1941-0093.
\newblock \doi{10.1109/TNN.2008.2005605}.
\newblock URL \url{http://doi.org/10.1109/TNN.2008.2005605}.

\bibitem[Sharma et~al.(2018)Sharma, Ding, Goodman, and
  Soricut]{sharma2018conceptual}
P.~Sharma, N.~Ding, S.~Goodman, and R.~Soricut.
\newblock Conceptual captions: A cleaned, hypernymed, image alt-text dataset
  for automatic image captioning.
\newblock In \emph{Proceedings of the 56th Annual Meeting of the Association
  for Computational Linguistics (Volume 1: Long Papers)}, pages 2556--2565,
  Stroudsburg, PA, USA, July 2018. Association for Computational Linguistics.
\newblock \doi{10.18653/v1/P18-1238}.
\newblock URL \url{https://aclanthology.org/P18-1238}.

\bibitem[Shetty et~al.(2017)Shetty, Rohrbach, Anne~Hendricks, Fritz, and
  Schiele]{shetty2017speaking}
R.~Shetty, M.~Rohrbach, L.~Anne~Hendricks, M.~Fritz, and B.~Schiele.
\newblock Speaking the same language: Matching machine to human captions by
  adversarial training.
\newblock In \emph{2017 IEEE International Conference on Computer Vision
  (ICCV)}, pages 4155--4164, Manhattan, New York, USA, 2017. IEEE.
\newblock \doi{10.1109/ICCV.2017.445}.
\newblock URL \url{http://doi.org/10.1109/ICCV.2017.445}.

\bibitem[Shuster et~al.(2019)Shuster, Humeau, Hu, Bordes, and
  Weston]{shuster2019engaging}
K.~Shuster, S.~Humeau, H.~Hu, A.~Bordes, and J.~Weston.
\newblock Engaging image captioning via personality.
\newblock In \emph{Proceedings of the IEEE Computer Society Conference on
  Computer Vision and Pattern Recognition}, volume 2019-June, pages
  12516--12526, Manhattan, New York, U.S., 2019. IEEE.
\newblock \doi{10.1109/CVPR.2019.01280}.
\newblock URL \url{http://doi.org/10.1109/CVPR.2019.01280}.

\bibitem[Shutterstock(2019)]{shutterstock}
Shutterstock.
\newblock Stock images, photos, vectors, video and music | shutterstock, Sep
  2019.
\newblock URL \url{https://www.shutterstock.com/}.

\bibitem[Sidorov et~al.(2020)Sidorov, Hu, Rohrbach, and
  Singh]{sidorov2020textcaps}
O.~Sidorov, R.~Hu, M.~Rohrbach, and A.~Singh.
\newblock Textcaps: a dataset for image captioning with reading comprehension.
\newblock In A.~Vedaldi, H.~Bischof, T.~Brox, and J.-M. Frahm, editors,
  \emph{Computer Vision -- ECCV 2020}, pages 742--758, Manhattan, New York,
  USA, 2020. Springer International Publishing.
\newblock ISBN 978-3-030-58536-5.

\bibitem[Simonyan and Zisserman(2015)]{simonyan2014very}
K.~Simonyan and A.~Zisserman.
\newblock Very deep convolutional networks for large-scale image recognition.
\newblock \emph{3rd International Conference on Learning Representations,
  {ICLR} 2015, San Diego, CA, USA, May 7-9, 2015, Conference Track
  Proceedings}, 75, May 2015.
\newblock ISSN 15352900.

\bibitem[Szegedy et~al.(2016)Szegedy, Vanhoucke, Ioffe, Shlens, and
  Wojna]{szegedy2016rethinking}
C.~Szegedy, V.~Vanhoucke, S.~Ioffe, J.~Shlens, and Z.~Wojna.
\newblock Rethinking the inception architecture for computer vision.
\newblock In \emph{Proceedings of the IEEE conference on computer vision and
  pattern recognition}, pages 2818--2826, Manhattan, New York, U.S., 2016.
  IEEE.
\newblock \doi{10.1109/CVPR.2016.308}.
\newblock URL \url{http://doi.org/10.1109/CVPR.2016.308}.

\bibitem[Szegedy et~al.(2017)Szegedy, Ioffe, Vanhoucke, and
  Alemi]{szegedy2017inception}
C.~Szegedy, S.~Ioffe, V.~Vanhoucke, and A.~A. Alemi.
\newblock Inception-v4, inception-resnet and the impact of residual connections
  on learning.
\newblock In \emph{Thirty-first AAAI conference on artificial intelligence},
  AAAI'17, page 4278–4284, Palo Alto, California, USA, 2017. AAAI Press.
\newblock URL
  \url{https://www.aaai.org/ocs/index.php/AAAI/AAAI17/paper/view/14806}.

\bibitem[Tang et~al.(2019)Tang, Zhang, Wu, Luo, and Liu]{tang2019learning}
K.~Tang, H.~Zhang, B.~Wu, W.~Luo, and W.~Liu.
\newblock Learning to compose dynamic tree structures for visual contexts.
\newblock In \emph{2019 IEEE/CVF Conference on Computer Vision and Pattern
  Recognition (CVPR)}, pages 6612--6621, Manhattan, New York, U.S., 2019. IEEE.
\newblock \doi{10.1109/CVPR.2019.00678}.
\newblock URL \url{http://doi.org/10.1109/CVPR.2019.00678}.

\bibitem[Vaswani et~al.(2017)Vaswani, Shazeer, Parmar, Uszkoreit, Jones, Gomez,
  Kaiser, and Polosukhin]{vaswani2017attention}
A.~Vaswani, N.~Shazeer, N.~Parmar, J.~Uszkoreit, L.~Jones, A.~N. Gomez,
  {\L}.~Kaiser, and I.~Polosukhin.
\newblock Attention is all you need.
\newblock In I.~Guyon, U.~V. Luxburg, S.~Bengio, H.~Wallach, R.~Fergus,
  S.~Vishwanathan, and R.~Garnett, editors, \emph{Advances in neural
  information processing systems}, volume~30, pages 5998--6008, Red Hook, New
  York, USA, 2017. Curran Associates, Inc.

\bibitem[Vedantam et~al.(2015)Vedantam, Lawrence~Zitnick, and
  Parikh]{vedantam2015cider}
R.~Vedantam, C.~Lawrence~Zitnick, and D.~Parikh.
\newblock Cider: Consensus-based image description evaluation.
\newblock In \emph{Proceedings of the IEEE conference on computer vision and
  pattern recognition}, volume 07-12-June-2015, pages 4566--4575, Los Alamitos,
  CA, USA, 2015. IEEE Computer Society.
\newblock \doi{10.1109/CVPR.2015.7299087}.
\newblock URL
  \url{https://doi.ieeecomputersociety.org/10.1109/CVPR.2015.7299087}.

\bibitem[Vinyals et~al.(2015)Vinyals, Toshev, Bengio, and
  Erhan]{vinyals2015show}
O.~Vinyals, A.~Toshev, S.~Bengio, and D.~Erhan.
\newblock Show and tell: A neural image caption generator.
\newblock In \emph{Proceedings of the IEEE conference on computer vision and
  pattern recognition}, pages 3156--3164, Los Alamitos, CA, USA, jun 2015. IEEE
  Computer Society.
\newblock \doi{10.1109/CVPR.2015.7298935}.
\newblock URL
  \url{https://doi.ieeecomputersociety.org/10.1109/CVPR.2015.7298935}.

\bibitem[Vinyals et~al.(2016)Vinyals, Blundell, Lillicrap, Wierstra,
  et~al.]{vinyals2016matching}
O.~Vinyals, C.~Blundell, T.~Lillicrap, D.~Wierstra, et~al.
\newblock Matching networks for one shot learning.
\newblock In D.~Lee, M.~Sugiyama, U.~Luxburg, I.~Guyon, and R.~Garnett,
  editors, \emph{Advances in neural information processing systems}, volume~29,
  pages 3630--3638, Red Hook, New York, USA, 2016. Curran Associates, Inc.
\newblock URL
  \url{https://proceedings.neurips.cc/paper/2016/file/90e1357833654983612fb05e3ec9148c-Paper.pdf}.

\bibitem[Wang et~al.(2019)Wang, Beck, and Cohn]{wang2019role}
D.~Wang, D.~Beck, and T.~Cohn.
\newblock On the role of scene graphs in image captioning.
\newblock In \emph{Proceedings of the Beyond Vision and LANguage: inTEgrating
  Real-world kNowledge (LANTERN)}, pages 29--34, Stroudsburg, PA, USA, 2019.
  Association for Computational Linguistics.
\newblock \doi{10.18653/v1/D19-6405}.
\newblock URL \url{http://doi.org/10.18653/v1/D19-6405}.

\bibitem[Wang et~al.(2020)Wang, Wang, Wang, Wang, Feng, and
  Tan]{wang2020learning}
J.~Wang, W.~Wang, L.~Wang, Z.~Wang, D.~D. Feng, and T.~Tan.
\newblock Learning visual relationship and context-aware attention for image
  captioning.
\newblock \emph{Pattern Recognition}, 98, Feb. 2020.
\newblock ISSN 0031-3203.
\newblock \doi{10.1016/j.patcog.2019.107075}.
\newblock URL \url{https://doi.org/10.1016/j.patcog.2019.107075}.

\bibitem[Wang and Chan(2018)]{wang2018cnn}
Q.~Wang and A.~B. Chan.
\newblock Cnn+ cnn: Convolutional decoders for image captioning.
\newblock \emph{CoRR}, abs/1805.09019, 2018.
\newblock URL \url{http://arxiv.org/abs/1805.09019}.

\bibitem[Wang et~al.(2022)Wang, Xu, and Sun]{wang2022end}
Y.~Wang, J.~Xu, and Y.~Sun.
\newblock End-to-end transformer based model for image captioning.
\newblock \emph{Proceedings of the AAAI Conference on Artificial Intelligence},
  36\penalty0 (3):\penalty0 2585--2594, Jun. 2022.
\newblock \doi{10.1609/aaai.v36i3.20160}.
\newblock URL \url{https://ojs.aaai.org/index.php/AAAI/article/view/20160}.

\bibitem[Wang et~al.(2018)Wang, Liu, Zeng, and Yuille]{wang2018scene}
Y.-S. Wang, C.~Liu, X.~Zeng, and A.~Yuille.
\newblock Scene graph parsing as dependency parsing.
\newblock In \emph{Proceedings of the 2018 Conference of the North American
  Chapter of the Association for Computational Linguistics: Human Language
  Technologies, Volume 1 (Long Papers)}, pages 397–--407, Stroudsburg, PA,
  USA, 2018. Association for Computational Linguistics.
\newblock \doi{10.18653/v1/N18-1037}.
\newblock URL \url{http://doi.org/10.18653/v1/N18-1037}.

\bibitem[Wu et~al.(2016)Wu, Shen, Liu, Dick, and Van Den~Hengel]{wu2016value}
Q.~Wu, C.~Shen, L.~Liu, A.~Dick, and A.~Van Den~Hengel.
\newblock What value do explicit high level concepts have in vision to language
  problems?
\newblock In \emph{2016 IEEE Conference on Computer Vision and Pattern
  Recognition (CVPR)}, pages 203--212, Manhattan, New York, USA, 2016. IEEE.
\newblock \doi{10.1109/CVPR.2016.29}.
\newblock URL \url{http://doi.org/10.1109/CVPR.2016.29}.

\bibitem[Xia et~al.(2021)Xia, Huang, Duan, Zhang, Ji, Sui, Cui, Bharti, and
  Zhou]{xia2021xgpt}
Q.~Xia, H.~Huang, N.~Duan, D.~Zhang, L.~Ji, Z.~Sui, E.~Cui, T.~Bharti, and
  M.~Zhou.
\newblock Xgpt: Cross-modal generative pre-training for image captioning.
\newblock In \emph{Natural Language Processing and Chinese Computing: 10th CCF
  International Conference, NLPCC 2021, Qingdao, China, October 13–17, 2021,
  Proceedings, Part I}, pages 786--797, Berlin, Heidelberg, 2021. Springer,
  Springer-Verlag.
\newblock ISBN 978-3-030-88479-6.
\newblock \doi{10.1007/978-3-030-88480-2_63}.
\newblock URL \url{https://doi.org/10.1007/978-3-030-88480-2_63}.

\bibitem[Xu et~al.(2017)Xu, Zhu, Choy, and Fei-Fei]{xu2017scene}
D.~Xu, Y.~Zhu, C.~B. Choy, and L.~Fei-Fei.
\newblock Scene graph generation by iterative message passing.
\newblock In \emph{2017 IEEE Conference on Computer Vision and Pattern
  Recognition (CVPR)}, pages 3097--3106, Manhattan, New York, U.S., 2017. IEEE.
\newblock \doi{10.1109/CVPR.2017.330}.
\newblock URL \url{http://doi.org/10.1109/CVPR.2017.330}.

\bibitem[Xu et~al.(2015)Xu, Ba, Kiros, Cho, Courville, Salakhudinov, Zemel, and
  Bengio]{xu2015show}
K.~Xu, J.~Ba, R.~Kiros, K.~Cho, A.~Courville, R.~Salakhudinov, R.~Zemel, and
  Y.~Bengio.
\newblock Show, attend and tell: Neural image caption generation with visual
  attention.
\newblock In F.~Bach and D.~Blei, editors, \emph{Proceedings of the 32nd
  International Conference on Machine Learning}, volume~37 of \emph{Proceedings
  of Machine Learning Research}, pages 2048--2057, Lille, France, 07--09 Jul
  2015. PMLR.

\bibitem[Xu et~al.(2019)Xu, Liu, Liu, Nie, and Su]{xu2019scene}
N.~Xu, A.-A. Liu, J.~Liu, W.~Nie, and Y.~Su.
\newblock Scene graph captioner: Image captioning based on structural visual
  representation.
\newblock \emph{Journal of Visual Communication and Image Representation},
  58:\penalty0 477--485, Jan. 2019.
\newblock ISSN 1047-3203.
\newblock \doi{https://doi.org/10.1016/j.jvcir.2018.12.027}.
\newblock URL \url{http://doi.org/10.1016/j.jvcir.2018.12.027}.

\bibitem[Xu et~al.(2020)Xu, Zhang, Liu, Nie, Su, Nie, and Zhang]{xu2019multi}
N.~Xu, H.~Zhang, A.-A. Liu, W.~Nie, Y.~Su, J.~Nie, and Y.~Zhang.
\newblock Multi-level policy and reward-based deep reinforcement learning
  framework for image captioning.
\newblock \emph{IEEE Transactions on Multimedia}, 22\penalty0 (5):\penalty0
  1372--1383, Sept. 2020.
\newblock \doi{10.1109/TMM.2019.2941820}.
\newblock URL \url{http://doi.org/10.1109/TMM.2019.2941820}.

\bibitem[Xu et~al.(2022)Xu, Li, Xu, Huang, Huang, and Cai]{xu2022image}
Y.~Xu, L.~Li, H.~Xu, S.~Huang, F.~Huang, and J.~Cai.
\newblock Image captioning in the transformer age.
\newblock \emph{arXiv preprint arXiv:2204.07374}, 2022.
\newblock \doi{10.48550/ARXIV.2204.07374}.
\newblock URL \url{https://arxiv.org/abs/2204.07374}.

\bibitem[Yang et~al.(2018)Yang, Lu, Lee, Batra, and Parikh]{yang2018graph}
J.~Yang, J.~Lu, S.~Lee, D.~Batra, and D.~Parikh.
\newblock Graph r-cnn for scene graph generation.
\newblock In V.~Ferrari, M.~Hebert, C.~Sminchisescu, and Y.~Weiss, editors,
  \emph{Computer Vision -- ECCV 2018}, pages 690--706, Manhattan, New York,
  USA, 2018. Springer International Publishing.
\newblock ISBN 978-3-030-01246-5.
\newblock \doi{10.1007/978-3-030-01246-5_41}.
\newblock URL \url{http://doi.org/10.1007/978-3-030-01246-5_41}.

\bibitem[Yang et~al.(2019)Yang, Tang, Zhang, and Cai]{yang2019auto}
X.~Yang, K.~Tang, H.~Zhang, and J.~Cai.
\newblock Auto-encoding scene graphs for image captioning.
\newblock In \emph{Proceedings of the IEEE/CVF Conference on Computer Vision
  and Pattern Recognition}, pages 10685--10694, Manhattan, New York, U.S.,
  2019. IEEE.
\newblock \doi{10.1109/CVPR.2019.01094}.
\newblock URL \url{http://doi.org/10.1109/CVPR.2019.01094}.

\bibitem[Yang et~al.(2022)Yang, Liu, and Wang]{yang2022reformer}
X.~Yang, Y.~Liu, and X.~Wang.
\newblock Reformer: The relational transformer for image captioning.
\newblock In \emph{Proceedings of the 30th ACM International Conference on
  Multimedia}, MM '22, page 5398–5406, New York, NY, USA, 2022. Association
  for Computing Machinery.
\newblock ISBN 9781450392037.
\newblock \doi{10.1145/3503161.3548409}.
\newblock URL \url{https://doi.org/10.1145/3503161.3548409}.

\bibitem[Yang et~al.(2016)Yang, He, Gao, Deng, and Smola]{yang2016stacked}
Z.~Yang, X.~He, J.~Gao, L.~Deng, and A.~Smola.
\newblock Stacked attention networks for image question answering.
\newblock In \emph{Proceedings of the IEEE conference on computer vision and
  pattern recognition}, volume 2016-December, pages 21--29, Manhattan, New
  York, U.S., 2016. IEEE.
\newblock \doi{10.1109/CVPR.2016.10}.
\newblock URL \url{https://doi.org/10.1109/CVPR.2016.10}.

\bibitem[Yao et~al.(2017)Yao, Pan, Li, Qiu, and Mei]{yao2017boosting}
T.~Yao, Y.~Pan, Y.~Li, Z.~Qiu, and T.~Mei.
\newblock Boosting image captioning with attributes.
\newblock In \emph{2017 IEEE International Conference on Computer Vision
  (ICCV)}, pages 4904--4912, Manhattan, New York, USA, 2017. IEEE.
\newblock \doi{10.1109/ICCV.2017.524}.
\newblock URL \url{http://doi.org/10.1109/ICCV.2017.524}.

\bibitem[Yao et~al.(2018)Yao, Pan, Li, and Mei]{yao2018exploring}
T.~Yao, Y.~Pan, Y.~Li, and T.~Mei.
\newblock Exploring visual relationship for image captioning.
\newblock In V.~Ferrari, M.~Hebert, C.~Sminchisescu, and Y.~Weiss, editors,
  \emph{Computer Vision -- ECCV 2018}, pages 711--727, Manhattan, New York,
  USA, 2018. Springer International Publishing.
\newblock ISBN 978-3-030-01264-9.
\newblock \doi{10.1007/978-3-030-01264-9_42}.

\bibitem[You et~al.(2016)You, Jin, Wang, Fang, and Luo]{you2016image}
Q.~You, H.~Jin, Z.~Wang, C.~Fang, and J.~Luo.
\newblock Image captioning with semantic attention.
\newblock In \emph{2016 IEEE Conference on Computer Vision and Pattern
  Recognition (CVPR)}, pages 4651--4659, Manhattan, New York, USA, 2016. IEEE.
\newblock \doi{10.1109/CVPR.2016.503}.
\newblock URL \url{http://doi.org/10.1109/CVPR.2016.503}.

\bibitem[Young et~al.(2014)Young, Lai, Hodosh, and Hockenmaier]{young2014image}
P.~Young, A.~Lai, M.~Hodosh, and J.~Hockenmaier.
\newblock From image descriptions to visual denotations: New similarity metrics
  for semantic inference over event descriptions.
\newblock \emph{Transactions of the Association for Computational Linguistics},
  2:\penalty0 67--78, February 2014.
\newblock ISSN 2307-387X.
\newblock \doi{10.1162/tacl_a_00166}.
\newblock URL \url{https://doi.org/10.1162/tacl_a_00166}.

\bibitem[Zellers et~al.(2018)Zellers, Yatskar, Thomson, and
  Choi]{zellers2018neural}
R.~Zellers, M.~Yatskar, S.~Thomson, and Y.~Choi.
\newblock Neural motifs: Scene graph parsing with global context.
\newblock In \emph{2018 IEEE/CVF Conference on Computer Vision and Pattern
  Recognition}, pages 5831--5840, Manhattan, New York, U.S., 2018. IEEE.
\newblock \doi{10.1109/CVPR.2018.00611}.
\newblock URL \url{http://doi.org/10.1109/CVPR.2018.00611}.

\bibitem[Zeng et~al.(2022)Zeng, Zhang, Song, and Gao]{zeng2022s2}
P.~Zeng, H.~Zhang, J.~Song, and L.~Gao.
\newblock S2 transformer for image captioning.
\newblock In L.~D. Raedt, editor, \emph{Proceedings of the Thirty-First
  International Joint Conference on Artificial Intelligence, {IJCAI-22}}, pages
  1608--1614. International Joint Conferences on Artificial Intelligence
  Organization, July 2022.
\newblock \doi{10.24963/ijcai.2022/224}.
\newblock URL \url{https://doi.org/10.24963/ijcai.2022/224}.

\bibitem[Zhang et~al.(2017)Zhang, Kyaw, Chang, and Chua]{zhang2017visual}
H.~Zhang, Z.~Kyaw, S.-F. Chang, and T.-S. Chua.
\newblock Visual translation embedding network for visual relation detection.
\newblock In \emph{2017 IEEE Conference on Computer Vision and Pattern
  Recognition (CVPR)}, pages 3107--3115, Manhattan, New York, U.S., 2017. IEEE.
\newblock \doi{10.1109/CVPR.2017.331}.
\newblock URL \url{http://doi.org/10.1109/CVPR.2017.331}.

\bibitem[Zhang et~al.(2021)Zhang, Li, Hu, Yang, Zhang, Wang, Choi, and
  Gao]{zhang2021vinvl}
P.~Zhang, X.~Li, X.~Hu, J.~Yang, L.~Zhang, L.~Wang, Y.~Choi, and J.~Gao.
\newblock Vinvl: Revisiting visual representations in vision-language models.
\newblock In \emph{Proceedings of the IEEE/CVF Conference on Computer Vision
  and Pattern Recognition}, pages 5579--5588, Manhattan, New York, U.S., 2021.
  IEEE.
\newblock \doi{10.1109/CVPR46437.2021.00553}.
\newblock URL \url{http://doi.org/10.1109/CVPR46437.2021.00553}.

\bibitem[Zheng et~al.(2021)Zheng, Lu, Zhao, Zhu, Luo, Wang, Fu, Feng, Xiang,
  Torr, et~al.]{zheng2021rethinking}
S.~Zheng, J.~Lu, H.~Zhao, X.~Zhu, Z.~Luo, Y.~Wang, Y.~Fu, J.~Feng, T.~Xiang,
  P.~H. Torr, et~al.
\newblock Rethinking semantic segmentation from a sequence-to-sequence
  perspective with transformers.
\newblock In \emph{2021 IEEE/CVF Conference on Computer Vision and Pattern
  Recognition (CVPR)}, pages 6877--6886, Los Alamitos, CA, USA, jun 2021. IEEE
  Computer Society.
\newblock \doi{10.1109/CVPR46437.2021.00681}.
\newblock URL
  \url{https://doi.ieeecomputersociety.org/10.1109/CVPR46437.2021.00681}.

\bibitem[Zhong et~al.(2020)Zhong, Wang, Chen, Yu, and
  Li]{zhong2020comprehensive}
Y.~Zhong, L.~Wang, J.~Chen, D.~Yu, and Y.~Li.
\newblock Comprehensive image captioning via scene graph decomposition.
\newblock In A.~Vedaldi, H.~Bischof, T.~Brox, and J.-M. Frahm, editors,
  \emph{Computer Vision -- ECCV 2020}, pages 211--229, Manhattan, New York,
  USA, 2020. Springer International Publishing.
\newblock ISBN 978-3-030-58568-6.
\newblock \doi{https://doi.org/10.1007/978-3-030-58568-6_13}.
\newblock URL
  \url{http://doi.org/https://doi.org/10.1007/978-3-030-58568-6_13}.

\bibitem[Zhou et~al.(2017)Zhou, Xu, Koch, and Corso]{zhou2016image}
L.~Zhou, C.~Xu, P.~Koch, and J.~J. Corso.
\newblock Watch what you just said: Image captioning with text-conditional
  attention.
\newblock In \emph{Proceedings of the on Thematic Workshops of ACM Multimedia
  2017}, Thematic Workshops '17, page 305–313, New York, NY, USA, 2017.
  Association for Computing Machinery.
\newblock ISBN 9781450354165.
\newblock \doi{10.1145/3126686.3126717}.
\newblock URL \url{https://doi.org/10.1145/3126686.3126717}.

\bibitem[Zhou et~al.(2020)Zhou, Palangi, Zhang, Hu, Corso, and
  Gao]{zhou2020unified}
L.~Zhou, H.~Palangi, L.~Zhang, H.~Hu, J.~Corso, and J.~Gao.
\newblock Unified vision-language pre-training for image captioning and vqa.
\newblock \emph{AAAI 2020 - 34th AAAI Conference on Artificial Intelligence},
  34\penalty0 (07):\penalty0 13041--13049, Apr. 2020.
\newblock ISSN 2159-5399.
\newblock \doi{10.1609/aaai.v34i07.7005}.
\newblock URL \url{http://doi.org/10.1609/aaai.v34i07.7005}.

\bibitem[Zhu et~al.(2017)Zhu, Park, Isola, and Efros]{zhu2017unpaired}
J.-Y. Zhu, T.~Park, P.~Isola, and A.~A. Efros.
\newblock Unpaired image-to-image translation using cycle-consistent
  adversarial networks.
\newblock In \emph{Proceedings of the IEEE international conference on computer
  vision}, pages 2223--2232, Manhattan, New York, U.S., 2017. IEEE.
\newblock \doi{10.1109/ICCV.2017.244}.
\newblock URL \url{http://doi.org/10.1109/ICCV.2017.244}.

\bibitem[Zhu et~al.(2020)Zhu, Su, Lu, Li, Wang, and Dai]{zhu2020deformable}
X.~Zhu, W.~Su, L.~Lu, B.~Li, X.~Wang, and J.~Dai.
\newblock Deformable detr: Deformable transformers for end-to-end object
  detection.
\newblock \emph{arXiv preprint arXiv:2010.04159}, 2020.

\end{thebibliography}






\end{document}